\documentclass[journal]{IEEEtran}
\usepackage{times}
\usepackage{subfigure}
\usepackage{epsfig}
\usepackage{graphicx}
\usepackage{amsthm}
\usepackage{amsmath}
\usepackage{amssymb}
\usepackage{algorithm,algpseudocode}

\newtheorem{theorem}{Theorem}

\newtheorem{lemma}{Lemma}
\newtheorem{definition}{Definition}
\newtheorem{assumption}{Assumption}
\newtheorem{corollary}{Corollary}

\usepackage{datetime}

\usepackage{amsthm}

\usepackage{color}

\def\b1{\color{black}}
\usepackage{multirow}
\usepackage{hhline}

\allowdisplaybreaks

\usepackage[noadjust]{cite}
\usepackage[sort&compress,numbers]{natbib}
\usepackage{tabu}

\ifCLASSOPTIONcompsoc
  \usepackage[nocompress]{cite}
\else
  \usepackage{cite}
\fi
\ifCLASSINFOpdf
\else
\fi
\usepackage{scalerel}
\usepackage{tikz}
\usetikzlibrary{svg.path}

\definecolor{orcidlogocol}{HTML}{A6CE39}
\tikzset{
    orcidlogo/.pic={
        \fill[orcidlogocol] svg{M256,128c0,70.7-57.3,128-128,128C57.3,256,0,198.7,0,128C0,57.3,57.3,0,128,0C198.7,0,256,57.3,256,128z};
        \fill[white] svg{M86.3,186.2H70.9V79.1h15.4v48.4V186.2z}
        svg{M108.9,79.1h41.6c39.6,0,57,28.3,57,53.6c0,27.5-21.5,53.6-56.8,53.6h-41.8V79.1z M124.3,172.4h24.5c34.9,0,42.9-26.5,42.9-39.7c0-21.5-13.7-39.7-43.7-39.7h-23.7V172.4z}
        svg{M88.7,56.8c0,5.5-4.5,10.1-10.1,10.1c-5.6,0-10.1-4.6-10.1-10.1c0-5.6,4.5-10.1,10.1-10.1C84.2,46.7,88.7,51.3,88.7,56.8z};
    }
}

\newcommand\orcidicon[1]{\href{https://orcid.org/#1}{\mbox{\scalerel*{
                \begin{tikzpicture}[yscale=-1,transform shape]
                \pic{orcidlogo};
                \end{tikzpicture}
            }{|}}}}
\begin{document}

%
\title{Stochastic Optimization for Non-convex Problem with Inexact Hessian Matrix, Gradient, and Function}
%
%
%
%

\author{Liu~Liu,~\IEEEmembership{Member,~IEEE,}
		Xuanqing~Liu,
		Cho-Jui~Hsieh,
		Dacheng~Tao,~\IEEEmembership{Fellow,~IEEE}
\IEEEcompsocitemizethanks{
	\IEEEcompsocthanksitem 
 L. Liu is with the Institute of Artificial Intelligence and State Key Lab of Software Development Environment, Beihang University (e-mail: liuliubh@gmail.com).
	\IEEEcompsocthanksitem 
	X. Liu and C. Hsieh are with  University of California, Los Angeles 
	(e-mail: xqliu@cs.ucla.edu; chohsieh@cs.ucla.edu)
  \IEEEcompsocthanksitem 
  D. Tao is with School of Computer Science, Sydney AI Centre, Faculty of Engineering, the University of Sydney,Sydney, NSW 2008, Australia 
	(e-mail: dacheng.tao@gmail.com).
 \IEEEcompsocthanksitem 
    Corresponding Author: Dacheng Tao
    }
}
\IEEEtitleabstractindextext{
\begin{abstract}
Trust-region (TR) and adaptive regularization using cubics (ARC) have proven to have some very appealing theoretical properties for non-convex optimization by concurrently computing function value, gradient, and Hessian matrix to obtain the next search direction and the adjusted parameters. Although stochastic approximations help largely reduce the computational cost, it is challenging to theoretically guarantee the convergence rate. In this paper, we explore a family of stochastic TR and ARC methods that can simultaneously provide inexact computations of the Hessian matrix, gradient, and function values. Our algorithms require much fewer propagations overhead per iteration than TR and ARC. We prove that the iteration complexity to achieve $\epsilon$-approximate second-order optimality is of the same order as the exact computations demonstrated in previous studies. Additionally, the mild conditions on inexactness can be met by leveraging a random sampling technology in the finite-sum minimization problem. Numerical experiments with a non-convex problem support these findings and demonstrate that, with the same or a similar number of iterations, our algorithms require less computational overhead per iteration than current second-order methods.
\end{abstract}

}

\maketitle

\IEEEdisplaynontitleabstractindextext

%
\IEEEpeerreviewmaketitle

\section{Introduction}\label{sec:introduction}

\IEEEPARstart{I}{n} this paper, we consider the following unconstrained optimization problem:
\begin{align}
\label{Newton:Problem}
    \mathop {\min }\limits_{x \in {\mathbb R^d}} f( x ) 
    =
    \frac{1}{n}\sum\nolimits_{i = 1}^n {{f_i}( x )},
\end{align}
where $f_i(x)$ is smooth but not necessarily convex. 
Optimization problems with this type of finite-sum structure are becoming increasingly popular in modern machine learning tasks, especially in deep learning, where each $f_i(\cdot)$ corresponds to the loss of a training sample and  $n$ is the number of data samples.
However, computing a full gradient and a Hessian matrix is cost-prohibitive for most large-scale tasks. 
Hence, first-order methods, such as gradient descent and stochastic gradient descent (SGD), have become the “go-to” methods for solving these problems because they are guaranteed to converge to stationary points in expectation.  Unfortunately, these stationary points can be saddle points or local minima, which could not be distinguished by only using the first-order information without the noise added\footnote{Some recent analyses indicate that SGD can escape from saddle points in certain cases~\cite{ge2017no,jin2017escape}, i.e., add noise to the stochastic gradient, but SGD is not the focus of this paper.}.

It is known that second-order methods can generally and more easily detect the negative direction curvature through the use of Hessian information. At each iteration, second-order methods typically build a quadratic approximation function around the current solution $x_k$, such as
\begin{align}
    f( {{x_k}} ) + \langle 
    {{\nabla f(x_k)},s} \rangle  + \frac{1}{2}\langle {s,{\nabla^2f(x_k)}s} \rangle,
\end{align}
where $\nabla f(x_k)$ is the gradient and $\nabla^2 f(x_k)$ is the Hessian matrix. 
A common strategy to update the current solution is to minimize this quadratic approximation within a small region. 

Algorithms based on this idea, such as the trust-region method (TR) \cite{conn2000trust} and adaptive regularization using cubics (ARC) \cite{cartis2011adaptivea,cartis2011adaptiveb}, the second-order methods have demonstrated promising performance on non-convex problems. 
However, with large-scale optimization problems, such as training deep neural networks, it is impossible to compute the gradient and Hessian matrix exactly in every iteration. 
As a result, some researchers have begun to explore inexact second-order methods.
The second-order method with stochastic Hessian (SH-) \cite{chen2018adaptive,bellavia2021adaptive, xu2017second, xu2017newton} use an inexact Hessian matrix computed from some samples rather than all the samples for TR and ARC, and demonstrated that the subsampled Hessian matrix did not affect the convergent rate.
The second-order information is approximated by the subsampled Hessian matrix, yet the gradient is still computed exactly. 
The second-order method with stochastic Hessian and stochastic gradient (SHG-) \cite{cartis2012oracle,kohler2017sub,cartis2018global,tripuraneni2018stochastic} makes use of inexact information of Hessian and gradient. 
Although the approaches mentioned above maintain the same order of convergence rate,
the accurate value of the function sometimes is difficult or impossible to obtain to adjust parameters. 
Recently, second-order methods with stochastic Hessian, stochastic gradient, and stochastic function value (SHGF-)~\cite{bergou2022subsampling, bellavia2021adaptive} employ the inexact information from  Hessian, gradient, and function value.

In this paper,  we devise simple and practical stochastic versions of the trust-region (STR) and adaptive regularization using cubics (SARC) methods, which are different from  \cite{bergou2022subsampling, bellavia2019adaptive} in defining the ratio and proof process. We aim to inexactly compute all the Hessian, gradient, and function value given a fixed sample size for TR and ARC while retaining the convergence order. The idea is to take the quadratic approximation in \eqref{Newton:Definition_m} and replace $\nabla f(x)$ with an approximate gradient $g(x_k)$ and $\nabla^2 f(x)$ with a Hessian approximation $B(x)$ that has a fixed approximation error.  Further, we also consider the most general case where the trust-region radius or cubic regularization parameter is adaptively adjusted by checking the \textit{subsampled} objective function value. This strategy can be achieved using fixed sample size, and the error bound can be approximated using particular sampling bounds, which avoid the proof difficulty caused by the approximated function, gradient, and Hessian.
Here, we present the iteration complexity\footnote{We use $\mathcal{O}(\cdot)$ to hide constant factors. Parameters are defined in Sec. \ref{Newton:Section:Preliminary}.} for both proposed methods STR and SARC, in which  parameters are defined in  Preliminary section.

\begin{itemize}
	\item 
    	\textbf{{STR}}: The total number of iterations is 
    	\begin{align*}
        	\mathcal{O}( {\text{max}}\{ ( \epsilon _H - {\epsilon _B} )^{-1}( \epsilon _{\nabla f} - \epsilon _g )^{ - 2},( \epsilon _H - \epsilon _B )^{ - 3} \} ).
    	\end{align*}
	\item 
    	\textbf{{SARC}}: The base iteration complexity of SARC is of the same order as that of STR.  However, if the conditions of the terminal criterion in  (\ref{Newton:SARC:Assumption_s1})-(\ref{Newton:SARC:Assumption_s3}) are satisfied, then the total number of iterations of SARC becomes 
    	\[\mathcal{O}( \text{max} \{ (\epsilon_{\nabla f} - \epsilon_g)^{ - 3/2},(\epsilon_H - \epsilon_B)^{ - 3}\}  ).\]
\end{itemize}

Note that we do not claim the above strategy induces  ``new"  STR and SARC algorithms because translating TR and ARC to the stochastic setting is an instinctive concept. The question is whether this simple strategy works in theory and in practice for STR and SARC. We affirmatively answer this question, and our contributions are summarized as follows:

\begin{itemize}	
	\item We provide a theoretical analysis of the convergence and iteration complexity for STR and SARC. In contrast to \cite{kohler2017sub}, the approximation errors of the Hessian matrix and the gradient estimation do not need to be related to the update step.
	In our framework, the function value, gradient, and Hessian matrix are all inexact, but that of \cite{xu2017newton, xu2017second} does not allow for an inexact gradient and the function value. 
	Similar to \cite{xu2017newton, xu2017second}, our proof of the framework follows the convention. 
	Even when the function value, gradient, and Hessian matrix are all inexact, we are able to show that the iteration complexity is of the same order as the current exact methods. 
	\item 	We present a new modified parameter $\tilde\rho$ within STR and SARC that can be used to determine whether or not  $\hat\rho$ has changed. 
	In addition, by combining the estimated function with the new modified sub-sampling rule, we guarantee the estimated gradient within a small difference to the ``original" gradient under our framework, which is the key element in the theoretical analysis. However, this is not guaranteed in general.
	\item	
	To deal with the difference and difficulty brought by the two extra inexact terms, we present a new yet non-trivial sub-sampling technique in the proof. 
	Specifically, we introduce an upper bound for the subsampled function value based on the operator Bernstein inequality, which means the TR radius and ARC parameter can be automatically adjusted using the subsamples. We also present the corresponding probabilistic convergence analysis.
\end{itemize}
The remainder of the paper is organized as follows. Section \ref{Newton:Section:relatedwork} presents the related works. Section \ref{Newton:Section:Preliminary} provides the preliminaries about the assumptions and definition.  
Section \ref{Newton:Section:STR} and  \ref{Newton:Section:SARC} present the stochastic trust region method and cubic regularization method, respectively, along with their convergence and iteration complexity.  
The sub-sampling method for estimating the corresponding function, gradient, and Hessian, and probabilistic convergence analysis are presented in Section \ref{Newton:Section:subsampling}.
Section \ref{Newton:Section:Experiment} presents the experimental results. Section \ref{Newton:Section:Conclusion} concludes our paper and Section \ref{Newton:Appendix:theorem} presents the proof process of the complexity iteration of the proposed algorithm.

\section{Related Works}\label{Newton:Section:relatedwork}
With the increasing size of data and models, stochastic optimization has become more popular, since computing the gradient and Hessian are prohibitively expensive. 
The simplicity and effectiveness of SGD \cite{zhang2004solving, 9512549} certainly make it the most popular stochastic first-order optimization method, especially for training deep neural networks and large-scale machine learning models. However, estimating a gradient inevitably induces noise such that the variance of the gradient may not approximate zero even when converging to a stationary point. Stochastic variance reduction gradient (SVRG) \cite{johnson2013accelerating} and SAGA \cite{defazio2014saga,8464090} are the two most common methods used to reduce the variance of the gradient estimator. Moreover, both lead to faster convergence, especially in the convex setting. Several other related variance reduction methods have been developed and analyzed for non-convex problems \cite{reddi2016stochastic,NIPS2017_6829}. However, the algorithms did not consider of escaping the saddle points.


Within second-order optimization techniques, there are several types of methods. Newton-type methods \cite{marteau2019globally,crane2019dingo, 9756647,9675797,9640489} rely on building a quadratic approximation around the current solution, avoiding saddle points by exploring the curvature information in non-convex optimization. To overcome the overhead of computing an exact Hessian matrix, the Broyden-Fletcher-Goldfarb-Shanno (BFGS) and Limited-BFGS \cite{nocedal1980updating} methods approximate the Hessian matrix using first-order information. Another essential technique for Hessian approximation is the subsampling function $f_i(\cdot)$ to obtain the estimated Hessian matrix. 
Both Richard \textit{et al.}~\cite{byrd2011use} 
and 
Murat
\cite{erdogdu2015convergence} used a stochastic Hessian matrix to reach global convergence. 
Richard \textit{et al.}~\cite{byrd2011use}, however,  required the $f_i(\cdot)$ to be smooth and strongly convex. 

The TR is a classical second-order method that searches the update direction only within a trusted region around the current point. The size of the trusted region, which is critical to the effectiveness of the search direction, is updated based on a measurement of whether or not the quadratic approximation is an adequate representation of the function. 
Other approaches outlined in 
\cite{xu2017newton, xu2017second}, which incorporate a subsampling based inexact Hessian matrix with TR.
Similar to TR methods, ARC methods  replaced an approximate matrix with a Hessian matrix \cite{cartis2011adaptivea,cartis2011adaptiveb} for unconstrained optimization. 
Kohler and Lucchi \cite{kohler2017sub}  applied an operator-Bernstein inequality to approximate the Hessian matrix and gradient to form a quadratic function in cubic regularization methods.
However, the sample approximate condition is subject to the search direction $s$. 
Thus, they need to increase the sample size at each step. To overcome this issue, 
Xu et al. 
\cite{xu2017newton} provided an alternative approximation condition for a Hessian matrix that does not depend on the search step $s$. 
However, the gradient must be computed exactly, which is not feasible in large-scale applications. 
More recently, 
Tripuraneni \textit{et al.}~\cite{tripuraneni2018stochastic}  
introduced a stochastic cubic regularization method, but the downside of this method is that there is no adaptive way to adjust the regularization parameters.
Turning to the focus of this paper, 
Yao \textit{et al.}~\cite{yao2018inexact} recently applied a second-order method based on an inexact gradient based on STR and SARC. 
However, the cost of computing the value of the function was not considered. Further, each update of $\rho$, which measures the adequacy of the function, demands the objective function to be fully computed, which adds to the likely already-high computational cost.
{Bellavia \textit{et al.}~\cite{bellavia2022adaptive} 
successively consider the random noise in derivatives and inexact function values for any order for smooth unconstrained optimization problems and adaptive cubic regularization method with dynamic inexact Hessian information.
}
Other papers such as
\cite{cao2023first,curtis2018stochastic,sun2022trust} also consider the stochastic trust-region method from the noise setting.

Finally, we would like to emphasize that our proposed algorithms are empirically much faster than the stochastic TR and ARC. Although algorithmically it is straightforward to simultaneously compute the inexact Hessian matrix, gradient, and function values, it is challenging and not trivial to strictly prove the convergence. This is mainly because the approximate function does not guarantee the corresponding bounds of gradient and Hessian.


\section{Preliminary}
\label{Newton:Section:Preliminary} 
For a vector $x$ and a matrix $X$,  we use $\|x\|$ and $\|X\|$ to denote the Euclidean norm and the matrix spectral norm, respectively. We use $\mathcal{S}$ to denote the set and $|\mathcal{S}|$ to denote its cardinality. The set $\{1,2,...,n\}$ is defined as $[n]$. For the matrix $X$,  $\lambda_{\text{min}}(X)$ and $\lambda_{\text{max} }(X)$  denote its smallest and largest eigenvalue, respectively.   We use $\mathbb{I}[event]$ to denote the indicator function of a probabilistic event. Some important assumptions and definitions follow, which include the characteristics of the function, the approximate conditions, and related bounds \footnote{Note that Assumptions \ref{Newton:Assumption:function}-
	\ref{Newton:Assumption:Bound:Variance} follows from \cite{nesterov2013introductory}, \cite{kohler2017sub}, \cite{xu2017newton}, and \cite{NIPS2017_6829}, respectively.}, and the definition of optimality.

\begin{assumption}(Lipschitz continuous)\label{Newton:Assumption:function}
	For the function $f(x)$, we assume that ${\nabla ^2}f( x )$ and ${\nabla}f( x )$  are Lipschitz continuous satisfying 
	$
	\| \nabla ^2f( x ) - {\nabla ^2}f( y ) \| \le L_H\| x - y\|,
	\| \nabla 	f( x ) - {\nabla}f( y ) \| \le L_{\nabla f}\| x - y\|,\forall x,y \in {\mathbb{R}^d}.
	$
\end{assumption}

\begin{assumption}
	(Approximation)\label{Newton:Assumption:approximation} 
	For the function $f(x)$, the  approximate gradient $g(x)$ and the approximate Hessian matrix $B(x)$ satisfy
	\begin{align}
	\label{Newton:Assumption:approximation-gB}
	    \left\| {\nabla f(x) - g(x)} \right\| \le {\epsilon_g},
	    \left\| {{\nabla ^2}f(x) - B(x)} \right\| \le v_0{\epsilon_B},
	\end{align}	
	where ${\epsilon_g},{\epsilon_B} > 0$, $v_0\in(0,1]$ is a constant. The approximated function $h(x)$ at the  $k$-iteration satisfies \footnote{Note that Lemma \ref{Newton:Subsample:Lemma} in Section \ref{Newton:Section:subsampling} illustrates that the approximated $h(x)$ is related to $s_k$ whenever the approximated value is $h(x_k)$ or $h(x_k+s_k)$. Moreover, unlike \cite{kohler2017sub} where the estimated Hessian matrix is dependent on $||s_k||$, the subproblem-solver in Algorithms \ref{Newton:STR:Algorithm} and \ref{Newton:SARC:Algorithm} does not require the value of the function. The setting of $v_0$ is used  for presenting the brief proof.}
	\begin{align}\label{Newton:Assumption:approximation-h}
	|{f(x_k) - h(x_k)} | \le {\epsilon_h}{\|s_k\|^2},{\epsilon_h} > 0.
	\end{align}
\end{assumption}

\begin{assumption}(Bound)\label{Newton:Assumption:bound} 
	For $i\in[n]$, the bound assumptions are that the function $f_i(x)$ satisfies $| {{f_i}({x})} | \le {\kappa _f}$, $\left\| {\nabla {f_i}({x})} \right\| \le {\kappa _{\nabla f}}$, and $\left\| {{\nabla ^2}{f_i}({x})} \right\| \le {\kappa _H}$.
\end{assumption}

\begin{assumption} (Bound) \label{Newton:Assumption:Bound:Variance}
	We assume $H_1$ and $H_2$ are the upper bounds on the variance of  $\nabla f_i(x)$ and $\nabla^2f_i(x)$, $i\in[n]$, i.e.,
	\begin{align}\label{Newton:Assumption:Bound:Variance-eq1}
		\frac{1}{n}\sum\nolimits_{i = 1}^n {{{\left\| {\nabla {f_i}\left( x \right) - \nabla f\left( x \right)} \right\|}^2}}  \le& H_1^2,
		\\
		\label{Newton:Assumption:Bound:Variance-eq2}
		\frac{1}{n}\sum\nolimits_{i = 1}^n {{{\left\| {{\nabla ^2}{f_i}\left( x \right) - {\nabla 	^2}f\left( x \right)} \right\|}^2}}  \le& H_2^2.
	\end{align}
\end{assumption}

\begin{figure}
	\centering
	\includegraphics[width=0.45\textwidth]{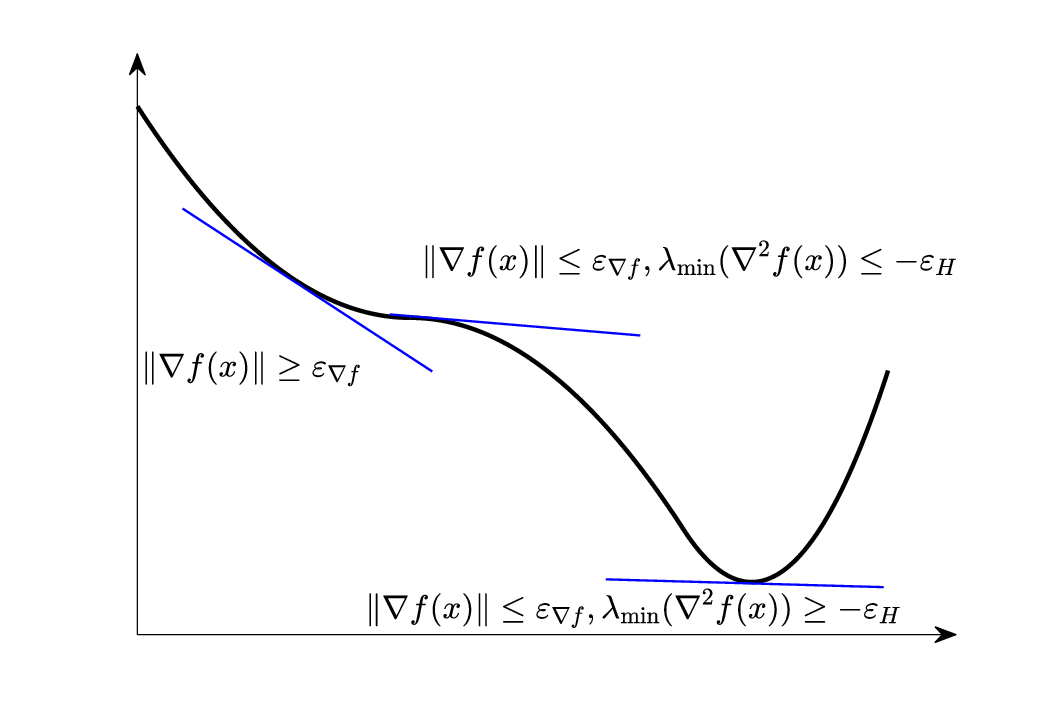}
	\caption{The three situations of convergence to the first and second critical points analyzed in this paper.}
	\label{Newton:Figure3Situation}
\end{figure}

\begin{definition}(($\epsilon_{{\nabla f}},\epsilon_H$)-optimality). 
	Given $\epsilon_{\nabla f}$ and ${\epsilon_H} \in [0,1]$, $x$ is an $( \epsilon_{{\nabla f}},{\epsilon_H})$-optimality solution to (\ref{Newton:Problem}) if
	\[	\| {\nabla f( {{x}} )} \| \le {\epsilon_{\nabla f}},\,\lambda_\text{min} ({\nabla ^2}f(x) ) \ge-{\epsilon_H }.\]
\end{definition}
This paper concerns the following three index sets:
\begin{align*}
\mathcal{S}_{{\nabla f}}&\mathop  = \limits^\text{def}\{x: \,  \|{\nabla 
	f( {{x}} )} \| \ge {\epsilon_{\nabla f}}\}, 
	\\
\mathcal{S}_{H}&\mathop  = \limits^\text{def}\{x: \|{\nabla 
	f( {{x}} )} \| \le {\epsilon_{\nabla f}} \,\text{and} 
\,\lambda_\text{min} 
({\nabla^2f(x )} ) \le - {\epsilon_H }\}, 
\\
\mathcal{S}_{*}&\mathop  = \limits^\text{def} \{x:\, \|
\nabla f( {{x}} ) \| \le \epsilon_{\nabla f}\,
\text{and}\, \lambda_\text{min} (\nabla^2f( x )) \ge  -\epsilon_H\}.	
\end{align*}
A simple geometric illustration is provided in Fig.  \ref{Newton:Figure3Situation} to illustrate these three situations.

The following lemma is a crucial tool for analyzing the convergence of the proposed algorithm. It characterizes the variance of the random variable, which decreases by a factor related to the set size.
\begin{lemma} \cite{9512549}
\label{Newton:Appendix:Tool:RandomSubset}
	If $v_1,...,v_n\in \mathbb{R}^d$ satisfy 
	$\sum\nolimits_{i = 1}^n {{v_i}}  = \vec 0$, and $\cal A$ is a non-empty, 	uniform random subset of $[n]$, $A=|\mathcal{A}|$, then
		${\mathbb{E}_{\mathcal{A}}} {{{\left\| {\frac{1}{A}\sum\nolimits_{b \in 
{\mathcal{A}}} {{v_b}} } \right\|}^2}}  \le \frac{{\mathbb{I}\left( {A < n} 		\right)}}{A}\frac{1}{n}\sum\nolimits_{i = 1}^n {\|v_i\|^2}.$
\end{lemma}

\section{Stochastic Second Optimization for Non-convex Problem}\label{Newton:Section:STR+SARC}
\subsection{Stochastic trust-region method}\label{Newton:Section:STR}
\begin{algorithm}[t]
	\caption{STR  with Inexact Hessian Matrix, Gradient, and Function}
	\label{Newton:STR:Algorithm}
	\begin{algorithmic}[1]
		\Require 
		given $x_0$, $r_2\ge 1> r_1$, $1>\eta>0$, $\epsilon>0$, ${\epsilon_{\nabla 	f}}$, ${\epsilon_g}$	and 	$\Delta_0,\Delta_\text{max}>0$. 
		\For{$k$=1 to T }
		\State Set the approximate gradient $g(x_k)$ and Hessian matrix $B(x_k)$ as in \eqref{Newton:Assumption:approximation-gB}.
		\State Compute the direction vector $s_k$
		\begin{align}\label{Newton:STR:Algorithm-solution}
		s_k=\mathtt{Subproblem-Solver}(g(x_k),B(x_k),\Delta_k).
		\end{align}	
		\If{$\|g\left( {{x_k}} \right)\| \le {\epsilon_{\nabla 	f}}+ {\epsilon_g}$}
		\State $\mathcal{S}_h=\mathcal{S}_g$
		\EndIf
		\State Set the approximate function $h(x_k)$ and $h(x_k+s_k)$ as in \eqref{Newton:Assumption:approximation-h}.
		\State 	Compute $
		{{\tilde \rho }_k} = \frac{{h\left( {{x_k}} \right) - h\left( {{x_k} + {s_k}} \right)}}{{{m_k}\left( 0 \right) - {m_k}\left( {{s_k}} \right)}}
		$
		and set $\hat\rho_k={{{\tilde \rho }_k} - \frac{{2{\epsilon_h}{\|s_k\|^2}}} 
			{{{m_k}\left( 0 \right) - {m_k}\left( {{s_k}} \right)}}}$						
		\State set ${x_{k + 1}} = \left\{ {\begin{array}{*{20}{l}}
			{{x_k} + {s_k},}&{\hat\rho_k  \ge \eta ,}\\
			{{x_k},}&{{\rm{otherwise}}{\rm{.}}}
			\end{array}} \right.$
		\State set
		${\Delta _{k + 1}} = \left\{ {\begin{array}{*{20}{l}}	{\text{min} \left\{ {{\Delta _{\text{max} }},{r_2}{\Delta _k}} \right\},}&{\hat\rho_k  > \eta 	,}\\{{r_1}{\Delta _k},}&{{\rm{otherwise}}.}	\end{array}} \right.$	
		\EndFor
	\end{algorithmic}
\end{algorithm}

In this section, we examine the STR method as a solution to constrained optimization problems. In the discussion that follows, we show that the iterative complexity of this method is of the same order as the classical TR but without exactly computing the function, gradient, and Hessian matrix. Rather, the objective function is approximated within a trust- region by a quadratic model at the $k^{th}$-iteration:
\begin{align}\label{Newton:STR:Objective}
\mathop {\min }\limits_{s\in \mathbb{R}^d} \,{m_k}( s ),\,\,\,\,	\text{subject to}\,\| s \| \le {\Delta _k},
\end{align}
where $\Delta_k$ is the radius of the trust-region,
${m_k}( s )$ is defined:
\begin{align}
\label{Newton:Definition_m}
    {m_k}(s)= h({{x_k}} ) 
    + \langle {{ g(x_k)},s} \rangle  
    + \frac{1}{2}\langle {s,{B(x_k)}s} \rangle,
\end{align}
in which the approximated function $h(x_k)$, gradient $g(x_k)$, and Hessian matrix $B(x_k)$ are formed based on the conditions in Assumption \ref{Newton:Assumption:approximation}. 
Note that, the problem in \eqref{Newton:STR:Objective} is defined as the subproblem, and the solution $s$ is obtained through $\mathtt{subproblem-solver()}$ in \cite{conn2000trust}.
The conditions for $\epsilon_B $ 
do not depend on the search direction $s$ the same as in \cite{xu2017newton,xu2017second}. Moreover, we define a new parameter $\epsilon_g$, which has the same characteristics as $\epsilon_B$.  Unlike \cite{xu2017newton} and \cite{kohler2017sub}, another modification is that we have replaced $f(x)$ and $f(x + s)$ with the approximate function $h(x)$ and $h(x + s)$ under the assumption \ref{Newton:Assumption:approximation}  and  implementation in \eqref{Newton:STR:Subsampling:h}-\eqref{Newton:STR:Subsampling:B}  to mitigate the expense of computing the finite-sum-structured $f(x)$ and $f(x + s)$. Condition (\ref{Newton:Assumption:approximation-h}) is subject to the search direction $s$. However, the approximate functions $h(x)$ and $h(x + s)$ can be derived after the solution for $s$ has been obtained. The process for updating $x$ and $\Delta$ is detailed in Algorithm \ref{Newton:STR:Algorithm}. The following theoretical analysis of the STR method consists of two parts: an analysis of the role of the radius $\Delta$ to ensure that it has a lower bound, and a calculation of the iteration complexity of the method under assumptions presented in Preliminary. 

There are three important definitions in this section: $\rho_k$, $\tilde{\rho}_k$, and $\hat\rho_k$, i.e., the original ratio, the approximated ratio, and the updated and approximated ratio. The two terms $\rho_k$ and $\tilde{\rho}$  are defined as
\begin{align*}
    {\rho _k} = \frac{{f\left( {{x_k}} \right) - f\left( {{x_k} + {s_k}} \right)}}{{{m_k}\left( 0 	\right) - {m_k}\left( {{s_k}} \right)}},{{\tilde \rho }_k} = \frac{{h\left( {{x_k}} \right) - h\left( {{x_k} + {s_k}} \right)}}{{{m_k}\left( 0 \right) - {m_k}\left( {{s_k}} \right)}}.
\end{align*}
Based on Eq. \eqref{Newton:Assumption:approximation-h} in Assumption 
\ref{Newton:Assumption:approximation}, we have
\begin{align*}
\left| {{\rho _k} - {{\tilde \rho }_k}} \right|
=&
\left| {\frac{{f\left( {{x_k}} \right) - h\left( {{x_k}} \right) - \left( {f\left( {{x_k} + {s_k}} \right) - h\left( {{x_k} 	+{s_k}}	\right)} \right)}}{{{m_k}\left( 0 \right) - {m_k}\left( {{s_k}} \right)}}} \right|\\
\le&
\frac{{2{\epsilon_h}\|s_k\|^2}}{{{m_k}\left( 0 \right) - {m_k}\left( {{s_k}} \right)}},
\end{align*}
i.e.,
${{\tilde \rho }_k}- \frac{{2{\epsilon_h}\|s_k\|^2}}{{{m_k}\left( 0 \right) - {m_k}\left( {{s_k}} \right)}} \le {\rho _k}  \le \frac{{2{\epsilon_h}{\|s_k\|^2}}}{{{m_k}\left( 0 \right) - {m_k}\left( {{s_k}} \right)}}+ {{\tilde \rho }_k}.
$
Then,  $\hat\rho_k$ can be defined as ${{\tilde \rho }_k} - \frac{{2{\epsilon _h}{\|s_k\|^2}}}{{{m_k}\left( 0 \right) - 
{m_k}\left( {{s_k}} \right)}} $.  If $\hat\rho \ge {\eta}$, we can derive ${\rho _k} \ge {\eta}$. Thus, in the following analysis, we consider the size of $\rho$ that derives the desired lower bound of the radius. With this established, the following analysis examines the size of $\hat\rho_k$, which is used to derive the desired lower bound of the radius. However, before presenting the formal analyses, we must briefly discuss why the radius  $\Delta$ does not approximate to zero.

If ${\hat\rho_k} > {\eta }$, the current iteration will be successful, and the radius $\Delta$ will increase by a factor of $r_2$. Therefore, it is worth considering whether there is a constant $C$ such that $\Delta<C$ and the current iteration can be successful simultaneously. If so, then constant $C$ would be the desired bound of $\Delta$, because $\Delta$ would increase again with successful iteration. Moreover, this constant would play a critical role in determining the iteration complexity. Therefore, instead of computing $\hat\rho$ directly, we consider another relationship:
\begin{align}\label{Newton:STR:1-p}
&
1 - \hat\rho_k
\nonumber\\  
= 
&
\frac{{{m_k}\left( 0 \right) - {m_k}\left( {{s_k}} \right) - \left( {h\left( {{x_k}} \right) - h\left( {{x_k} + {s_k}} \right)} \right) + 2{\epsilon_h}{\|s_k\|^2}}}{{{m_k}\left( 0 \right) - {m_k}\left( {{s_k}} \right)}}.
\end{align}
Here, as  long as $1-\hat\rho_k<1-\eta$, we can see that $\hat\rho_k>\eta$. 

Additionally, the solution $s_k$ for the subproblem-solver is based on the subproblem in (\ref{Newton:Definition_m}). In cases where $x_k\in\mathcal{S}_{\nabla f}$, we use the Cauchy point \cite{conn2000trust} {\color{blue}\footnote{Cauchy point is the minimizer of $m_k(s)$ within the trust region}}, while in cases where $x_k\in\mathcal{S}_{H}$, there are many methods to detect the negative curvature, such as shift-and-invert  \cite{garber2016faster}, lanczos \cite{kuczynski1992estimating}, and negative-curvature \cite{carmon2018accelerated}. The details of these techniques are beyond the scope of this paper, but we encourage interested readers to review the source papers.
The following analysis includes the complexity of both successful and unsuccessful iterations, including $x\in \mathcal{S}_{\nabla f}$ and $x\in \mathcal{S}_H$ followed by the total number of iterations.


\begin{theorem}\label{Newton:STR:Theorem:Iteration}
	In Algorithm \ref{Newton:STR:Algorithm}, suppose the Assumptions \ref{Newton:Assumption:function}-\ref{Newton:Assumption:Bound:Variance} hold, let $| \mathcal{S}_h | = \text{min} 	\{ n,\text{max} \{ H_1/\epsilon_g,H_2/\epsilon_B \} \}$, $\{ f( x ) \}$ be bounded below by $f_\text{low}$. Thus, the number of successful iterations ${T_\text{suc}}$  can be	no larger than
	${\kappa _3}\text{max}\{( \epsilon_H - \epsilon_B )^{ - 1} ( \epsilon_{\nabla f} - \epsilon_g )^{ - 2},(\epsilon_H - \epsilon_B)^{ - 3} \}$,
	where $\kappa _3 = 2( f( x_0 ) - f_\text{low} )\text{max}\{ 1/( \eta {\kappa _1} ),1/( \eta \kappa _2^2 ) \}$, 
	$\kappa_1$ 	and $\kappa_2$ are two positive constants.  
	Further, the number of unsuccessful iterations $T_\text{unsuc}$ can, at most, be $\frac{1}{-\log {r_1}}( \log ( \Delta_{\text{max}}/\Delta _\text{min} ) - T\log {r_2} ),$	where $\Delta _\text{max}$ and $\Delta _\text{min} $ are two positive constants,	$1>{r}_1>0$, and $r_2\ge 1$. Thus, the total number of iterations is 
	$\mathcal{O}( \text{max} \{ ( \epsilon_H - \epsilon_B )^{ - 1}(\epsilon_{\nabla f} - {\epsilon_g})^{ - 2},(\epsilon_H - \epsilon_B)^{ - 3} \} )$.
\end{theorem}

After  $|T_\text{suc}|+|T_\text{unsuc}|$ iterations, Algorithm \ref{Newton:STR:Algorithm}  will fall into $\mathcal{S}_*$ and converge to a stationary point. 
Two conclusions can be drawn from Theorem \ref{Newton:STR:Theorem:Iteration}. First, if the parameters $\epsilon_g$ and $\epsilon_B$  are set properly according to  $\epsilon_{\nabla f}$ and $\epsilon_H$, respectively, the iteration complexity is of the same order as \cite{cartis2011adaptiveb} and \cite{xu2017second}. Second, when $| {{\mathcal{S}_h}} |<n$, there are fewer total iterations than that of \cite{cartis2011adaptiveb} and \cite{xu2017second}, including those for computing the function, and, when $| {{\mathcal{S}_h}} |=n$, there are an equal number of iterations. From these results, we find that, overall, our algorithm is a more general solution.


\subsection{Stochastic adaptive regularization using cubics}
\label{Newton:Section:SARC}
\begin{algorithm}[t]
	\caption{SARC  with the Inexact Hessian Matrix, Gradient, and Function}
	\label{Newton:SARC:Algorithm}
	\begin{algorithmic}[1]
		\Require given $x_0$, $r_2\ge 1> r_1$, $1>\eta>0$, $\epsilon>0$ and  
		$\sigma_\text{min}>0$.
		\Ensure
		\For{$k$=1 to T }
		\State Set the approximate gradient $g(x_k)$ and Hessian matrix $B(x_k)$ as in \eqref{Newton:Assumption:approximation-gB}.
		\State Compute the direction vector $s_k$
		\begin{align}\label{Newton:SARC:Algorithm-solution}
		s_k=\mathtt{Subproblem-Solver}(g(x_k),B(x_k),\sigma_k)
		\end{align}	
		\If{$\|g\left( {{x_k}} \right)\| \le {\epsilon_{\nabla 	f}}+ {\epsilon_g}$}
		\State $\mathcal{S}_h=\mathcal{S}_g$
		\EndIf
		\State Set the approximate function $h(x_k)$ and $h(x_k+s_k)$ as in \eqref{Newton:Assumption:approximation-h}.	
		\State 	Compute $
		{{\tilde \rho }_k} = \frac{{h\left( {{x_k}} \right) - h\left( {{x_k} + {s_k}} \right)}}{{{p_k}\left( 0 \right) - {p_k}\left( {{s_k}} \right)}}
		$
		\State Set $\rho={{{\tilde \rho }_k} - \frac{{2{\epsilon_h}/{\sigma^ 2_k}}}{{{p_k}\left( 0 	\right) - {p_k}\left( {{s_k}} \right)}}}$			
		\State set ${x_{k + 1}} = \left\{ {\begin{array}{*{20}{l}}	{{x_k} + {s_k},}&{\rho  \ge \eta ,}\\
		{{x_k},}&{{\rm{otherwise}}{\rm{.}}}
			\end{array}} \right.$
		\State set
		${\sigma _{k + 1}} = \left\{ {\begin{array}{*{20}{l}}	{\max \left\{ {{\sigma _{\text{min} }},{r_1}{\sigma _k}} \right\},}&{\rho  > \eta 	,}\\{{r_2}{\sigma _k},}&{{\rm{otherwise}}.}
			\end{array}} \right.$	
		\EndFor
	\end{algorithmic}
\end{algorithm}

This subsection describes the stochastic adaptive cubic regularization (SARC) method of stochastic adaptive regularization. Our analysis shows that our proposed SARC has an iteration complexity of the same order as ARC but without exactly computing the function, gradient, and Hessian matrix. SARC solves the following unconstrained minimization problem in each iteration:
\begin{align}\label{Newton:SARC:definition_P}
    \mathop {\min }\nolimits_{s\in \mathbb{R}^d}  p_k( s ) := m_k(s) + \frac{{{\sigma _k}}}{3}{\| s \|^3},
\end{align}
where $m_k(s)$ is defined in (\ref{Newton:Definition_m}), and $\sigma_k$ is an adaptive parameter that can be considered as the reciprocal of the trust-region radius. The process for updating  $x_k$ and $\sigma_k$ is provided in Algorithm \ref{Newton:SARC:Algorithm}. 
 Note that, the problem in \eqref{Newton:SARC:Algorithm-solution} is defined as the subproblem, and the solution $s$ is obtained through $\mathtt{subproblem-solver()}$ in \cite{conn2000trust}.
Similar to 
the analysis in \cite{cartis2011adaptivea}, $\sigma_k$ in the cubic term actually perform one more task in addition to accounting for the discrepancy between the objective function and its corresponding second-order Taylor expansion. It also accounts for the difference between the exact and the approximate functions, gradients, and Hessian matrices. The update rules for  $\sigma$ are analogous to the stochastic region method. However, $\sigma$ will decrease if there is a sufficient decrease in some measure of the relative objective chance; otherwise, it will increase. Similar to STR, we present the definition of $\hat\rho_k$ directly:
\begin{align}\label{Newton:SARC:1-p}
&
1 - \hat\rho_k
\nonumber	\\  
=& 
\frac{{{p_k}( 0 ) - {p_k}( {{s_k}} ) - ( {h( {{x_k}} ) - h( {{x_k} + {s_k}} )} ) + 2{\epsilon_h}\|s_k\|^2}} {{{p_k}( 0 ) - {p_k}( {{s_k}} )}}.
\end{align}

The corresponding lower and upper bounds of the numerator and denominator to satisfy the inequality 
are provided 
in the Appendix
(Lemma \ref{Newton:SARC:Lemma:UpperboundOfP0Ps}-Lemma 
\ref{Newton:SARC:Lemma:Upperbound_p-h}).
Moreover, to widen the scope of convergence analysis and iteration complexity, we  also consider the step size $s_k$ following a similar subspace analysis of the cubic model outlined in \cite{cartis2011adaptivea} and \cite{cartis2011adaptiveb} under the following conditions:
\begin{align}
\label{Newton:SARC:Assumption_s1}
&\left\langle {g\left( {{x_k}} \right),{s_k}} \right\rangle  + s_k^T{B_k}{s_k} + {\sigma _k}{\left\| {{s_k}} \right\|^3} = 0,\nonumber\\
&s_k^T{B_k}{s_k} + {\sigma _k}{\left\| {{s_k}} \right\|^3} \ge 0,\\
\label{Newton:SARC:Assumption_s3}
&\| \nabla {p_k}( s_k ) \| \le {\theta _k}\|  g( x_k ) \|,\nonumber\\
&{\theta _k} \le {\kappa _\theta }\text{min}\{ {1,\left\| s_k \right\|} \},{\kappa _\theta } < 1.
\end{align}


Based on the above discussion, we present the iteration complexity. Unlike STR, our analysis of SARC involves two kinds of complexity – the difference being whether or not the conditions in (\ref{Newton:SARC:Assumption_s3}) are satisfied. If the conditions are not satisfied, the iteration complexity to convergence on the stationary point is of the same order as STR. However, if the conditions are satisfied, the iteration complexity improves to better than or equal to STR.

\begin{theorem}\label{Newton:SARC:theorem:Iteration}
	In Algorithm \ref{Newton:SARC:Algorithm}, suppose the Assumptions \ref{Newton:Assumption:function}-\ref{Newton:Assumption:Bound:Variance} hold. Let $| {\mathcal{S}_h} | = \text{min}\{ n,\text{max} \{ H_1/\epsilon_g,H_2/\epsilon_B \} \}$, $\{ f( x_k ) \}$ be 	bounded below by $f_{\text{low}}$. Thus, the number of  iterations ${T}$  can be no larger than
	$\mathcal{O}(\text{max} \{ (\epsilon_{\nabla f} - \epsilon_g)^{ - 2},( 
	\epsilon_H - \epsilon_B )^{ - 3} \})$.
	If  conditions  (\ref{Newton:SARC:Assumption_s1})-(\ref{Newton:SARC:Assumption_s3}) are satisfied, then the number of 	iterations 	$T$ will, at most, be 
	$\mathcal{O}(\text{max} \{ (\epsilon_{\nabla f} - \epsilon_g)^{ - 3/2},( 	\epsilon_H- \epsilon_B )^{ - 3} \})$.
\end{theorem}
As shown by these results, our proposed method has the same order of iteration complexity as  \cite{xu2017newton} and \cite{xu2017second}. However, our algorithm does not require full computation of the function and gradient and, thus, has a lower computational cost.

\section{Sub-sampling for finite-sum minimization}\label{Newton:Section:subsampling}
Given the finite-sum problem \eqref{Newton:Problem}, $f(x)$, $\nabla f(x)$, and $\nabla^2f(x)$ can be estimated by random subsampling, which, in turn, can drastically reduce the computational complexity. Here,  $\mathcal{S}_h$, $\mathcal{S}_g$, and $\mathcal{S}_B$  denote the random sample collections for estimating $f(x)$, $\nabla f(x)$, and $\nabla^2f(x)$, respectively, where $\mathcal{S}_h$, $\mathcal{S}_g$,  and $\mathcal{S}_B\subseteq [n]$. The approximated ones are formed by
\begin{align}
    \label{Newton:STR:Subsampling:h}
    h(x)=&\frac{1}{\left| {{\mathcal{S}_h}} \right|}\sum\limits_{i \in {\mathcal{S}_h}} {{f_i}\left( x \right)},\\
    \label{Newton:STR:Subsampling:g}
    g( {{x}} ) =& \frac{1}{{\left| {{\mathcal{S}_g}} \right|}}\sum\limits_{i \in {\mathcal{S}_g}}^{} {\nabla {f_i}\left( {{x}} \right)},\\
    \label{Newton:STR:Subsampling:B}
    B( {{x}} ) =& \frac{1}{{| {{\mathcal{S}_B}} |}}\sum\limits_{i \in {\mathcal{S}_B}}^{} {{\nabla ^2}{f_i}( {{x}} )}.
\end{align}

The simultaneous approximation of the Hessian matrix, gradient, and function values brings difficulty in proving the convergence, e.g., the adjustment of the parameter $\rho$ is inexplicable. Because the gradient and Hessian matrix of the approximated function $h(x)$ do not ensure error bounds, i.e. if $| h( x ) - f( x ) | \le \varepsilon$, we can guarantee neither $\| \nabla h( x ) - \nabla f( x ) \| \le \varepsilon$ nor $\| \nabla ^2h( x ) - \nabla ^2f( x ) \| \le \varepsilon$. However, the sampled $h(x)$ has the characteristics bound as in Lemma \ref{Newton:Appendix:Tool:RandomSubset}, which can avoid the dilemma to obtain the corresponding error bounds as shown in Lemma \ref{Newton:Subsample:Lemma}, i.e., $\left\| {\nabla h\left( x \right) - \nabla f\left( x \right)} \right\| \le \frac{{\mathbb{I}\left( {\left| {{S_h}} \right| < n} \right)}}{{\left| {{S_h}} \right|}}{H_1}$ and $\left\| {\nabla^2 h\left( x \right) - \nabla^2 f\left( x \right)} \right\| \le \frac{{\mathbb{I}\left( {\left| {{S_h}} \right| < n} \right)}}{{\left| {{S_h}} \right|}}{H_2}$. 
Furthermore, considering the probability-based models \cite{blanchet2019convergence,gratton2018complexity,chen2018stochastic}, the difference lies in 1) we give the proof of our proposed method from the stochastic sampling view rather than using the martingale process such that the proof process of ours is more intuitive than that of the martingale based method; 2) Our assumptions regarding the $\|\nabla f(x)-g(x)\|\le \epsilon_g$ and   $\|\nabla^2 f(x)-B(x)\|\le \epsilon_B$  do not depend on the stepsize or the ball $B(x, \delta)$. Note that the function approximate makes use of stepsize after the stepsize is obtained, which is reasonable.

\subsection{Randomized sub-sampling}

Most approaches use the operator-Bernstein inequality to probabilistically guarantee these approximations. For example, an approximate matrix multiplication is used to result in a fundamental primitive in RandNLA, which controls the approximation error of $\nabla f(x)$ \cite{drineas2006fast,mahoney2011randomized}. Alternatively, a vector-Bernstein inequality \cite{candes2011probabilistic,gross2011recovering} is applied to determine the subsample bound of the gradient \cite{ kohler2017sub}.  Based on the above techniques, we derive Lemma \ref{Newton:Subsample:Lemma_1}, which gives the sample size required of the inexact gradient $g(x)$, and inexact Hessian $B(x)$.

\begin{lemma} \label{Newton:Subsample:Lemma_1}
	If $| {{\mathcal{S}_g}} | \ge 16\log ( \frac{2d}{\delta_0}){{L_f^2}}/{{\epsilon_g^2}}$, then $g(x)$ formed by 	(\ref{Newton:STR:Subsampling:g})  satisfies $\| \nabla f( x ) - g(x) \| \le {\epsilon}_{g}$ with a probability 	of $1-\delta_0$.
	If $| \mathcal{S}_B | \ge \log ( \frac{2d}{\delta_0 } ) {{16L_B^2}}/(v^2_0\epsilon_B^2)$, then  $B(x)$ formed by (\ref{Newton:STR:Subsampling:B})	satisfies $\| {\nabla ^2}f( x ) -{B(x)}\| \le 	v_0{\epsilon}_B$ with a probability of	$1-\delta_0$.
\end{lemma}

To reduce the computational cost for updating the ratio $\rho$, the approximate function is derived by subsampling combined with the operator-Bernstein inequality. The resulting function probabilistically satisfies (\ref{Newton:Assumption:approximation-h}) in the following lemma. However, it does not guarantee the corresponding bounds of $\left\| {\nabla h\left( x \right) - \nabla f\left( x \right)} \right\|$ and $\| {{\nabla ^2}h( x ) - {\nabla ^2}f( x )} \|$. Thus, to derive the upper bound, we also make use of the finite-sum structure and Lemma \ref{Newton:Appendix:Tool:RandomSubset} to obtain an important step, which is used to analyze the update of ratio $\rho$.
\begin{lemma}\label{Newton:Subsample:Lemma} 
	If 	 $|{\cal S}_h| \ge 16\kappa _f^2\log (\frac{2d}{\delta_0})/( \epsilon_h^2  \Delta_c^4 )$, where $\Delta_c>0$ is a constant, upper bounded by $\text{max}\{{(\Delta_\text{max})^2},(1/\sigma_\text{min})^2\}$, then $h(x)$ formed by (\ref{Newton:STR:Subsampling:h}) satisfies $| f(x ) - h(x)| \le {\epsilon}{_h}\|s_k\|^2$ with a probability of $1-\delta_0$. Furthermore, suppose  Assumptions \ref{Newton:Assumption:Bound:Variance} holds,  we can also have the upper bounds of the gradient and the Hessian matrix for $h(x)$:
	    ${\left\| {\nabla h\left( x \right) - \nabla f\left( x \right)} \right\|^2} 
	    \le
	    \frac{{\mathbb{I}\left( {\left| {{\mathcal{S}_h}} \right| < n} \right)}}{{\left| 	{{\mathcal{S}_h}} \right|}}{H_1}$,
	    ${\left\| {{\nabla ^2}h\left( x \right) - {\nabla ^2}f\left( x \right)} \right\|^2} 
	    \le
	    \frac{{\mathbb{I}\left( {\left| {{\mathcal{S}_h}} \right| < n} \right)}}{{\left| {{\mathcal{S}_h}} \right|}}{H_2}$.
	    where $H_1$ and $H_2$ are defined in \eqref{Newton:Assumption:Bound:Variance-eq1} and \eqref{Newton:Assumption:Bound:Variance-eq2}.
\end{lemma}

\subsection{Probabilistic convergence analysis}
Here, we give the probabilistic iteration complexity of Algorithm \ref{Newton:STR:Algorithm} and Algorithm \ref{Newton:SARC:Algorithm} under inexact function, gradient, and Hessian matrix according to Lemma \ref{Newton:Subsample:Lemma_1} and Lemma \ref{Newton:Subsample:Lemma}. 
Thus, the approximations are the probabilistic construction, in order to ensure the success iterations, we require a small failure probability. Specifically, in order to get an accumulative success probability of $1-\delta$ for the whole $T$ iterations, the \textit{par-iteration failure probability} is set as  $\delta_0=(1-\sqrt[T]{(1-\delta)})\in \mathcal{O}(\delta/T)$.\footnote{
Assume that the par-iteration failure probability is $0<p<1$, we have $(1-\sqrt[T]{(1-\delta)})=p
    \Leftrightarrow (1-p)^T=1-\delta$. Based on approximation limit and the Taylor expansion, we have $(1-p)^T\approx e^{-pT}\approx 1-pT \Rightarrow p\approx \frac{\delta}{T}$.
}
This failure probability appears only in the “log factor” for sample size in all of our results, and so it is not the dominating cost, e.g.  $| {{\mathcal{S}_g}} | \ge 16\log ( \frac{2d}{\delta_0}){{L_f^2}}/{{\epsilon_g^2}}$ in Lemma \ref{Newton:Subsample:Lemma_1}. Hence, requiring that all $T$ iterations are successful for a large $T$, only need a small (logarithmic) increase in the sample size. 
For example, for $T\in\mathcal{O}( \text{max} \{ ( \epsilon_H - \epsilon_B )^{-1}(\epsilon_{\nabla f} - {\epsilon_g})^{-2},(\epsilon_H - \epsilon_B)^{-3} \} )$, as in Theorem \ref{Newton:STR:Theorem:Iteration}, we can set the per-iteration failure probability to $\delta\mathcal{O}( \text{max} \{(\epsilon_H - \epsilon_B )(\epsilon_{\nabla f}-{\epsilon_g})^2,(\epsilon_H-\epsilon_B)^3\})$, 
and ensure that when Algorithm \ref{Newton:STR:Algorithm} terminates, all Hessian  approximations have been, accumulatively, successful with probability of $1-\delta$. Thus, we derive the probabilistic iteration complexity for solving the finite-sum problem in (\ref{Newton:Problem}) for both STR and SARC. Since the proof is similar to Theorem \ref{Newton:STR:Theorem:Iteration} and Theorem  \ref{Newton:SARC:theorem:Iteration}, the proofs are omitted.
\begin{corollary}\label{Newton:STR:corollary:Iteration}
	In Algorithm \ref{Newton:STR:Algorithm}, 
	let the sample size of $ \mathcal{S}_h $, $\mathcal{S}_g$, and $\mathcal{S}_B$ be as in Lemma \ref{Newton:Subsample:Lemma_1} and \ref{Newton:Subsample:Lemma}, where the per-iteration failure probability is $\delta\mathcal{O}( \text{max} \{ ( \epsilon_H - \epsilon_B )(\epsilon_{\nabla f} - {\epsilon_g})^{2},(\epsilon_H - \epsilon_B)^{3} \} )$. Suppose the Assumptions \ref{Newton:Assumption:function}, \ref{Newton:Assumption:bound}, and \ref{Newton:Assumption:Bound:Variance} hold,  the total number of iterations is 
	$\mathcal{O}( \text{max} \{ ( \epsilon_H - \epsilon_B )^{ - 1}(\epsilon_{\nabla f} - {\epsilon_g})^{ - 2},(\epsilon_H - \epsilon_B)^{ - 3} \} )$  with probability of $1-\delta$.
\end{corollary}
\begin{corollary}
\label{Newton:SARC:corollary:Iteration}
	In Algorithm \ref{Newton:SARC:Algorithm}, suppose Assumptions \ref{Newton:Assumption:function}, \ref{Newton:Assumption:bound}, and \ref{Newton:Assumption:Bound:Variance} hold, 
	let the sample size of $ \mathcal{S}_h $, $\mathcal{S}_g$, and $\mathcal{S}_B$, be as in Lemma \ref{Newton:Subsample:Lemma_1} and \ref{Newton:Subsample:Lemma}, where the per-iteration failure probability is $\delta\mathcal{O}(\text{max} \{ (\epsilon_{\nabla f} - \epsilon_g)^{2},( 
	\epsilon_H - \epsilon_B )^{3} \})$ (or $\delta\mathcal{O}(\text{max} \{ (\epsilon_{\nabla f} - \epsilon_g)^{3/2},( \epsilon_H- \epsilon_B )^{3} \})$). Thus, the number of  iterations ${T}$  can be no larger than
	$\mathcal{O}(\text{max} \{ (\epsilon_{\nabla f} - \epsilon_g)^{ - 2},( 
	\epsilon_H - \epsilon_B )^{ - 3} \})$ with probability of $1-\delta$.
	If  conditions  (\ref{Newton:SARC:Assumption_s1})-(\ref{Newton:SARC:Assumption_s3}) are satisfied, then the number of 	iterations 	$T$ will, at most, be 
	$\mathcal{O}(\text{max} \{ (\epsilon_{\nabla f} - \epsilon_g)^{ - 3/2},( \epsilon_H- \epsilon_B )^{ - 3} \})$ with probability of $1-\delta$.
\end{corollary}

\section{Experiments}
\label{Newton:Section:Experiment}
In this section, we empirically evaluate the performance of the proposed algorithms, in which we consider two kinds of optimization problems arising in practice, i.e., non-linear least square and deep learning.

\begin{figure*}[t]
 	\centering
 	\subfigure{
 		\begin{minipage}[b]{0.3\textwidth}
 		\includegraphics[width=1.0\textwidth]{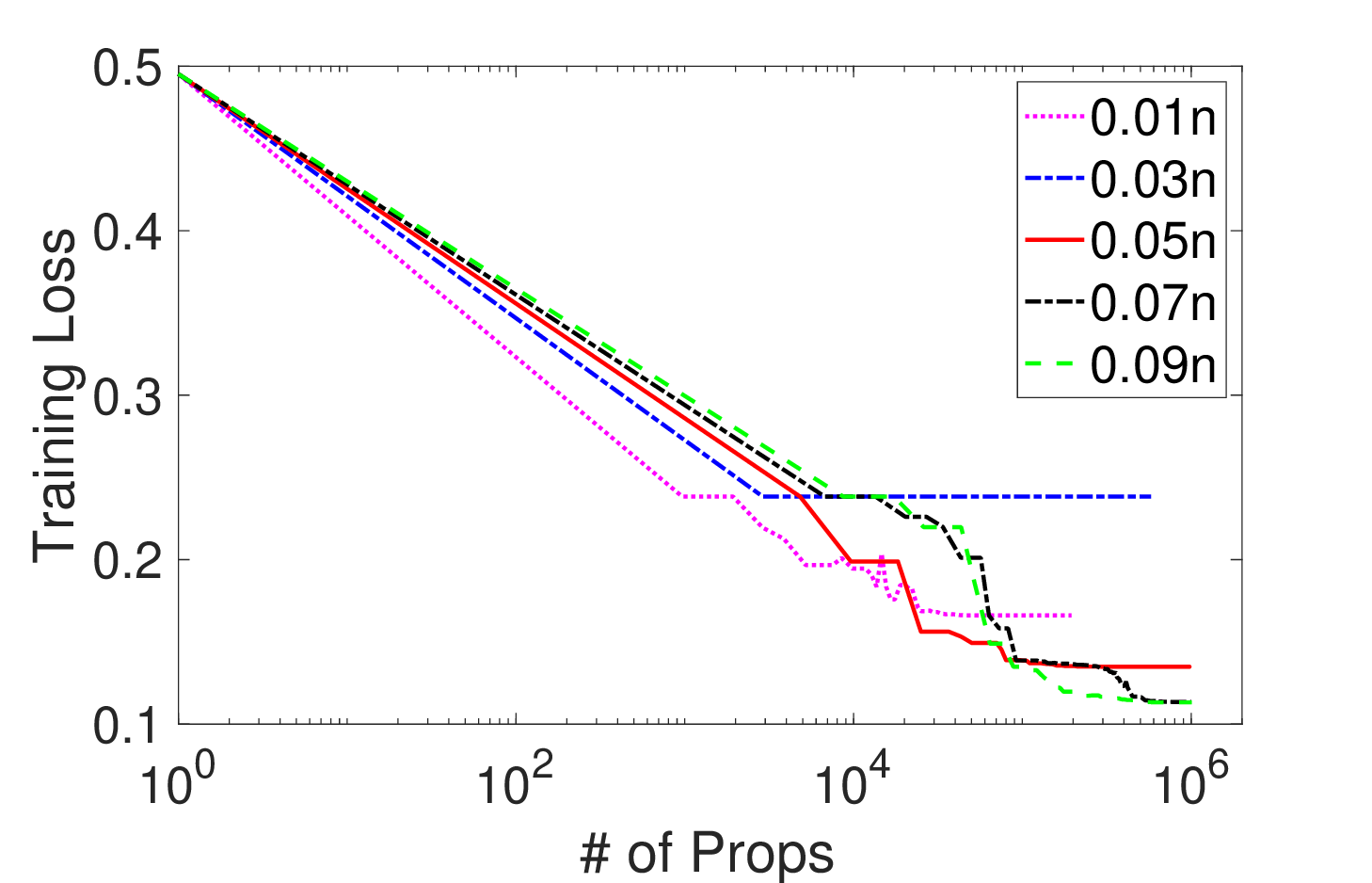}
 		\\
 		\includegraphics[width=1.0\textwidth]{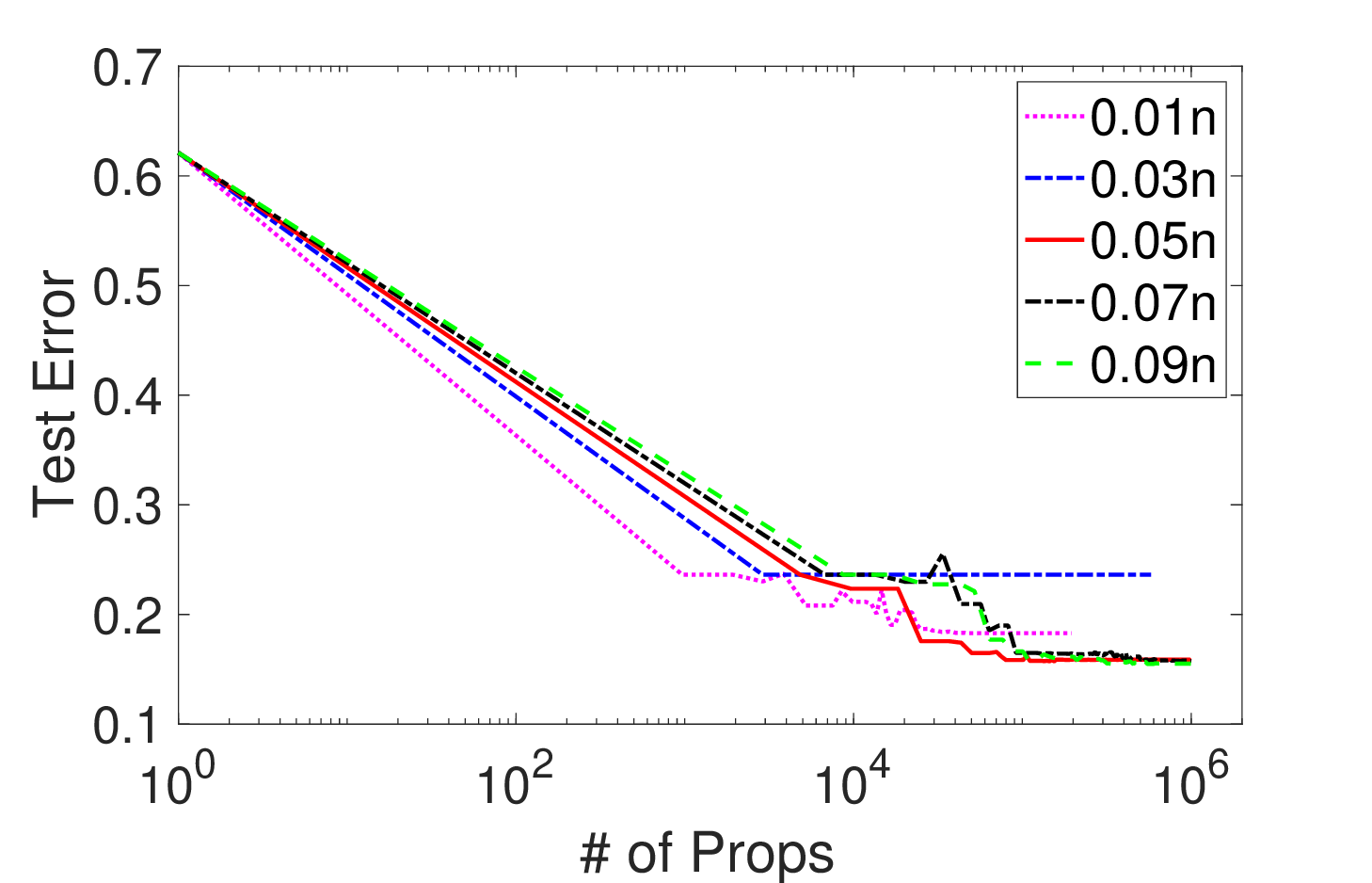}
 		\end{minipage}
 	}
 	\subfigure{
 		\begin{minipage}[b]{0.3\textwidth}
 		\includegraphics[width=1.0\textwidth]{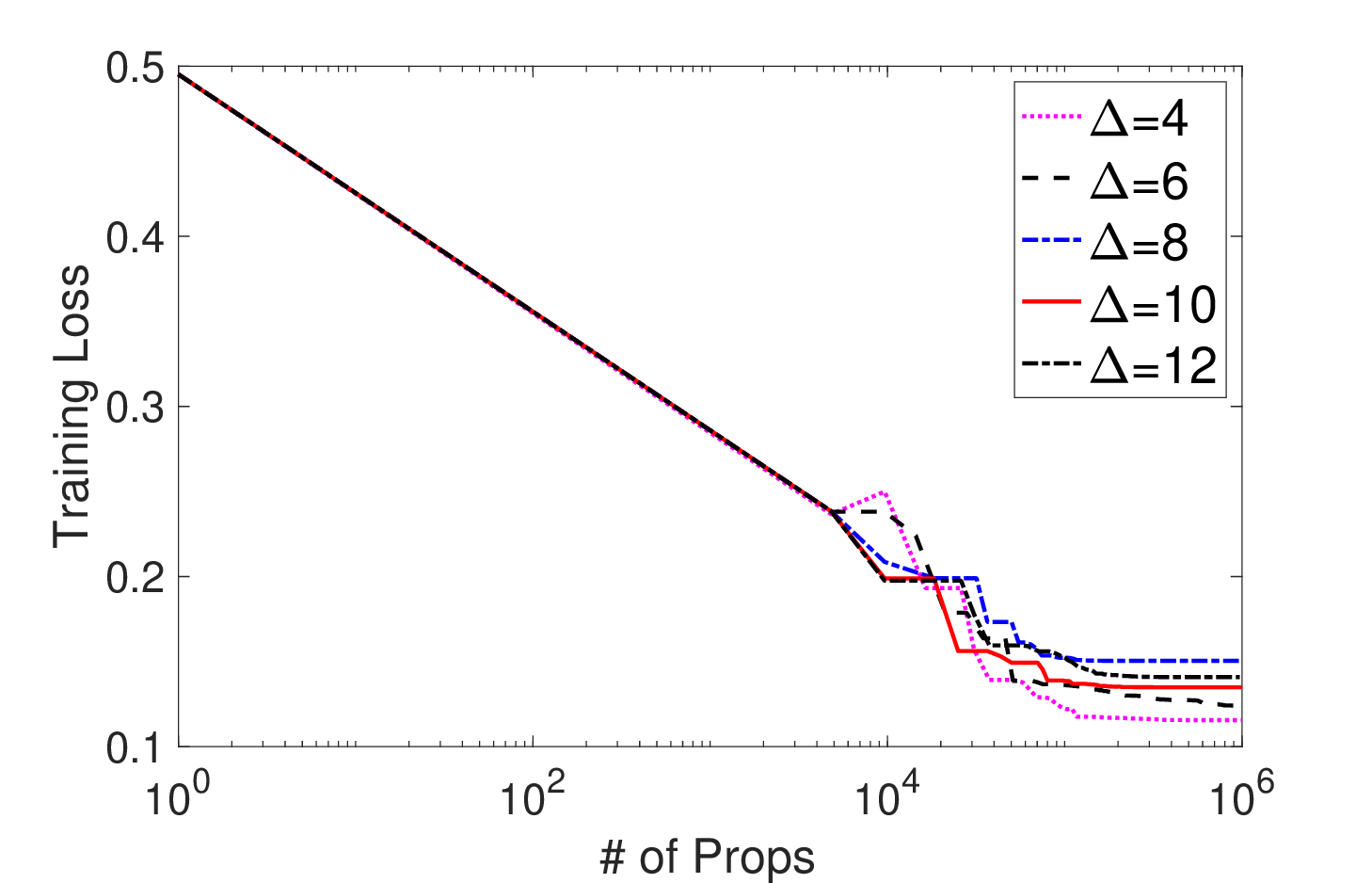}
 		\\
 		\includegraphics[width=1.0\textwidth]{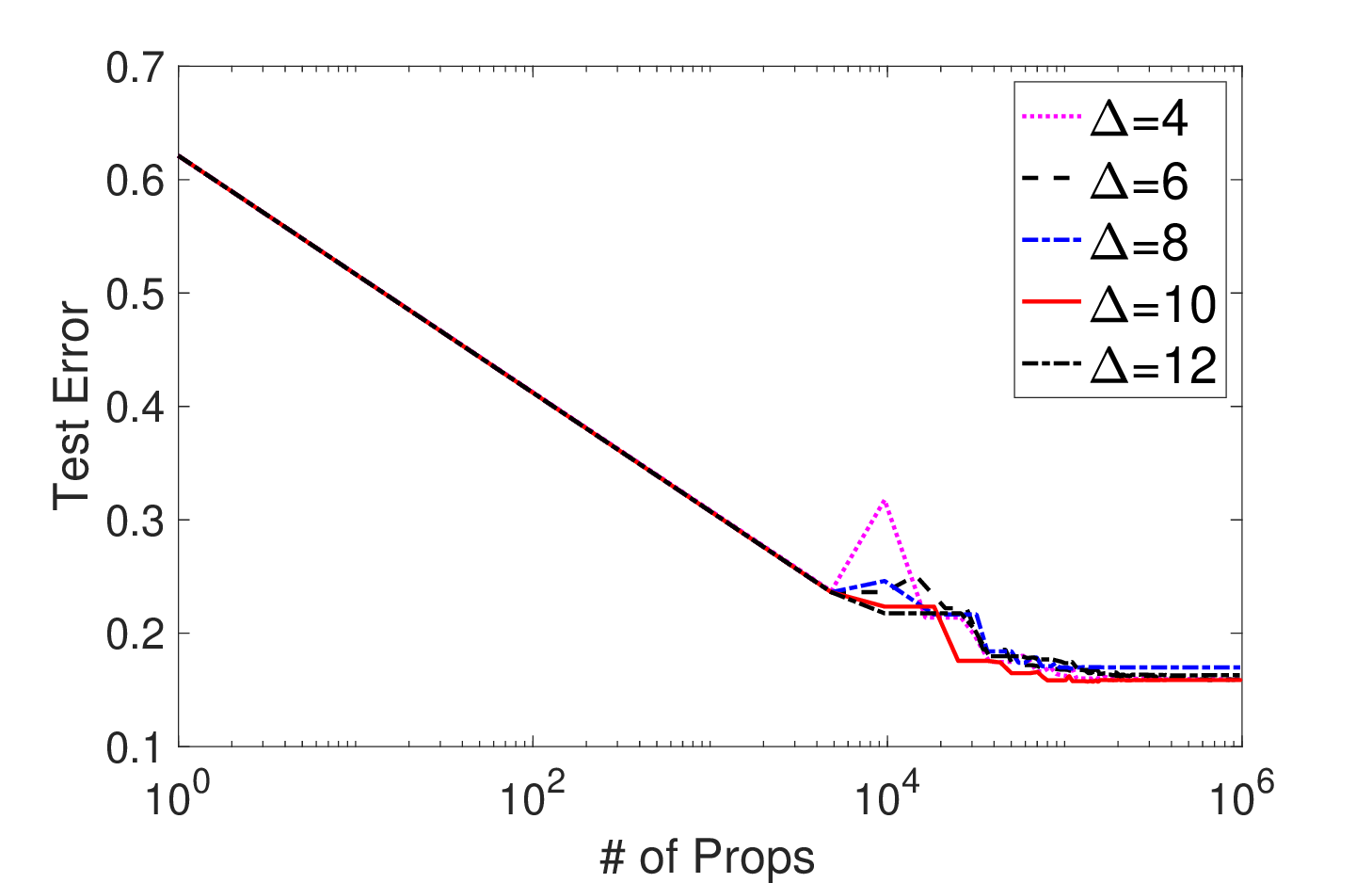}
 		\end{minipage}
 	}
 	\subfigure{
 		\begin{minipage}[b]{0.3\textwidth} 			\includegraphics[width=1.0\textwidth]{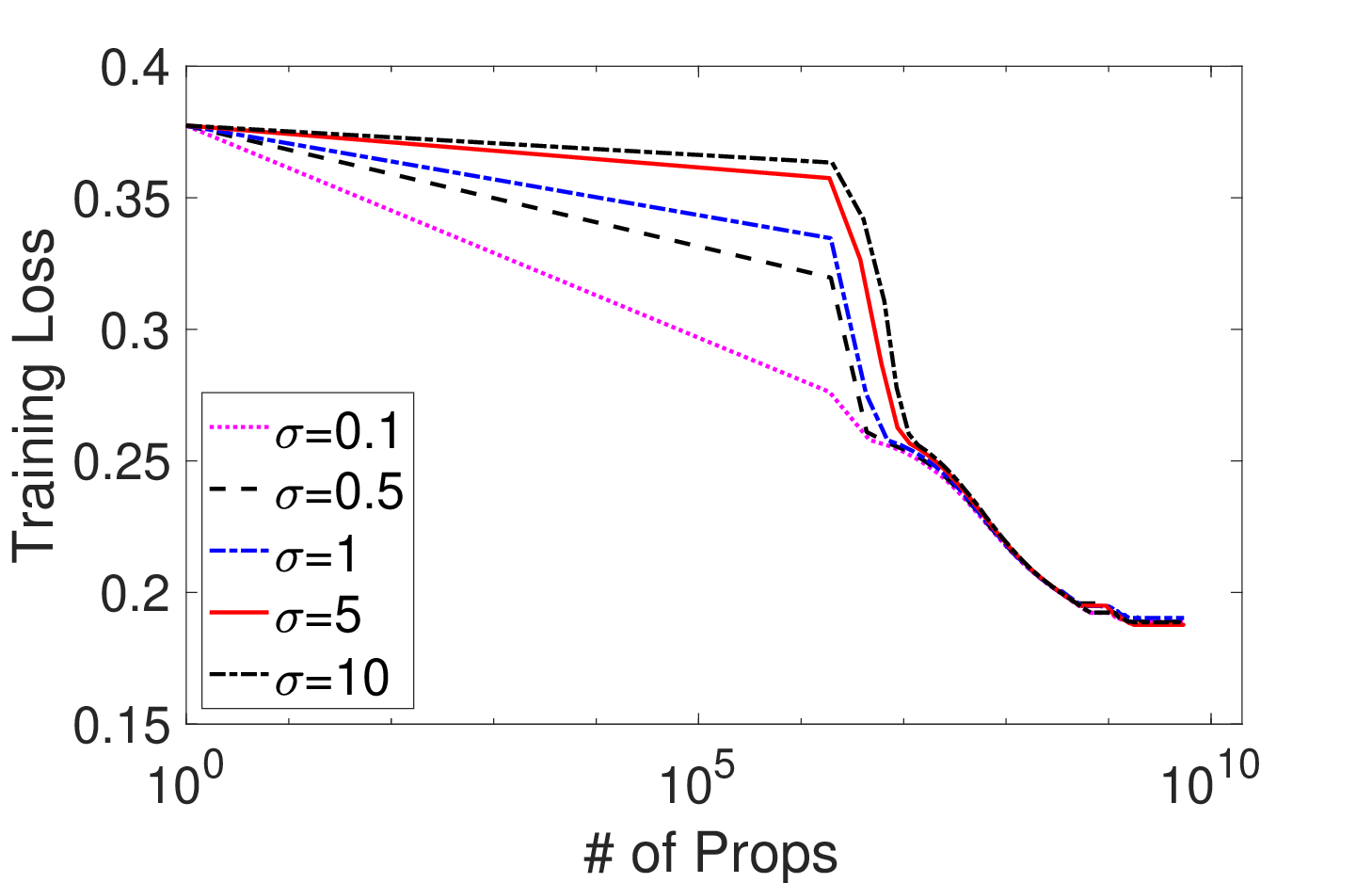}
 		\\
 		\includegraphics[width=1.0\textwidth]{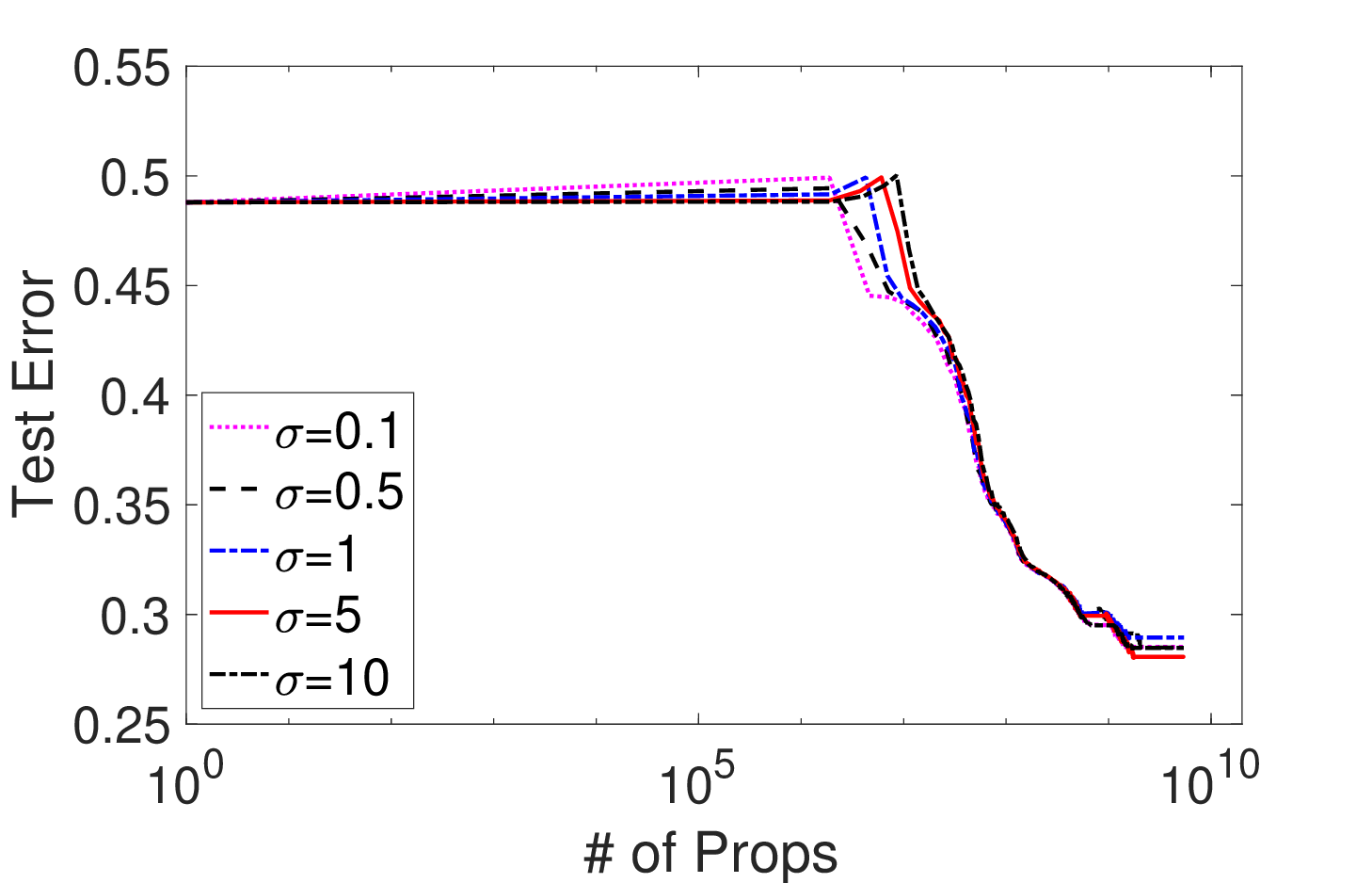}
 		\end{minipage}
 	}
 	\caption{ 
 	Discussion of the parameter:
 	Data: a9a.
 	initialization: random.
    Method: SHGF-TR and SHGF-ARC. 
    Axis: x-axis is the number of propagations and y-axis represents the training loss and test error.
    \textit{The first column}: Sample size: $0.01n$, $0.03n$, $0.05n$, $0.07n$, and $0.09n$, $n$ is the sample size; 
    \textit{The second column}: the radius $\Delta=4,6,8,10,12$; 
    \textit{The third column}: the parameters $\sigma=0.1,0.5,1,5,10$.
}
\label{Newton:Figure:Data:9a9:Dsicussioon}
\end{figure*}

\subsection{Experimental  Setting}
In our experimental results, we keep the measurement with trust-region based methods {\color{blue}\cite{xu2017second}\cite{xu2017newton}}. We plot the training loss and test error quantities v.s. the total number of propagations up to each iteration, in which counting the number of propagations is more appropriate as stated in \cite{xu2017second}. The total number of propagation per iteration for the algorithm considered in the paper are listed in TABLE \ref{Newton:Table:methods-main},
i.e.,
$\# \;{\rm{of}}\;{\rm{Props}} = \left| {{\mathcal{S}_h}} \right| + 2\left| {{\mathcal{S}_g}} \right| + 2\gamma\left| {{\mathcal{S}_B}} \right|$,
where ${\mathcal{S}_h}$, ${\mathcal{S}_g}$, and ${\mathcal{S}_B}$ are the sample sets used to compute the function value, gradient, and Hessian matrix, respectively;  $\gamma$ is the number of Hessian-vector products used to solve the subproblems in  \eqref{Newton:STR:Algorithm-solution} and  \eqref{Newton:SARC:Algorithm-solution}, and is given by the subproblem solver.
As our focus is to verify the efficiency of the proposed algorithms, we selected the original TR/ARC \cite{conn2000trust,cartis2011adaptivea,cartis2011adaptiveb} and several modified versions given different conditions for comparison – for example, an estimated Hessian matrix \cite{chen2018adaptive,bellavia2021adaptive, xu2017newton, xu2017second}, an estimated Hessian matrix and gradient \cite{cartis2012oracle,kohler2017sub,cartis2018global,tripuraneni2018stochastic}, and estimated Hessian matrix, gradient, and function in Algorithm \ref{Newton:STR:Algorithm} and Algorithm \ref{Newton:SARC:Algorithm}.  Note that we denote Full-TR/ARC by the original algorithm of TR and ARC, SH-TR/ARC by the stochastic algorithm of TR and ARC using the estimated Hessian, SHG-TR/ARC by the stochastic algorithm of TR and ARC using the estimated Hessian and gradient, and SHGF-TR/ARC by the stochastic algorithm of TR and ARC using the estimated Hessian, gradient, and function value.

\begin{table}[t]
	\centering
	\caption{The comparison methods and their numbers of propagation.}
	\begin{tabular}{|l|l|}
		\hline \multicolumn{1}{|c|}{Method} & \multicolumn{1}{|c|}{\# of Props} 
		\\
		\hline
		Full-TR/ARC \cite{conn2000trust,cartis2011adaptivea,cartis2011adaptiveb}
		& $n+2n+2\gamma n$ 
		\\
		\hline
		SH-TR/ARC \cite{chen2018adaptive,bellavia2021adaptive, xu2017newton, xu2017second}
		&  $n+2n+2\gamma\left| {{\mathcal{S}_B}} \right|$ 
		\\
		\hline
		SHG-TR/ARC \cite{cartis2012oracle,kohler2017sub,cartis2018global,tripuraneni2018stochastic}
		&$n+2\left| {{\mathcal{S}_g}} \right|+2\gamma\left| {{\mathcal{S}_B}} 
		\right|$ 
		\\
		\hline
		SHGF-TR/ARC (Ours)
		&$\left| {{\mathcal{S}_h}} \right|+2\left| {{\mathcal{S}_g}} 
		\right|+2\gamma\left| {{\mathcal{S}_B}} \right|$\\
		\hline
	\end{tabular}%
	\label{Newton:Table:methods-main}%
\end{table}

\begin{figure*}[h]
	\centering
	\subfigure{		
	\begin{minipage}[b]{0.3\textwidth}		\includegraphics[width=1.0\textwidth]{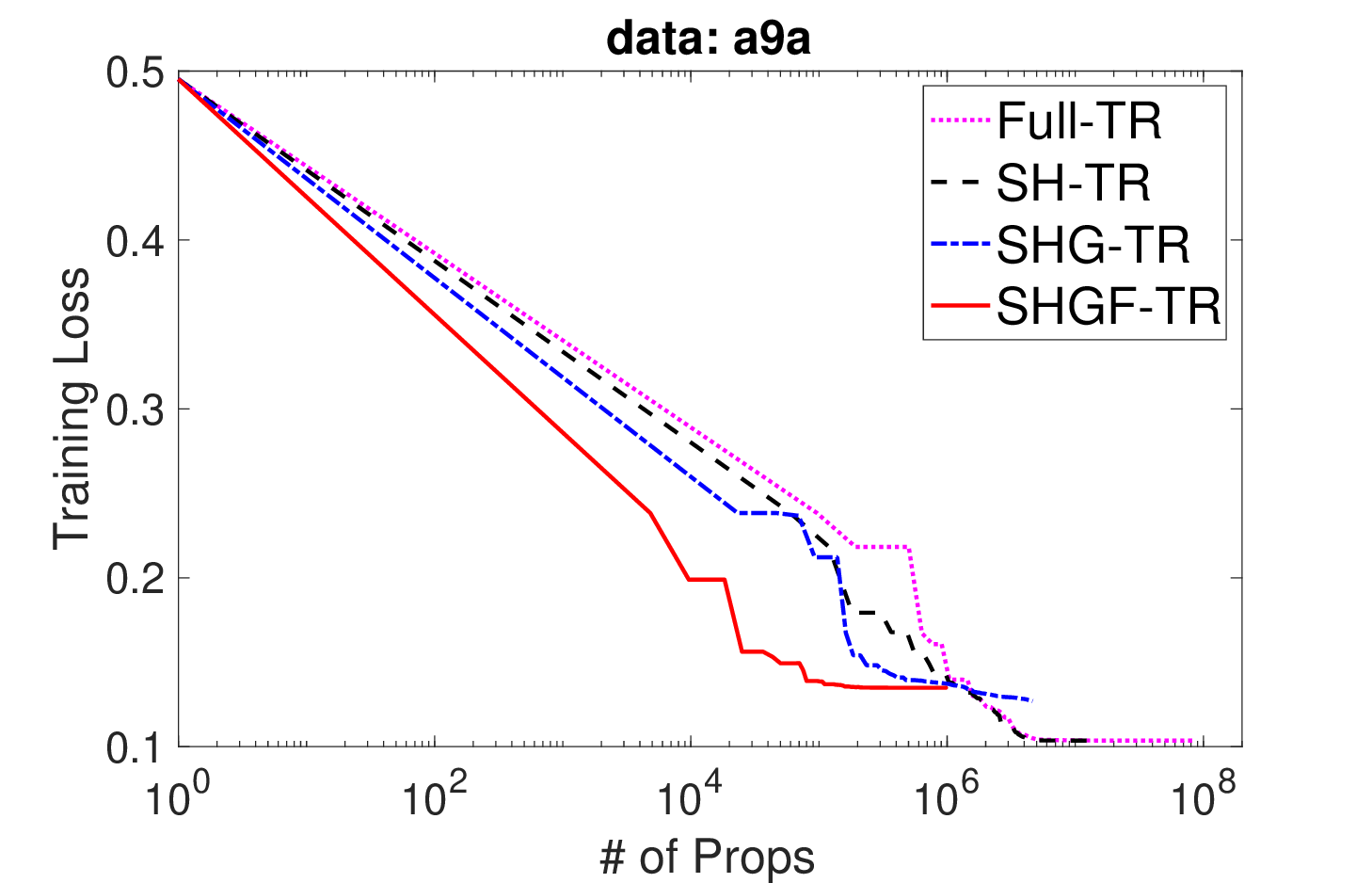}\\	\includegraphics[width=1.0\textwidth]{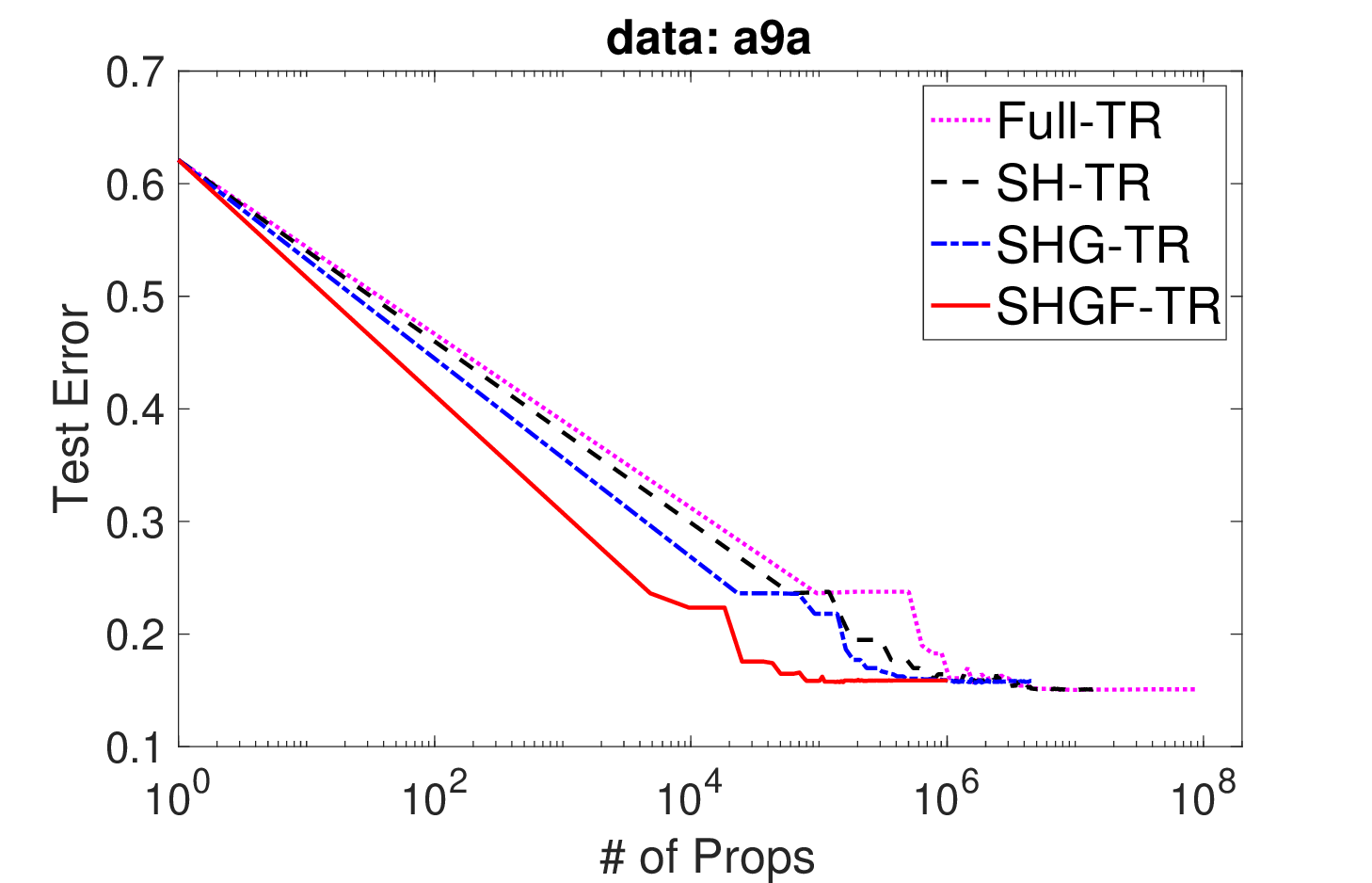}
		\end{minipage}
	}
	\subfigure{
	\begin{minipage}[b]{0.3\textwidth}
    \includegraphics[width=1.0\textwidth]{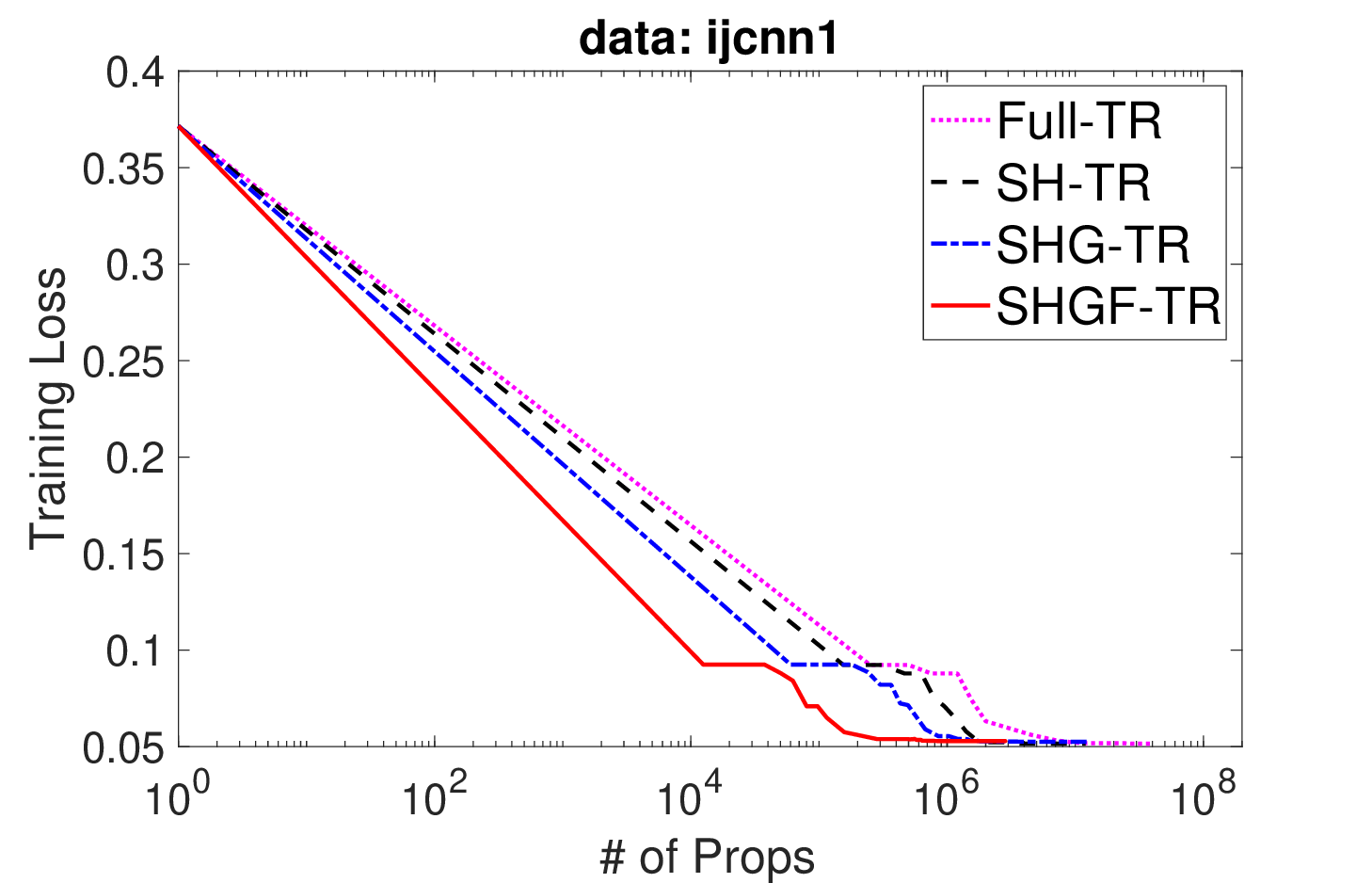}\\	\includegraphics[width=1.0\textwidth]{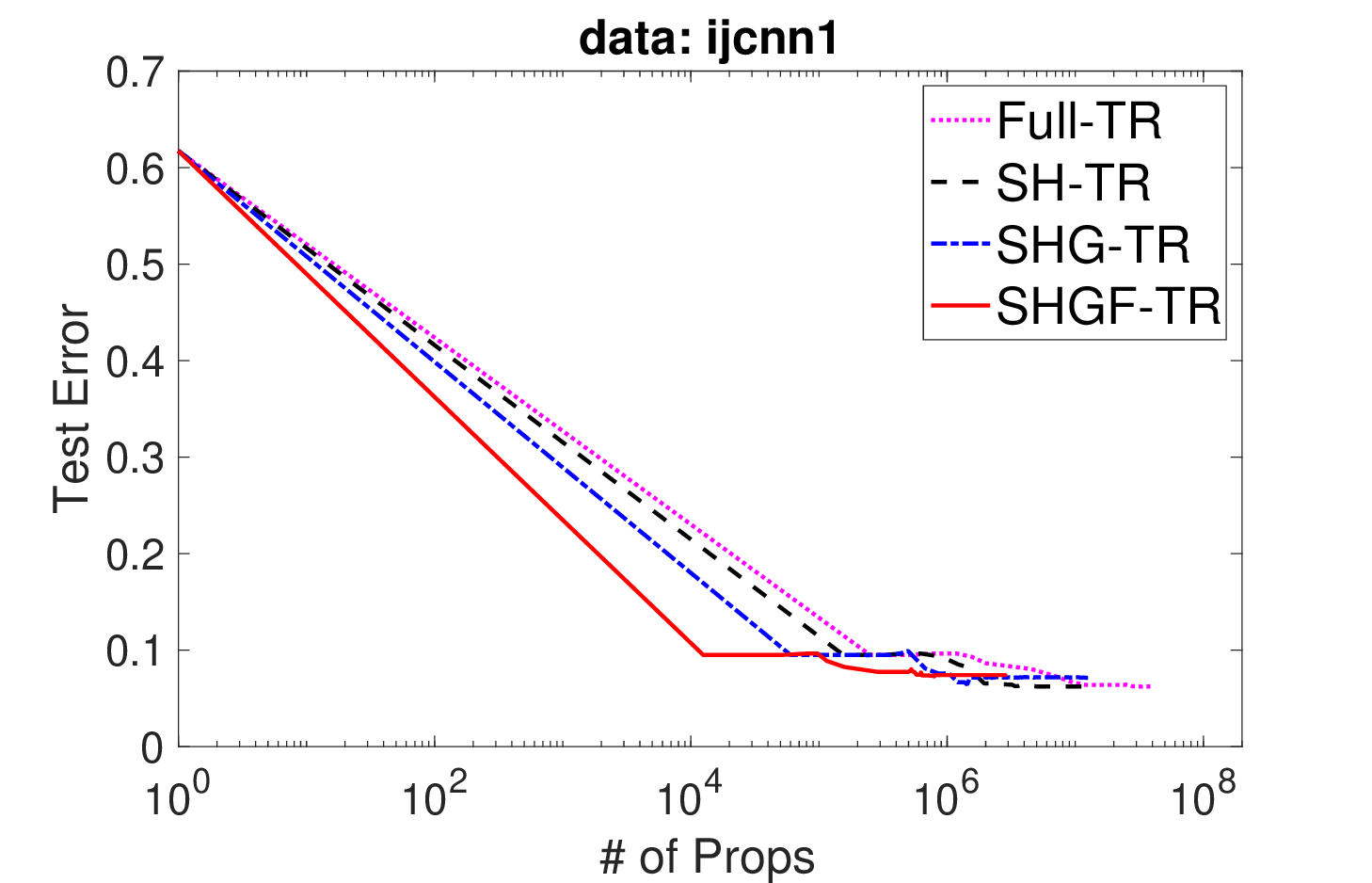}
		\end{minipage}
	}
	\subfigure{	\begin{minipage}[b]{0.3\textwidth}
    \includegraphics[width=1.0\textwidth]{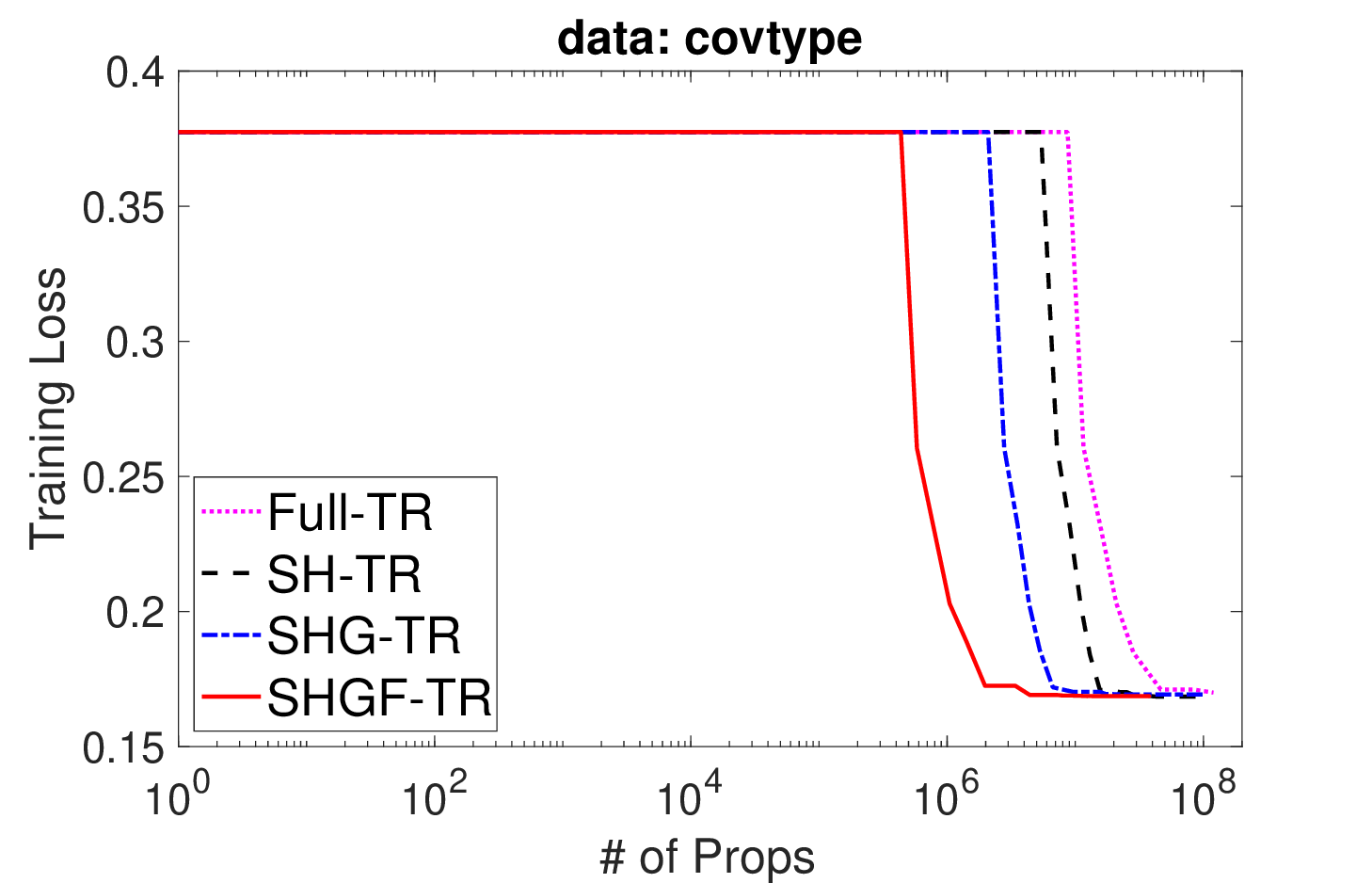}\\	\includegraphics[width=1.0\textwidth]{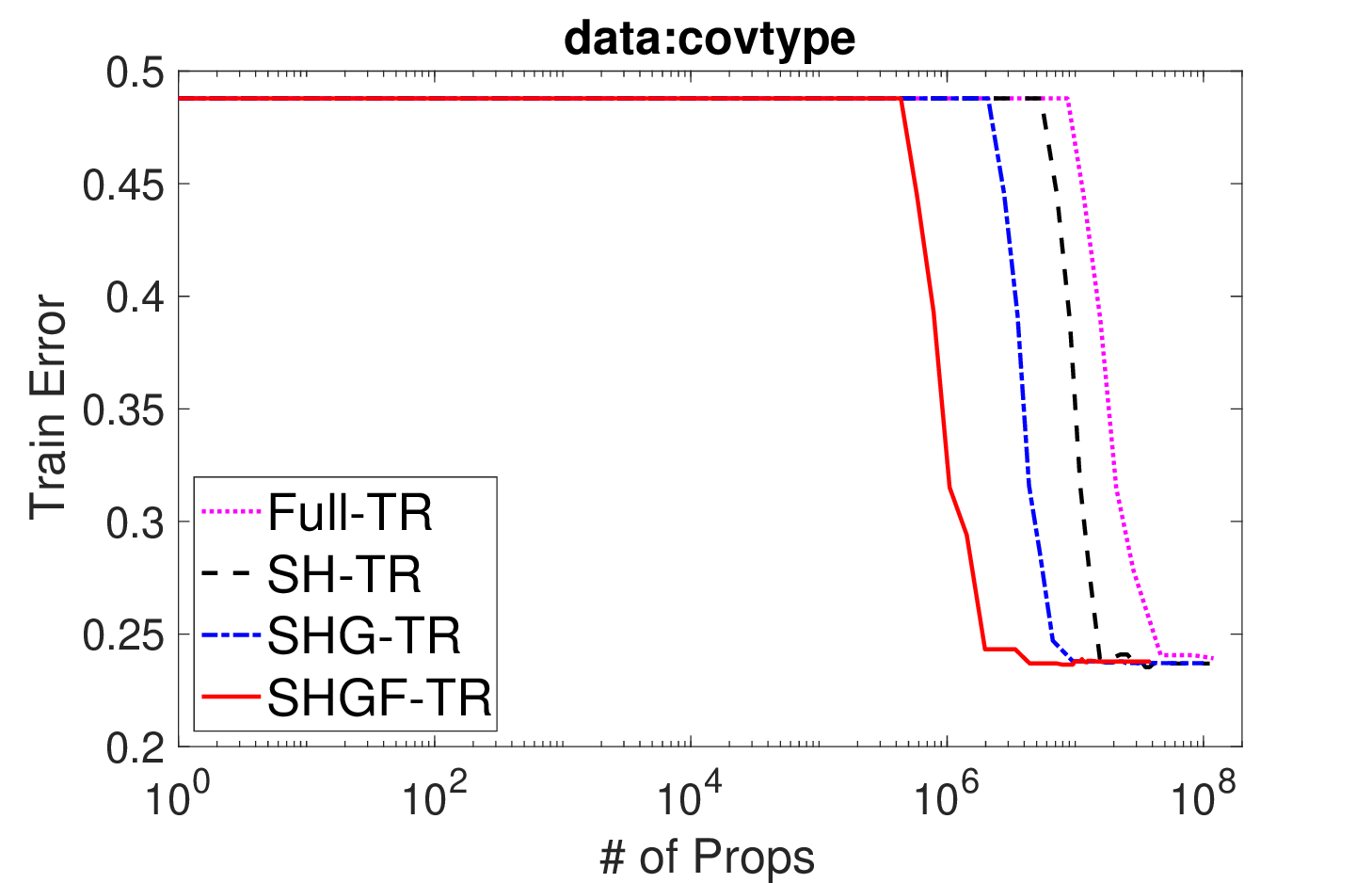}
		\end{minipage}
	}
	\caption{Data: ijcnn1, a9a, and covtype.
	Method: Trust-region. 
	Method: Full-TR, SH-TR, SHG-TR, and SHGF-TR(Ours).
	Axis: x-axis is the number of propagations and y-axis represents the training loss and test error.}
	\label{Newton:Figure:Method:TR:Data:ijcnn_a9a_cov}
\end{figure*}

\subsection{Non-Linear Least Squares}
We present the detailed numerical experiments conducted with STR and SARC on a non-linear least squares problem
The experimental problem, the non-linear least squares problem, is comparative and was designed to verify our theoretical results. 
\begin{table}[t]
	\centering
	\caption{Datasets description.}
	\begin{tabular}{|c|c|c|c|}
		\hline
		Data  & Dimension &\# of Train  & \# of Test \\
		\hline
		ijcnn1 & 22    & 49,990 & 91,701 \\
		\hline
		covtype & 54    & 581,012 & - \\
		\hline
		a9a   & 123   & 19,276
		& 16,281 \\
		\hline
	\end{tabular}%
	\label{Newton:Experiment:Dataset}%
\end{table}%
Consider a non-linear least squares loss with a logistic function. This experiment involves minimizing the following empirical risk problem,
\begin{align*}
\mathop {\min }\nolimits_{x \in {\mathbb{R}^d}} \frac{1}{n}\sum\nolimits_{i = 1}^n {{{\left( {{y_i} -\phi \left( {\left\langle {{a _i},x} \right\rangle } \right)} \right)}^2}}+\frac{1}{2}\|x\|^2, 
\end{align*}
 where $\phi ( z ) = 1/( {1 + {e^{ - z}}} )$ is the  sigmoid function  and $( {{a_i},{y_i}} ), {a_i} \in {\mathbb{R}^d},y \in \{ {0,1} \},i \in [n]$ are the observation sample, which are used for the binary linear classification. 
To verify our method, we minimized the above problem with three datasets and three other methods in addition to our own.  The details of datasets\footnote{https://www.csie.ntu.edu.tw/~cjlin/libsvmtools/datasets/binary.html. Note that the dataset covtype does not provide the test dataset.} are provided in TABLE \ref{Newton:Experiment:Dataset}.

\begin{figure*}[h]
	\centering
	\subfigure{		
	\begin{minipage}[b]{0.3\textwidth}		\includegraphics[width=1.0\textwidth]{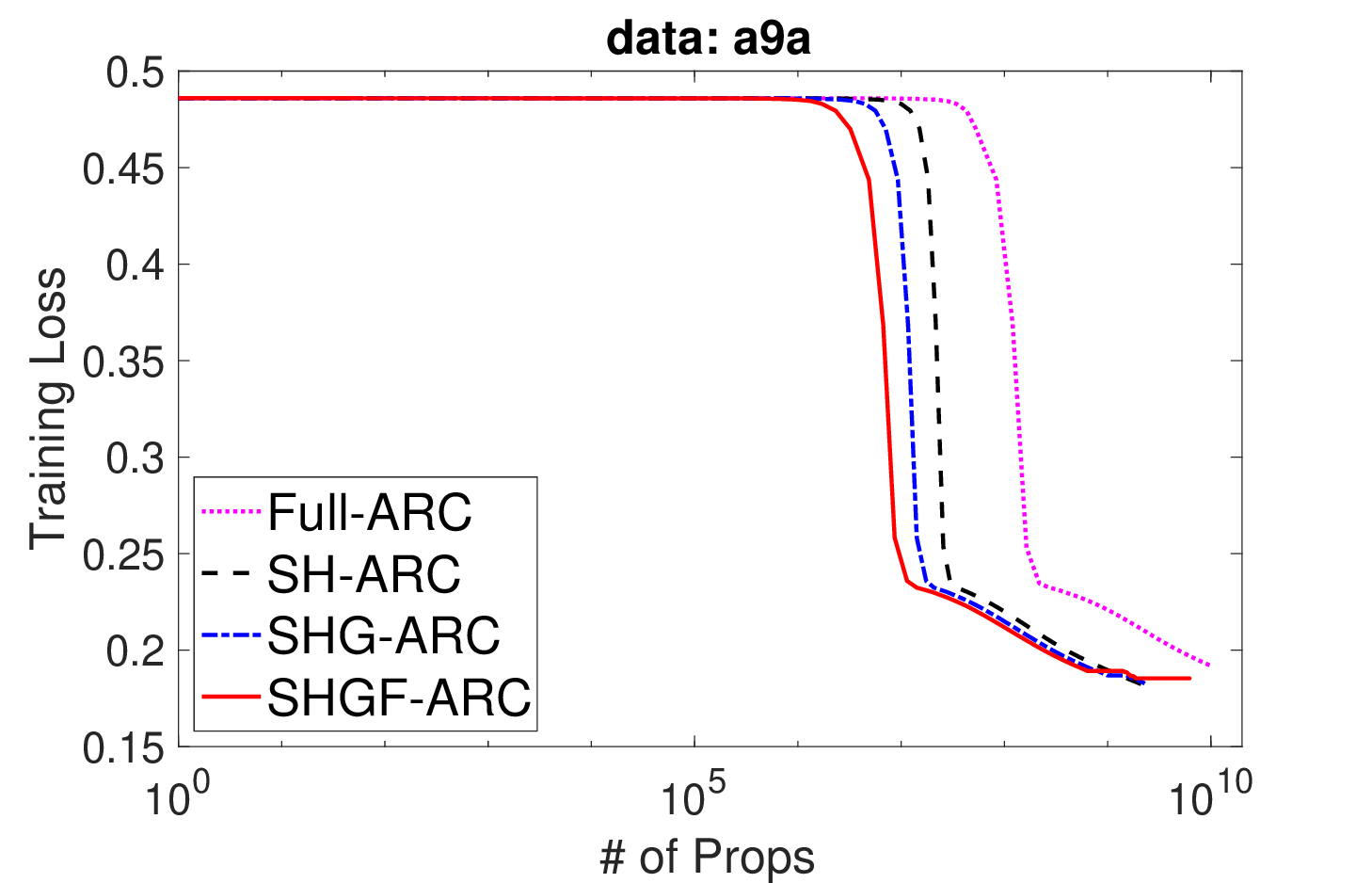}\\	\includegraphics[width=1.0\textwidth]{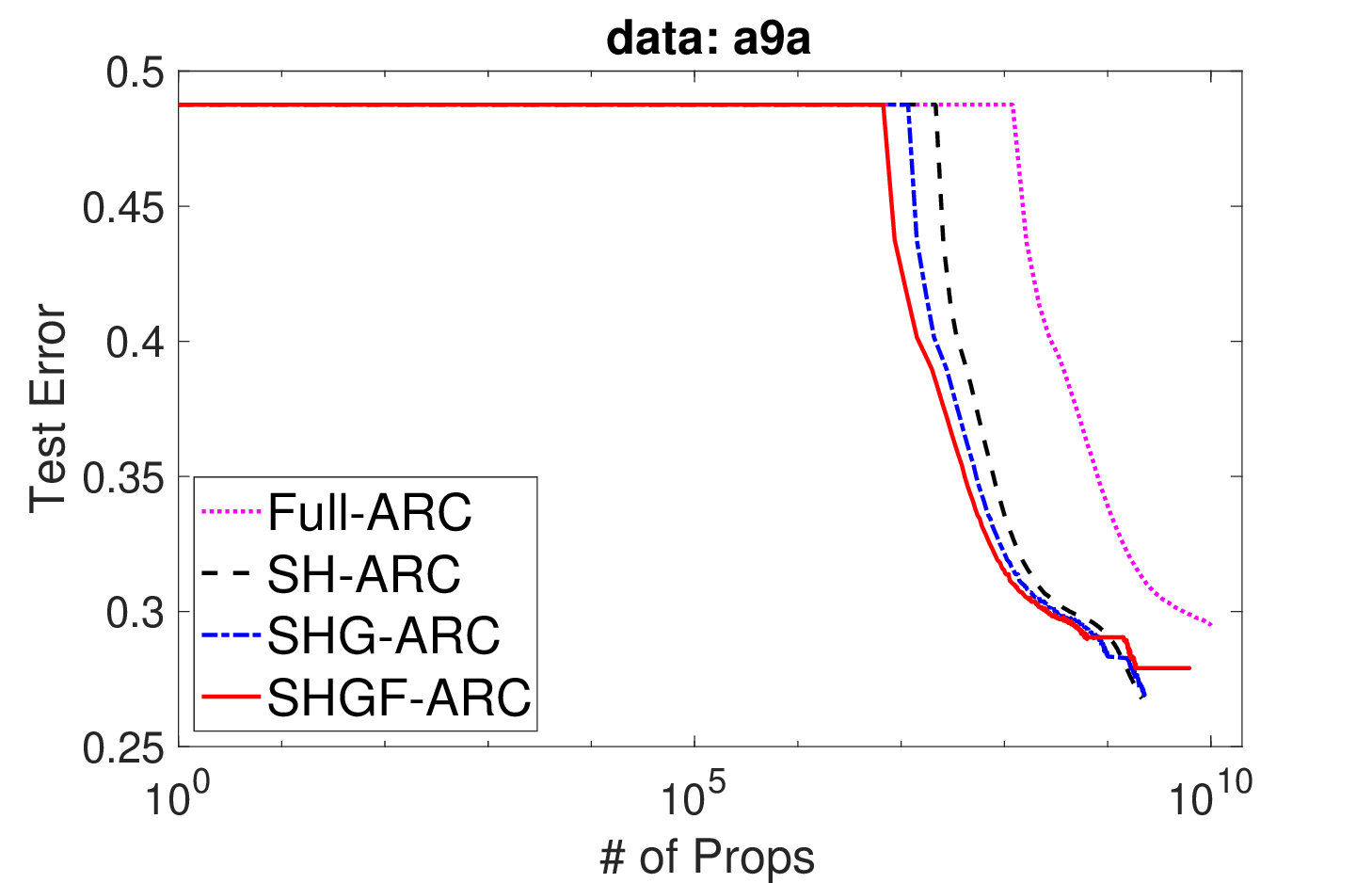}
		\end{minipage}
	}
	\subfigure{
	\begin{minipage}[b]{0.3\textwidth}
    \includegraphics[width=1.0\textwidth]{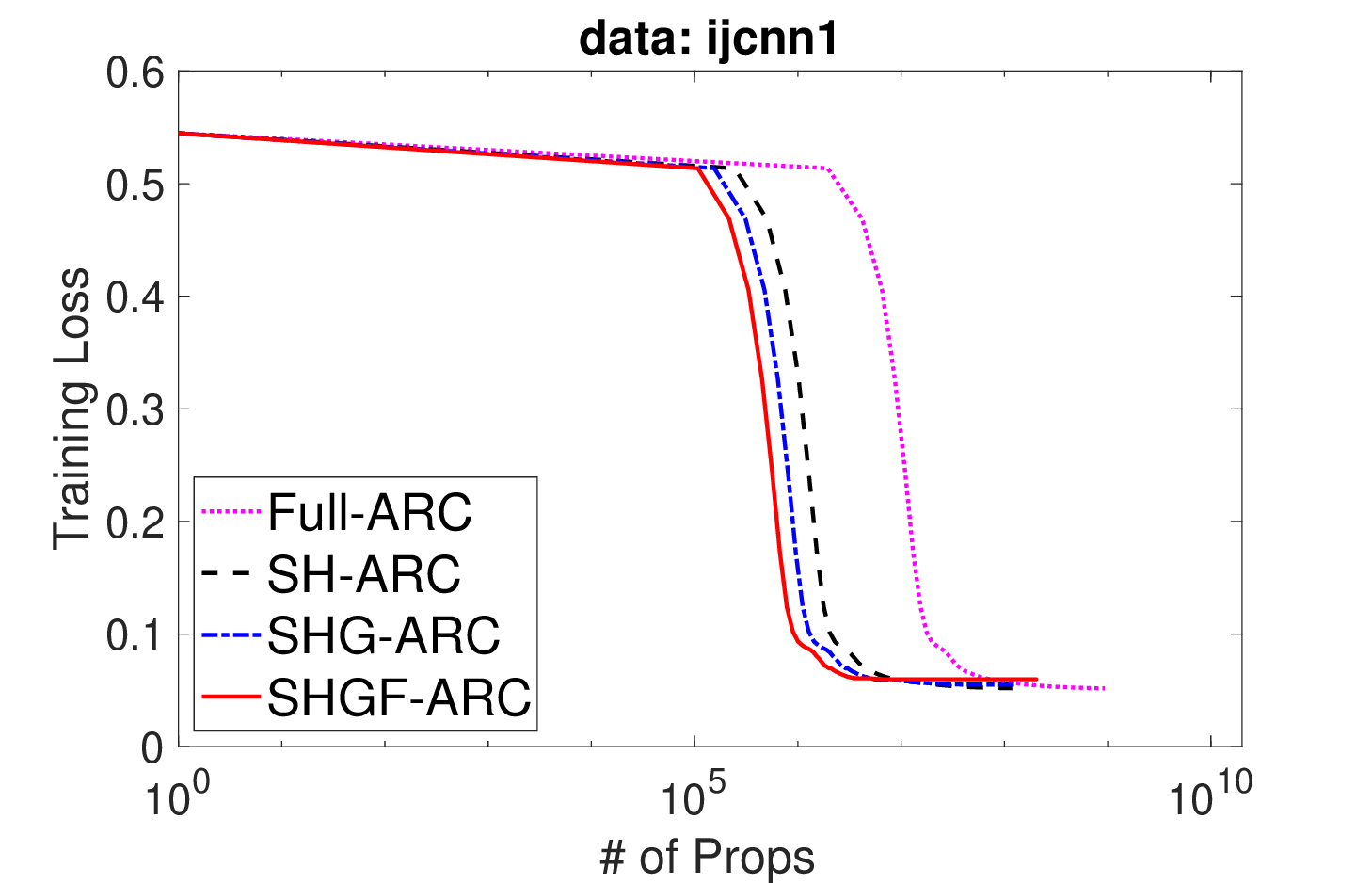}\\	\includegraphics[width=1.0\textwidth]{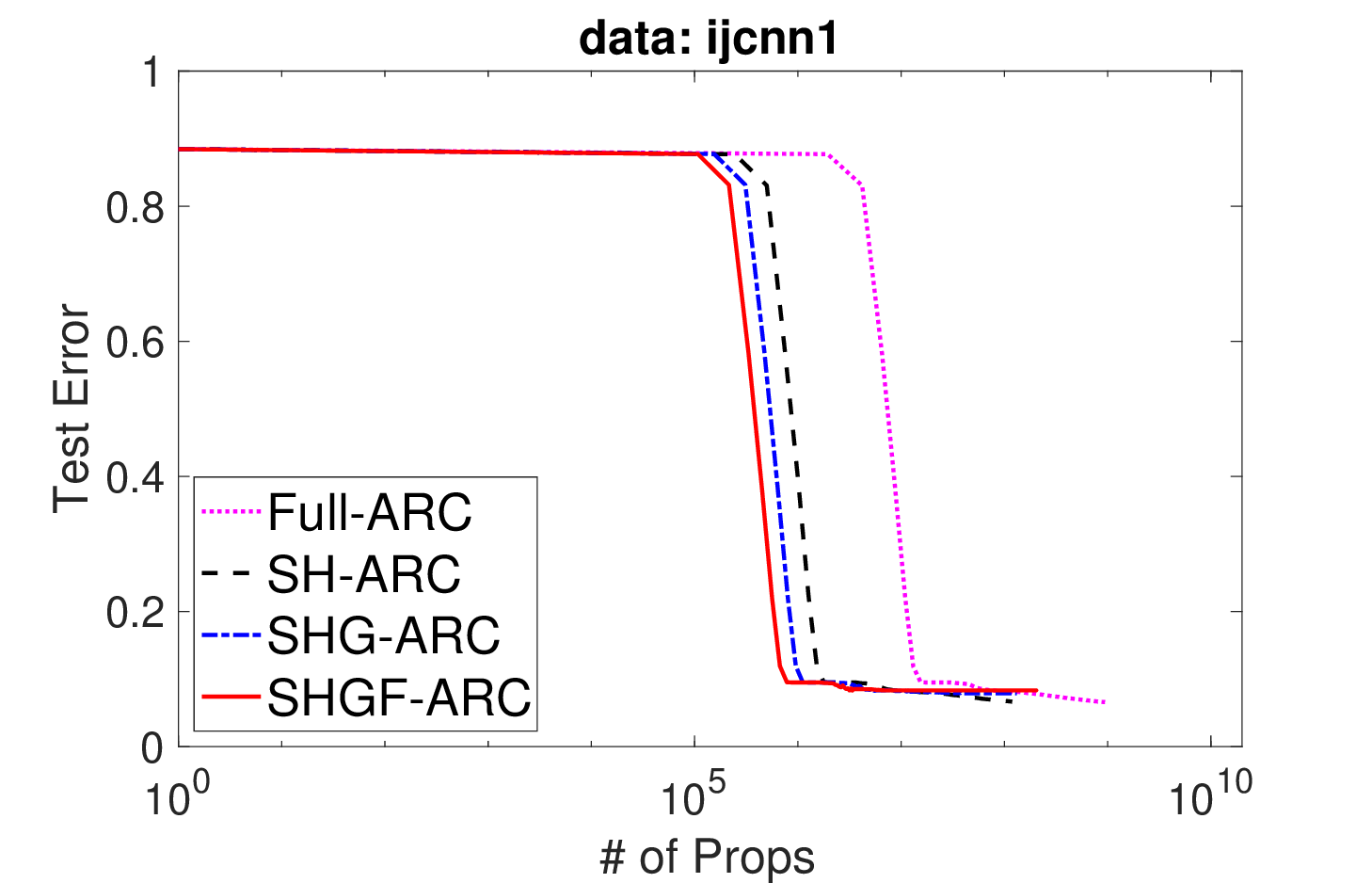}
		\end{minipage}
	}
	\subfigure{	\begin{minipage}[b]{0.3\textwidth}
    \includegraphics[width=1.0\textwidth]{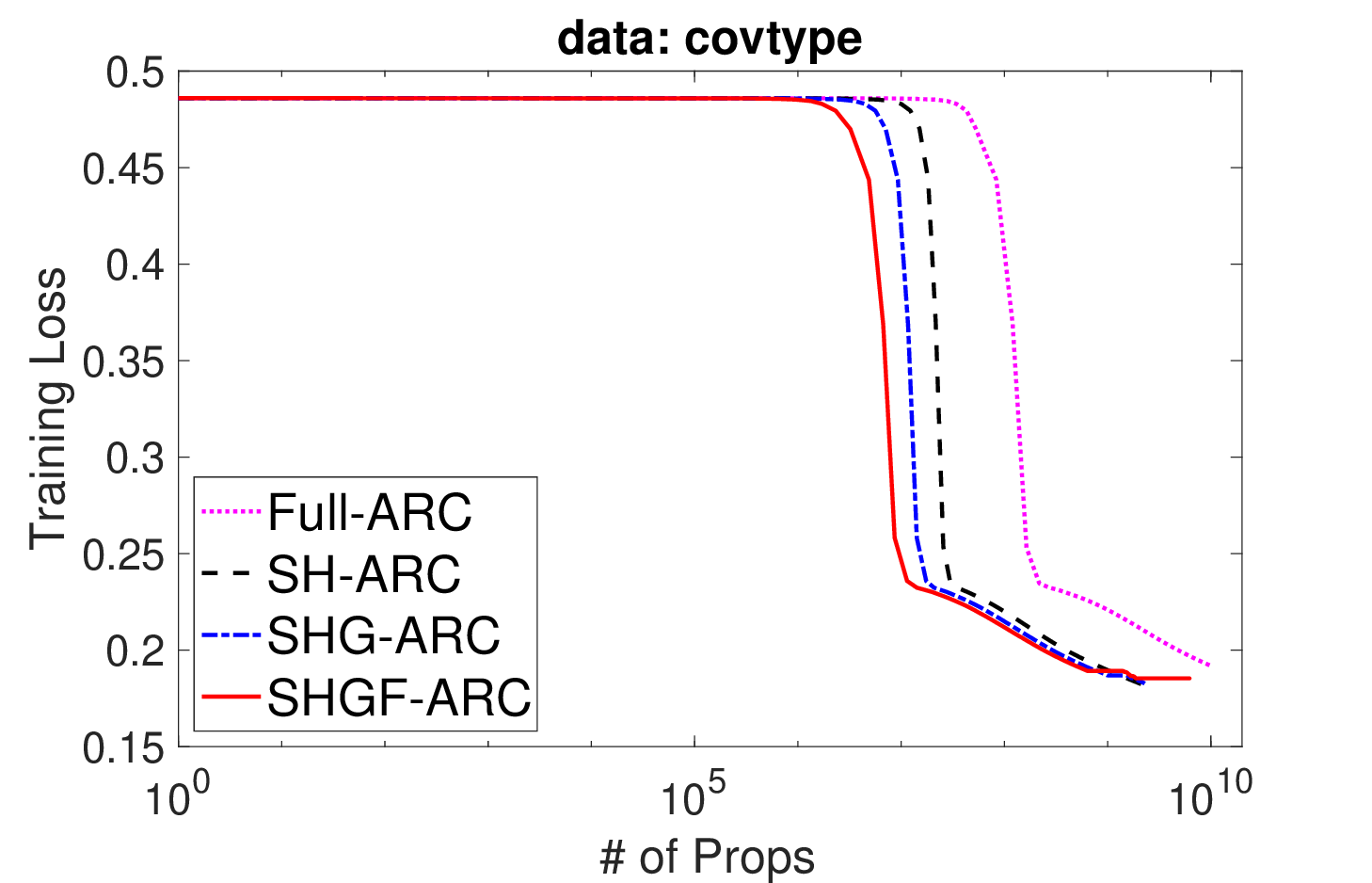}\\	\includegraphics[width=1.0\textwidth]{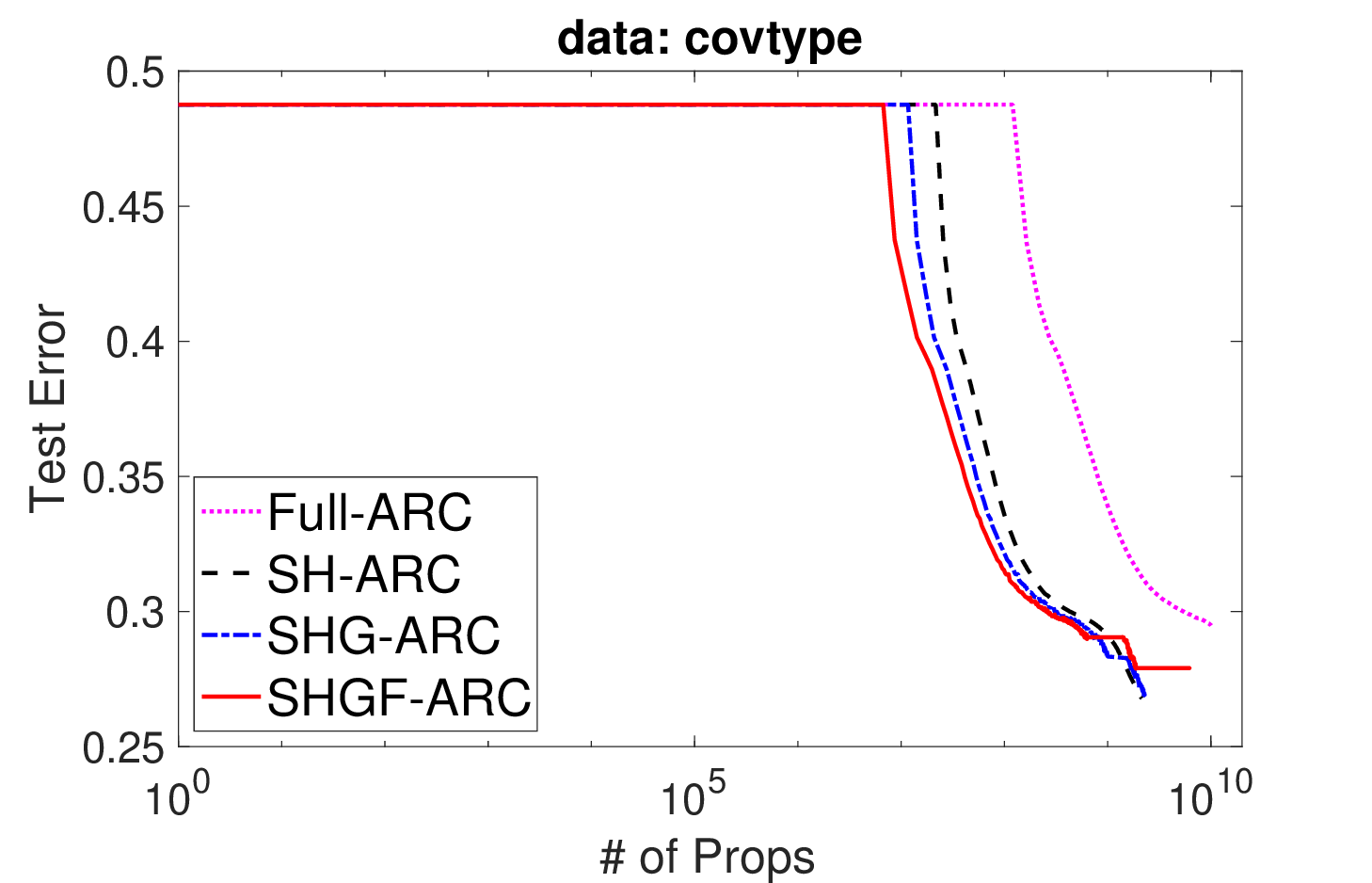}
		\end{minipage}
	}
	\caption{Data: ijcnn1, a9a, and covtype.
	Method: Adaptive regularization using cubics. 
	Method: Full-ARC, SH-ARC, SHG-ARC, and SHGF-TR(Ours).
	Axis: x-axis is the number of propagations and y-axis represents training loss and test error.}
	\label{Newton:Figure:Method:ARC:Data:ijcnn_a9a_cov}
\end{figure*}

\begin{figure*}
	\centering
	\subfigure{
		\begin{minipage}[b]{0.3\textwidth}
		\includegraphics[width=1.0\textwidth]{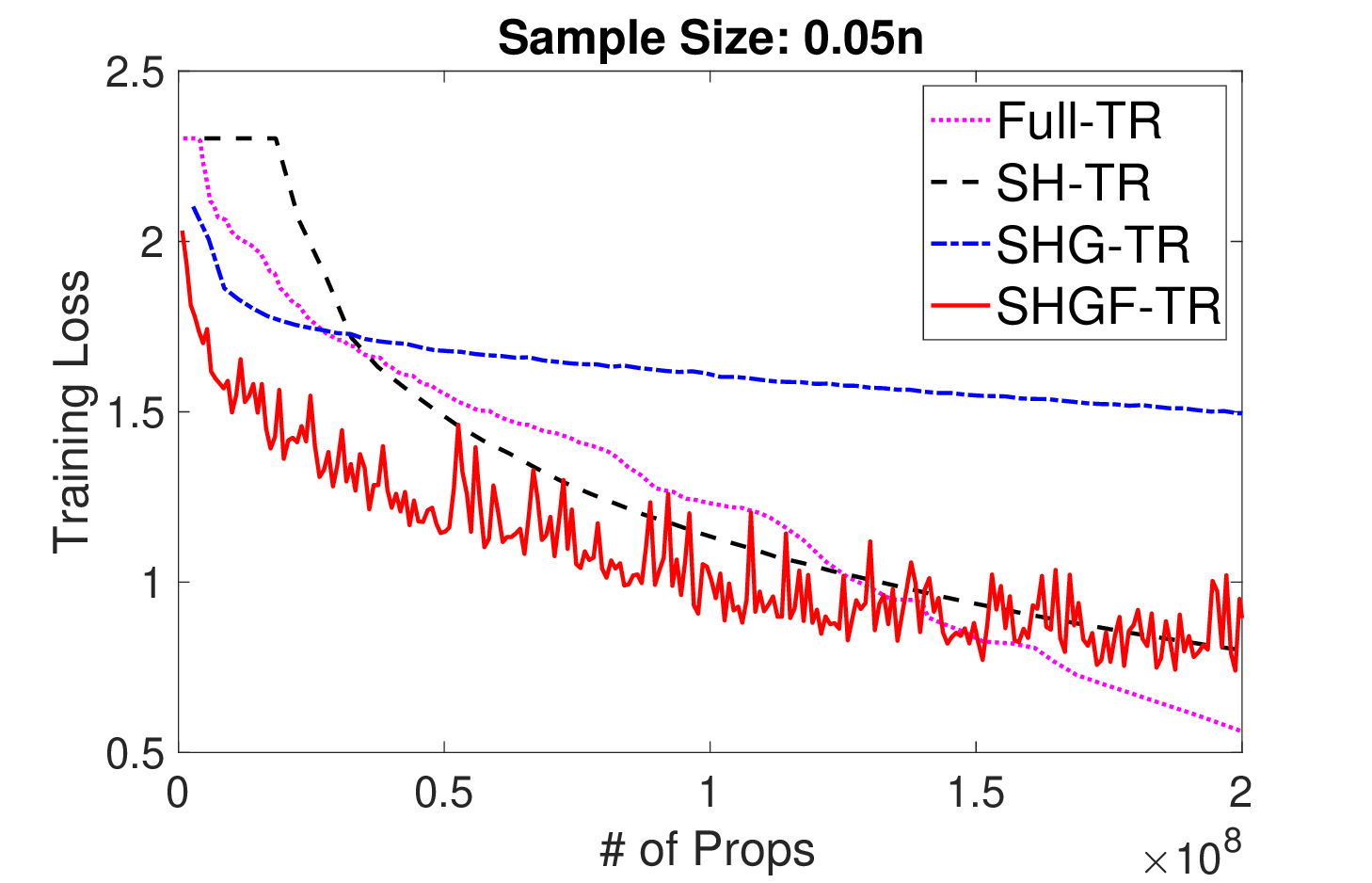}
		\\
		\includegraphics[width=1.0\textwidth]{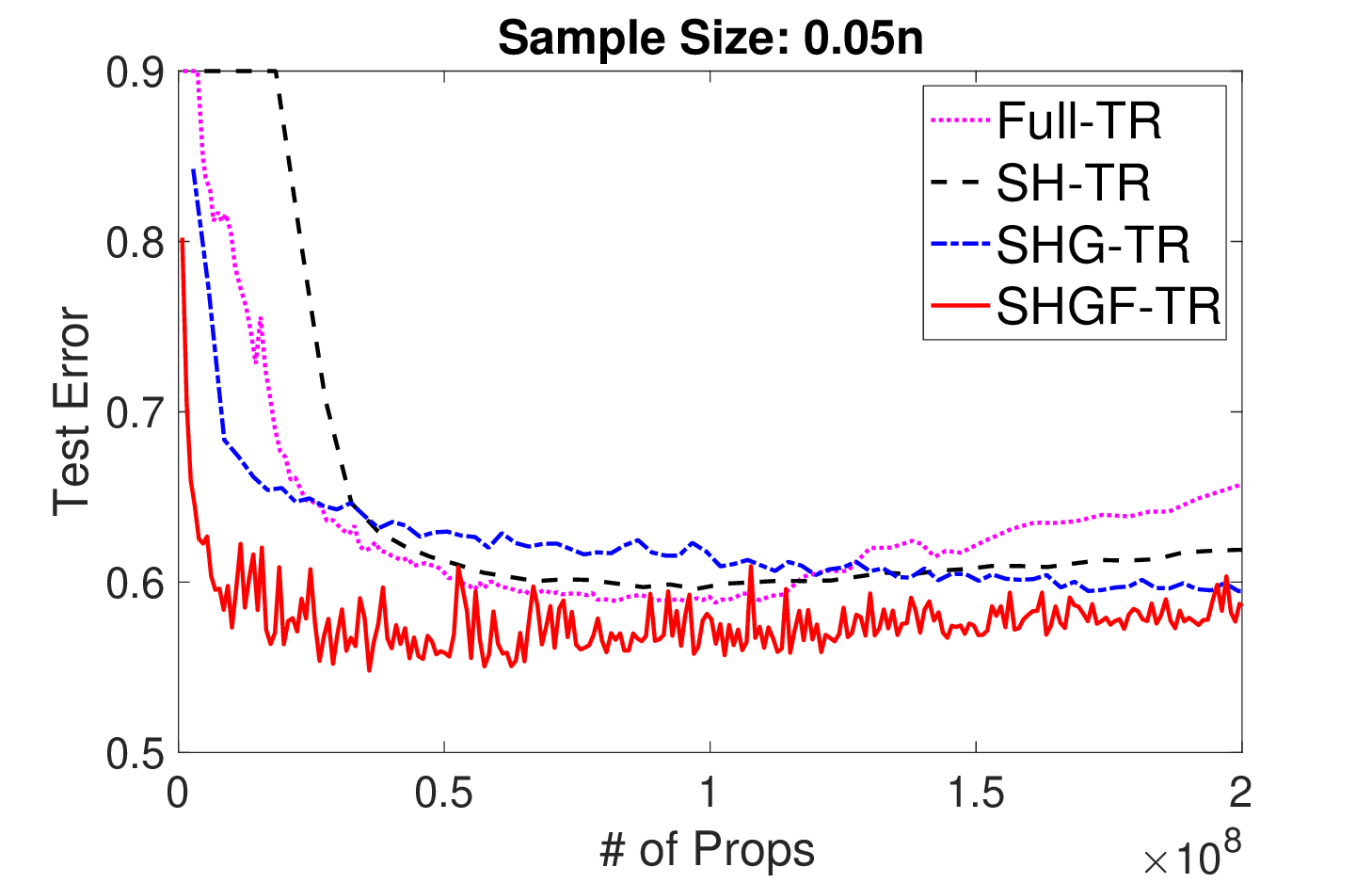}
		\end{minipage}
	}
	\subfigure{
		\begin{minipage}[b]{0.3\textwidth}
		\includegraphics[width=1.0\textwidth]{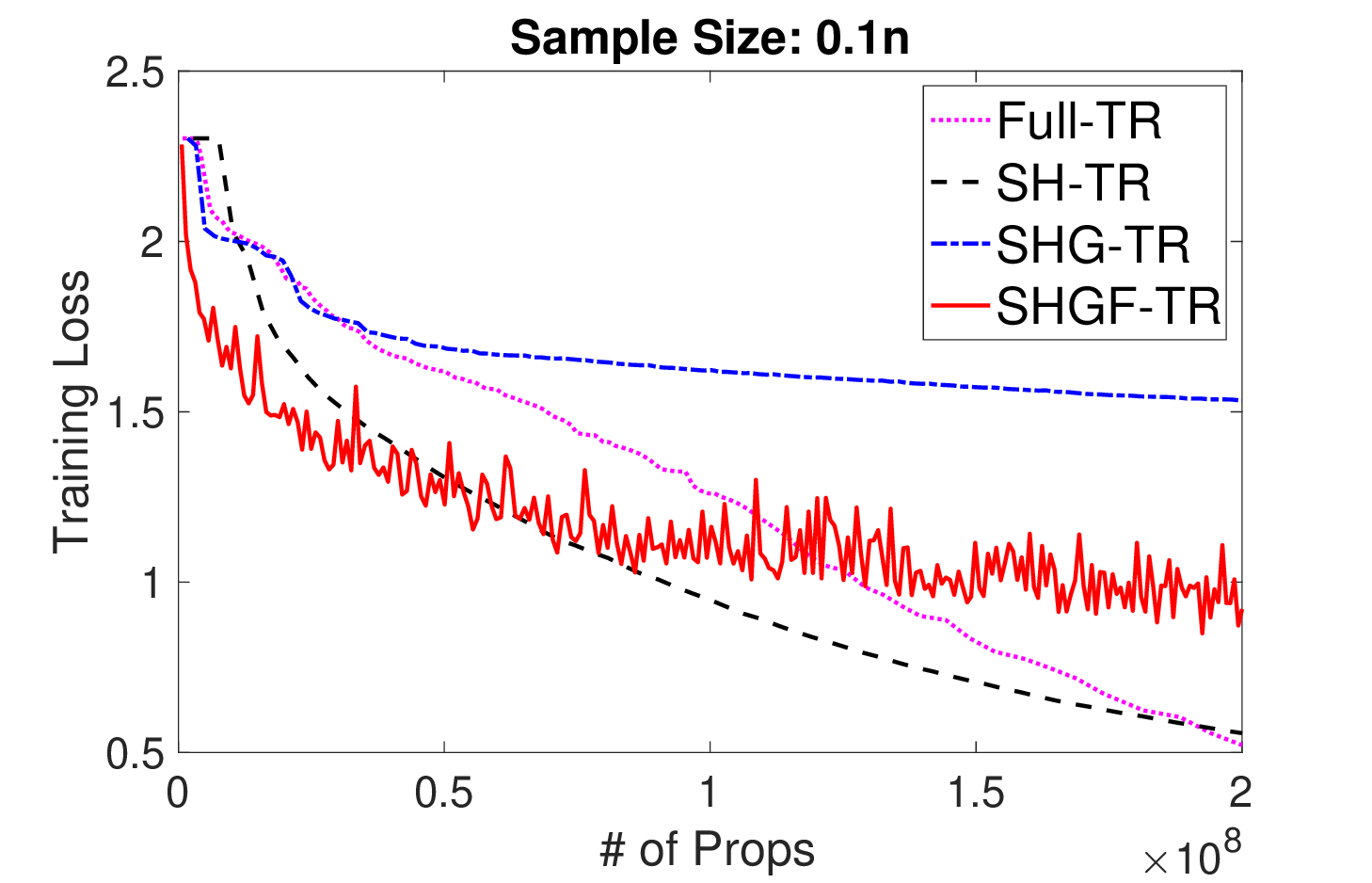}
		\\
		\includegraphics[width=1.0\textwidth]{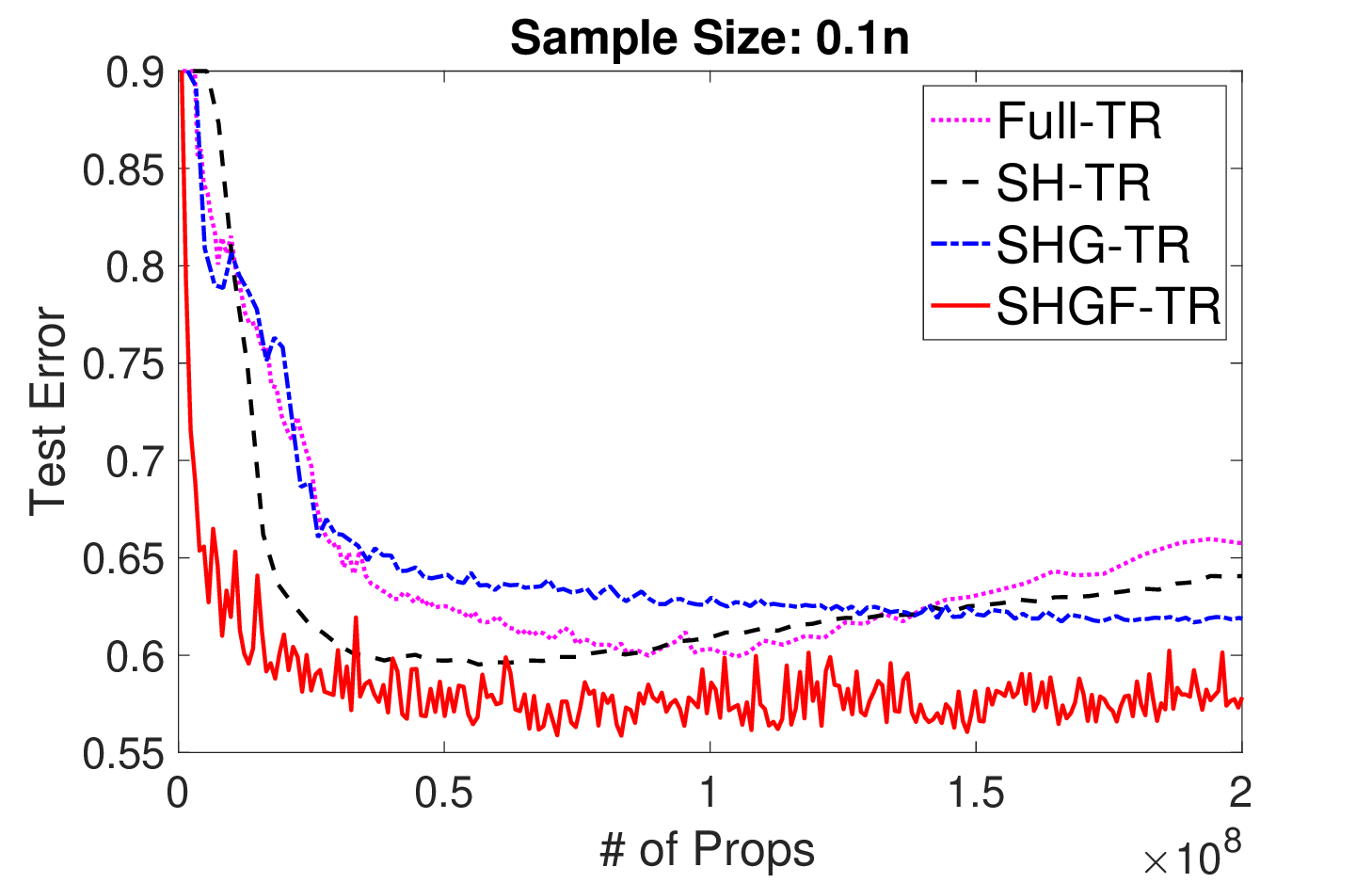}
		\end{minipage}
	}
	\subfigure{
		\begin{minipage}[b]{0.3\textwidth}
		\includegraphics[width=1.0\textwidth]{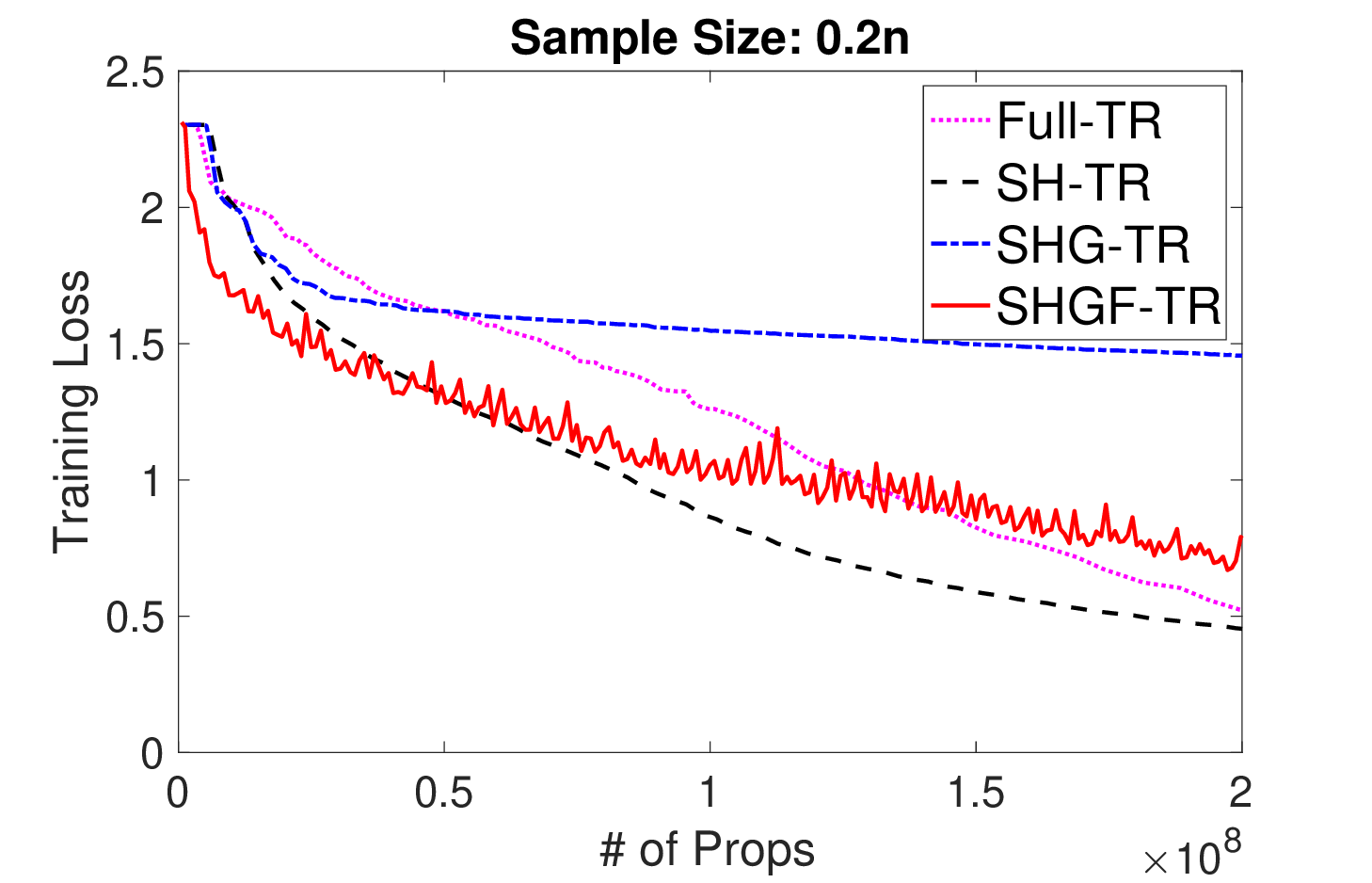}
		\\
		\includegraphics[width=1.0\textwidth]{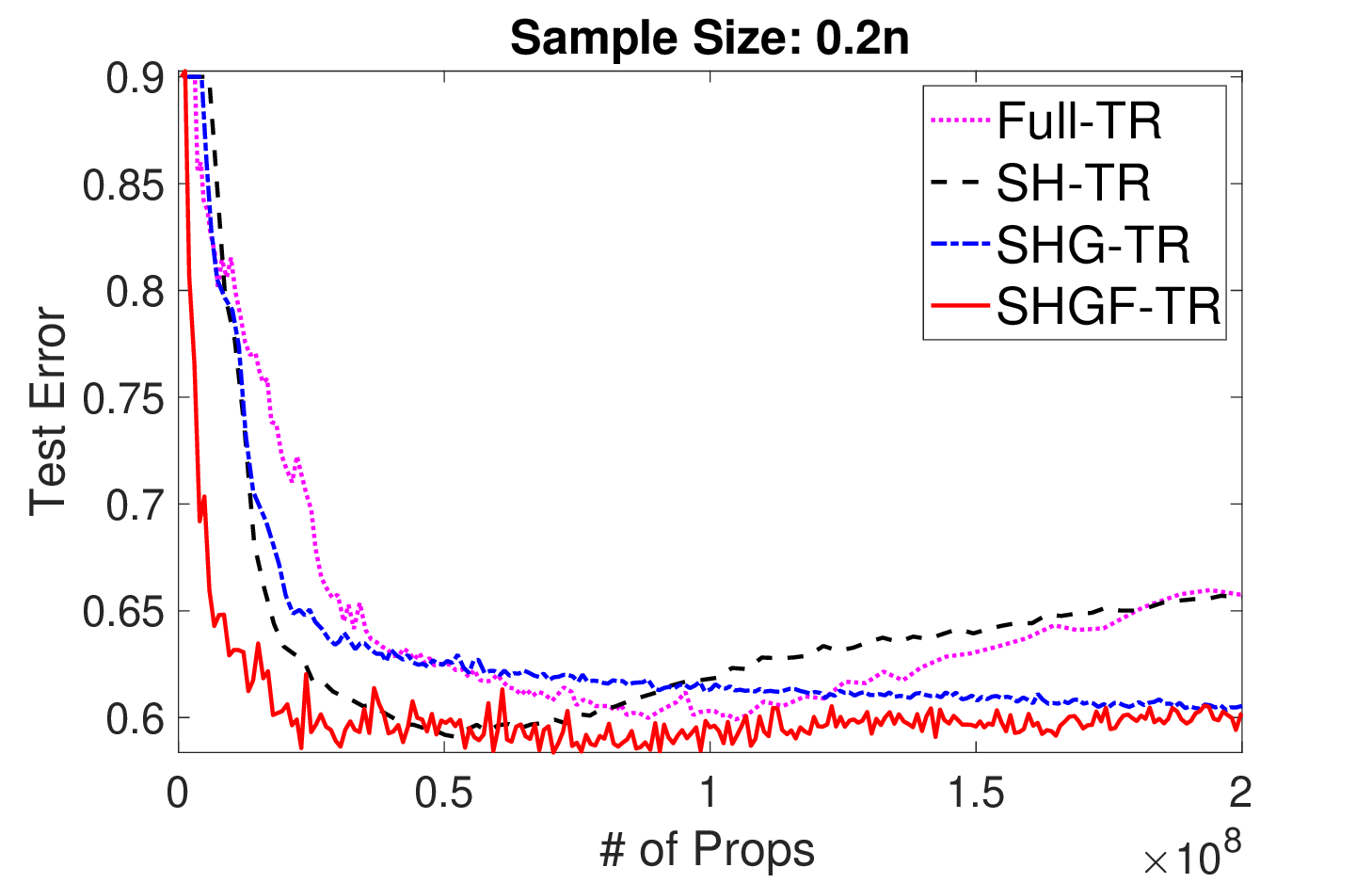}
		\end{minipage}
	}
	\caption{Data: cifar10;
	Method: Full-TR, SH-TR, SHG-TR, and SHGF-TR(Ours).
	The sample size: $0.05n$, $0.1n$, and $0.2n$.
	Axis: x-axis is the number of propagations and y-axis represents training loss and test error.
		}
	\label{Newton:Figure:Method:TR:Data:cifar10}
\end{figure*}

\subsubsection{Discussion of the parameter}

We first consider the different sizes of subsets ${\mathcal{S}_h}$, ${\mathcal{S}_g}$, and ${\mathcal{S}_B}$,  then discuss the relationships between these subsets, and finally compare our proposed algorithm with the fully stochastic version method. 
In  Fig.~\ref{Newton:Figure:Data:9a9:Dsicussioon}, the first column shows the performance of train loss and test error under different sizes of subsets. We present an initialization scheme: the initialization vector is set by random selection. Moreover, we set the size of samples as $0.01n$, $0.03n$, $0.05n$, $0.07n$, and $0.09n$ \footnote{The sample size 0.09n is not too small:1) experimental result shows its better performance; 2) the size of sample is based on $n$, while the size of $| {{\mathcal{S}_g}} |$ is based on $\epsilon_g$, $L_f$, and factor of $log$.}, where $n$ is the number of dataset.  
From Fig.~\ref{Newton:Figure:Data:9a9:Dsicussioon}, we observe that the smaller size presents the best performance at the beginning, but the final result is no better than others. The large size presents the best performance at the end of the iteration, but there will be more propagation. In the following experiment, we set the subset $0.05n$. Note that the performance of ``0.01n" and ``0.03n" are both worse, this is because we need sufficient samples to guarantee the convergence  such as we obtained the lower bound in Lemmas \ref{Newton:Subsample:Lemma_1}.
The second and third column of Fig.~\ref{Newton:Figure:Data:9a9:Dsicussioon} shows the performance of train loss and test error under different values of radius $\Delta$ and parameter $\sigma$, i.e.,  radius $\Delta=4,6,8,10,12$ and the parameters $\sigma=0.1,0.5,1,5,10$. We can see that our proposed algorithms have great robustness to the choice of its main hyper-parameter, i.e., the initial trust region$\Delta$ and parameter $\rho$. Consequently, in practice, one can run Algorithm \ref{Newton:STR:Algorithm} or \ref{Newton:SARC:Algorithm} merely one time, with almost any initial trust region radius $\Delta$ and parameter $\rho$, and expect reasonable performance in a reasonable amount of time.

\subsubsection{Experimental results}
In keeping with the performance evaluation in \cite{martens2012training} and \cite{xu2017second}, we measured performance by the total number of propagations per successful iteration in TABLE \ref{Newton:Table:methods-main}. We also considered two kinds of results – train loss and test error – given the initialization under random vectors.
All the results for the TR- and ARC-related algorithms are shown in Fig. \ref{Newton:Figure:Method:TR:Data:ijcnn_a9a_cov} and Fig. \ref{Newton:Figure:Method:ARC:Data:ijcnn_a9a_cov}.
From these results, we observe that the order of performance from best to worst was SHGF-TR/ARC, SHG-TR/ARC, SH-TR/ARC, then TR/ARC, which verifies the iteration complexity indicated by our theoretical analysis.  We can see that our proposed algorithms, the stochastic TR and ARC using estimated function value, gradient, and Hessian, would have the best performance on both train loss and test error, which is efficient and effective.
In particular,  the performances of SHGR-TR  are obviously better than that of SHG-TR, SH-TR, and Full-TR.

\subsection{Deep learning}
We apply our proposed algorithm for training deep learning. We consider 3-hidden layer network for image classification task using cifar10 dataset
\footnote{https://www.cs.toronto.edu/~kriz/ cifar.html.} 
to verify our proposed methods, in which we almost follow the same experimental setting as in \cite{xu2017second}. Specifically, the hidden layer size are [128,512,128]. The active functions are sigmoid functions. The last activation function of a neural network to normalize the output of a network is the softmax function.
In order to be consistent to the simulation experiment, we also do experiments on different sizes of samples.
The sizes of the samples are set to 5\%, 10\%, and 20\% of the total training set, i.e., $0.05n$, $0.1n$, and $0.2n$, where $n$ is the size of the dataset.
We apply the CG-Steihaug method \cite{nocedal2006numerical} to solve the subproblem in \eqref{Newton:Definition_m}.
Figure \ref{Newton:Figure:Method:TR:Data:cifar10} presents the experimental results of training loss and test error with random starting points. We can see that the performance of our proposed method is better than that of other methods in terms of test error.
Moreover, for different sizes of samples, our proposed methods outperform other methods.
Finally, from the perspective of the popular stochastic optimization in ML and DL, we proposed a sub-sample trick that is well fitted to the proof of the stochastic second-order methods under the inexact Hessian matrix, gradient, and function, while maintaining the convergence order and improving the performance.


\section{Conclusion}\label{Newton:Section:Conclusion}

In this paper, we presented a family of stochastic trust-region (STR) and stochastic cubic regularization (SARC) methods conditioned on an inexact function value, gradient, and a Hessian matrix. In addition, we also presented a subsampling technique to reduce the computational cost of estimating the function value of $f(x)$ when evaluating the role of $\rho$. A theoretical analysis of convergence and iteration complexity with the two approaches shows that iteration complexity stays of the same order but that the computational cost per iteration is reduced. This forms the main contribution of the paper.
Experiments on a non-linear, non-convex least squares problem 
show that the proposed methods are faster than the existing second-order methods. 
An interesting direction for future research is the use of the sampling technique to other classical optimizations in escaping shallow local minima while maintaining the same order of the convergence.




\ifCLASSOPTIONcaptionsoff
\newpage
\fi



{\small
\bibliographystyle{IEEEtran}
\bibliography{IEEEabrv,Main}
}

\appendices
\newpage

\onecolumn

\section{Proofs}\label{Newton:Appendix:theorem}

We present the proof frameworks of the stochastic trust-region and stochastic ARC, respectively.
\subsection{Proof for Stochastic Trust- Region}\label{Newton:Appendix:STR}

Here, we consider the upper and lower bounds of denominator and numerator in (\ref{Newton:STR:1-p}) in Lemmas       
 \ref{Newton:STR:Lemma:Upperbound_mx-msx} and Lemma \ref{Newton:STR:Lemma:Upperbound_fxs-msx}, respectively. Moreover, we separately give the corresponding bound under the index sets $\mathcal{S}_{\nabla f}$ and $\mathcal{S}_H$. Note that the detailed proofs of Lemma \ref{Newton:STR:Lemma:Upperbound_mx-msx}-\ref{Newton:STR:Lemma:Bound:Radius} are in Appendix.

\begin{lemma}\label{Newton:STR:Lemma:Upperbound_mx-msx}
	Suppose Assumptions 
	\ref{Newton:Assumption:approximation} and 
	\ref{Newton:Assumption:bound} hold,  $m_k(s)$ is defined in 
	(\ref{Newton:Definition_m}). For the case of
	$x_k\in\mathcal{S}_{\nabla f}$, if $m_k(s_k)\le m_k(s_k^C)$, where $s_k^C$ is the Cauchy 
	point, 	then we 	have
	\[{m_k}(0) - {m_k}({s_k}) \ge \frac{1}{2}({\epsilon_{\nabla f}} - {\epsilon_g})\min 
	\left\{ {{\Delta _k},({{{\epsilon_{\nabla f}} - {\epsilon_g}}})/{{{\kappa _H}}}} 
	\right\},\]
	For the case of $x_k\in \mathcal{S}_H$,  
	that is 
	${\lambda _{\text{min} }}( {\nabla^2f(x_k)} ) \le  - {\epsilon_H}$, there exists a vector $s_k$ such that $\langle 
	{ g({x_k}),{s_k}} \rangle \le 0$, $ {s_k^T\nabla^2f(x_k){s_k}}<v_{0}\lambda_\text{min}(\nabla^2f(x_k))
	\Delta_k^2$, and $\left\| {{s_k}} \right\| = {\Delta _k}$, where $v_{0}\in(0,1]$ is a constant,	then we have
    $
	m_k( 0) - {m_k}(s_k )\ge \frac{1}{2} ( {{\epsilon_H} - {\epsilon_B}})v_0\Delta _k^2.
	$
\end{lemma}

\begin{lemma}\label{Newton:STR:Lemma:Upperbound_fxs-msx}
	Suppose  Assumptions 
	\ref{Newton:Assumption:function},
	\ref{Newton:Assumption:approximation}, and \ref{Newton:Assumption:Bound:Variance} hold, 
	based 
	on the definition of $m_k(s)$ in 
	(\ref{Newton:Definition_m}),  we have
	{\small
	\begin{align*}
	&{m_k}\left( 0 \right) - {m_k}\left( {{s_k}} \right) - \left( {h\left( {{x_k}} \right) - h\left( {{x_k} + {s_k}} \right)} \right)\\ 
	\le& 2\left( {\frac{{\mathbb{I}\left( {\left| {{\mathcal{S}_h}} \right| < n} \right)}}{{\left| {{\mathcal{S}_h}} \right|}}H_1 + {\epsilon_g}} 	\right){\Delta _k}	
	+ \frac{3}{2}\left( {\frac{{\mathbb{I}\left( {\left| {{\mathcal{S}_h}} \right| < n} \right)}}{{\left| {{\mathcal{S}_h}} \right|}}{H_2} +L_H{\Delta _k} + {\epsilon_B}} \right)\Delta _k^2.
	\end{align*}
	}
	If $\mathcal{S}_h=\mathcal{S}_g$, we have $2\left( {\frac{{\mathbb{I}\left( {\left| {{\mathcal{S}_h}} \right| < n} \right)}}{{\left| {{\mathcal{S}_h}} \right|}}H_1 + {\epsilon_g}} 	\right){\Delta _k}=0.$
\end{lemma}

Note that, in Algorithm 
\ref{Newton:STR:Algorithm}, for the case of $\|g\left( {{x_k}} \right)\| \le {\epsilon	_{\nabla f}}+ {\epsilon_g}$, that is 
\begin{align*}
    \left\| {\nabla f\left( {{x_k}} \right)} \right\| 
    \le&
    \left\| {g\left( {{x_k}} \right)} \right\| + \left\| {\nabla f\left( {{x_k}} \right) - g\left( {{x_k}} \right)} \right\|
    \nonumber\\
    \le&
    {\epsilon_{\nabla f}} + {\epsilon_g} + {\epsilon_g} = {\epsilon_{\nabla f}} + 2{\epsilon_g}.
\end{align*}
we set $\mathcal{S}_h=\mathcal{S}_g$. 
$x\in\mathcal{S}_H$, in which $\left\| {\nabla f\left( {{x_k}} \right)} \right\| \le {\epsilon_{\nabla f}}$, satisfies such a  case\footnote{In the case of $\epsilon_{\nabla f}+2\epsilon_{g}>\|\nabla f(x_k)\|>\epsilon_{\nabla f}$, we have  $\mathcal{S}_h=\mathcal{S}_g$ such that the equality (\ref{Newton:STR:Lemma:Bound:Radius-3}) in the Appendix become 
	$\frac{{4{\epsilon_g} + \frac{3}{2}\left( {{L_H}{\Delta _k} + 2{\epsilon_B} + \frac{4}{3}{\epsilon_h}} \right){\Delta _k}}}{{\frac{1}{2}\left( {{\epsilon_{\nabla f}} - {\epsilon_g}} \right)}}$, which is smaller than equality (\ref{Newton:STR:Lemma:Bound:Radius-3}), 
	thus $\Delta_{{\rm{min1}}}$ is also satisfying such case. In this paper, in order to simplify the analysis, we consider the case $\nabla f(x)>\epsilon_{\nabla f}$ without the requirement of $\mathcal{S}_{\nabla f}=\mathcal{S}_H$.}. Thus, we give the $\Delta_{\text{min2}}$ in  (\ref{Newton:STR:Lemma:Bound:Radius-equality-main}) based on such implementation, which is key for analyses. The reason we make such an implementation is to ensure that there is a lower bound of the radius. What's more, the parameters' setting is more simple. Based on the above lemmas, we analyze the minimal radius.


\begin{lemma}\label{Newton:STR:Lemma:Bound:Radius}
	In Algorithm \ref{Newton:STR:Algorithm}, suppose Assumptions 
	\ref{Newton:Assumption:function}-
	\ref{Newton:Assumption:Bound:Variance} hold,  let $| {{\mathcal{S}_h}} | = \min 
	\{ n,\max \{ H_1/\epsilon_g,H_2/\epsilon_B 
	\} \}$, $1>r_1>0$, there 
	will be a  non-zero radium
	\begin{align}
	\label{Newton:STR:Lemma:Bound:Radius-equality-main}
    	{\Delta _\text{min}} = \min \{ {{\Delta _\text{min1}},{\Delta 	_\text{min2}}} 	\},
	\end{align}		
	where the parameters satisfy
	{\small
	\begin{align}
	\label{Newton:STR:Lemma:Bound:Radius-equality1}
	    {\Delta _\text{min1}} =& {\kappa _1}\left( {{\epsilon_{\nabla f}} - {\epsilon_g}} 
	    \right),\nonumber\\
	    {\kappa _1} =& {r_1}\min \left\{ {\frac{1}{{{\kappa _H}}},\frac{1}{{48}}\left( {1 - \eta } \right),\sqrt {\frac{1}{{12{L_H}}}\left( {1 - \eta } \right)} } \right\},\nonumber\\
	    {\Delta _\text{min2}} =& {\kappa _2}( {{\epsilon_H} - 	{\epsilon_B}} ),{\kappa _2} = {r_1}( {1 - \eta } )/(6{L_H}),\nonumber\\
	    {\epsilon_g} \le & \frac{1}{{16}}( {1 - \eta } )( {{\epsilon_{\nabla f}} - {\epsilon_g}} ),
	    {\epsilon_h} \le{\epsilon_B} \le   \frac{1}{{10}}( {1 - \eta } )( {{\epsilon_H} - {\epsilon_B}} ).\nonumber
	\end{align}
	}
	In particular, 	$\Delta_\text{min1}$ belongs to the case of $x\in 
	\mathcal{S}_{\nabla f}$ and $\Delta_\text{min2}$ belongs to the case of 
	$x\in 
	\mathcal{S}_{H}$.
\end{lemma}

\noindent\textbf{Proof of Theorem \ref{Newton:STR:Theorem:Iteration}}
\begin{proof}
	Consider two index sets: $\mathcal{S}_{\nabla f}$ and $\mathcal{S}_{H}$, we 
	separately 
	analyze the number of successful iterations based on results in Lemmas~\ref{Newton:STR:Lemma:Upperbound_mx-msx} and
	\ref{Newton:STR:Lemma:Bound:Radius}. And then add both of them to form the most successful iterations. Let  $f_\text{low}$ be the minimal value of the objective, we 
	have two kinds of successful iterations:	
	\begin{itemize}
		\item Consider the case of $\|	{\nabla f( x_k )} \| \ge {\epsilon
			_{\nabla f}}$, if $k$th iteration is successful, we have
		\begin{align*}
		&f\left( {{x_k}} \right) - f\left( {{x_k} + {s_k}} \right) 
		\ge {\eta 
		}\left( {{m_k}\left(0 \right) - {m_k}\left( { 
				{s_k}} 
			\right)} \right)\\
		\ge& \frac{1}{2}\eta ({\epsilon_{\nabla f}} \!-\! {\epsilon_g}){
		}\min \left\{ {{\Delta _k},\frac{{{\epsilon_{\nabla f}} \!-\! 
					{\epsilon_g}}}{{{\kappa _H}}}} \right\}
		\ge\frac{1}{2}{\eta}({\epsilon_{\nabla f}} - {\epsilon_g}){\Delta 
			_{\text{min}}},
		\end{align*}
		where the last two inequalities are based on Lemma 
		\ref{Newton:STR:Lemma:Upperbound_mx-msx} and Lemma 
		\ref{Newton:STR:Lemma:Bound:Radius}. Let $T_{1}$ denote 
		the number of successful iterations for $k\in \mathcal{S}_{\nabla f}$. Applying the above inequality, we can obtain 
		$
		f\left( {{x_0}} \right) - {f_\text{low}}
		\ge \frac{1}{2}{T_1}({\epsilon_{\nabla f}} - {\epsilon_g}){\eta}{\Delta 
			_\text{min}},
		$
		where $f_\text{low}$ is the lower bound of the objective.
		\item Consider the case of ${\lambda _{\text{min} }}( \nabla^2f(x_k) ) 
		\le  - 	{\epsilon_H}$, if $k$th iteration is successful, based on Lemma 
		\ref{Newton:STR:Lemma:Upperbound_mx-msx} and Lemma 
		\ref{Newton:STR:Lemma:Bound:Radius}, we have
		\begin{align*}
		    f\left( {{x_k}} \right) - f\left( {{x_k} + {s_k}} \right) 
		    &\ge 
		    {\eta }\left( {{m_k}\left(0 \right) - {m_k}\left( { {s_k}} \right)} \right)
		    \\& \ge 
		    \frac{1}{2}\eta({\epsilon_H} - {\epsilon_B})\Delta _\text{min}^2.
		\end{align*}
		Let $T_2$  denote 
		the number of successful iterations for $k\in \mathcal{S}_{H}$, we obtain 
        $
		f\left( {{x_0}} \right) - {f_\text{low}}
		\ge \frac{1}{2}{T_2}{\eta}({\epsilon_H} - {\epsilon
			_B})\Delta _{\text{min}}^2.
			$
	\end{itemize}
	Let $T_\text{suc}$ denote the number of  successful iterations, combining 
	above iteration and  $\Delta_{\text{min}}$ in Lemma \ref{Newton:STR:Lemma:Bound:Radius},  we have
	\begin{align*}
	T_\text{suc} \le& {T_1} + {T_2}
	\\
	\le& \frac{{2\left( {f\left( {{x_0}} \right) - 	{f_{{\rm{low}}}}} \right)}}{{({\epsilon_{\nabla f}} - {\epsilon_g})\eta {\Delta _\text{min}}}} + \frac{{2\left( {f\left( {{x_0}} \right) - {f_{{\rm{low}}}}} 	\right)}}{{\eta ({\epsilon_H} - {\epsilon_B}){\Delta^2_\text{min}}}}
	\\
	\le& {\kappa _3}{\rm{max}}\{ {( {{\epsilon _H} - {\epsilon _B}} )^{-1}{{( {{\epsilon _{\nabla f}} - {\epsilon _g}} )}^{ - 2}},{{( {{\epsilon _H} - {\epsilon _B}} )}^{ - 3}}} \},
	\end{align*}
	where ${\kappa _3} = 4\left( {f\left( {{x_0}} \right) - {f_\text{low}}} 
	\right){\rm{max}}\left\{ 
	{1/\left( {\eta {\kappa _1^2}} \right),1/\left( {\eta \kappa _2^2} \right)} \right\}$.

	Let $T_\text{unsuc}$ denote the number of unsuccessful iterations, we have $	{{r}_1}{\Delta_k} \le 
	{\Delta _{k + 1}}$;	Let $T_\text{suc}$ denote the number of successful iterations, we have 
	${r_2}{\Delta_k} \le {\Delta _{k + 1}}$.	Thus, we inductively deduce,
	\[{\Delta _{\text{min} }}r_2^{T_\text{suc}}r_1^{T_\text{unsuc}} \le {\Delta _{\text{max} 
	}},\]
	where $\Delta _\text{min}$ is defined in  (\ref{Newton:STR:Lemma:Bound:Radius-equality-main}) 
	and 
	${\Delta _{\text{max} }}$ is defined in Algorithm \ref{Newton:STR:Algorithm}.
	Thus, the number of unsuccessful index set is at most
	{\small
	\begin{align}
	{T_\text{unsuc}} \ge \frac{1}{{ - \log {r_1}}}\left( {\log \left( {\frac{{{\Delta_{{\text{max}}}}}}{{{\Delta_{{\text{min}}}}}}} \right) - {T_\text{suc}}\log {r_2}} \right).
	\end{align}
	}
	Combining with the successful iteration, we can obtain the total 
	iteration complexity,
	{
	\small
	\begin{align*}
	    &{T_\text{suc}} + {T_\text{unsuc}}
	    \\
	    =& {T_\text{suc}}\left({1+\log\left({\frac{{{\Delta_{{\rm{max}}}}}}{{{\Delta_{{\rm{min}}}}}}}\right)}\right)-\log\left({\frac{{{\Delta_{{\rm{max}}}}}}{{{\Delta _{{\rm{min}}}}}}} \right)\frac{1}{{\log {r_1}}}\\
	    =& \mathcal{O}( {\rm{max}}\{ {( {{\epsilon _H} - {\epsilon _B}} )^{-1}{{( {{\epsilon _{\nabla f}} - {\epsilon _g}} )}^{ - 2}},{{( {{\epsilon _H} - {\epsilon _B}} )}^{ - 3}}} \} ).
	\end{align*}
	}
\end{proof}

\subsection{Proof for Stochastic ARC}\label{SFSNewton:Appendix:SARC}



 Based on the lower bound of $p_k(0)-p_k(s_k)$ in \cite{conn2000trust}, we obtain the Lemma \ref{Newton:SARC:Lemma:UpperboundOfP0Ps}.
Note that the detailed proofs of Lemma \ref{Newton:SARC:Lemma:UpperboundOfP0Ps-2} and Lemma \ref{Newton:SARC:Lemma:bound_sigmal} are in Appendix.

\begin{lemma}\label{Newton:SARC:Lemma:UpperboundOfP0Ps} 
	Suppose that the step size $s_k$ satisfies ${p_k}\left( {{s_k}} \right) \le {p_k}\left( {s_k^C} \right)$, where $s_k^C$ is a Cauchy point, 	defined 	as
	$s_k^C =  - {\alpha _k}g\left( {{x_k}} \right),	{\alpha _k} =\mathop {\arg \min }\nolimits_{\alpha  \in {\mathbb{R}_ + 	}} 	\left\{ {{p_k}\left( x_k{ - \alpha g\left( {{x_k}} \right)} \right)} \right\}
	$, 
	for all $k\ge 0$,
	we have that 
	\begin{align*}
	p_k(0) - {p_k}( s_k)
	\ge 
	\frac{\left\| {{g_k}} \right\|}{10}\min \left\{ {{\left\| {{g_k}} \right\|}/{{\left\| {{B_k}} \right\|}},\sqrt {{{\left\| {{g_k}} \right\|}}/\sigma_k} } 	\right\}.
	\end{align*}
	Specifically, we set ${\alpha} = 2/(\| B_k	\| + \sqrt 	{{\| B_k \|}^2 + 4{\sigma _k}\| g_k \|} )$ and $s_k=-\alpha g(x_k)$ that satisfy above inequality. Furthermore, we can also obtain the upper bound of the step $\| s_k \|$, $k>0$, which satisfies 
	$\| s_k \| \le {{11}}/{4}\max \{ {{{\| 	{B({x_k})} \|}}/{{{\sigma _k}}},\sqrt {{{\| {g({x_k})} 	\|}}/{{{\sigma _k}}}} } \}.$	
\end{lemma}

\begin{lemma}\label{Newton:SARC:Lemma:UpperboundOfP0Ps-2} 
	Given the  	conditions of 	$s_k$ in 	(\ref{Newton:SARC:Assumption_s1}) 
	we have
	\begin{align*}
	p_k( 0) - {p_k}( s_k ) \ge 	\frac{{{\sigma_k}}}{6}{\| {{s_k}} \|^3}.
	\end{align*}
	Furthermore, suppose Assumptions \ref{Newton:Assumption:function} and	\ref{Newton:Assumption:approximation},  and the condition (\ref{Newton:SARC:Assumption_s3}) hold, for the norm of the gradient $\| {g( {{x_{k + 1}}} )} \| \ge {\epsilon_{\nabla f}} - 	{\epsilon_g}$, we have	\begin{align}\label{Newton:SARC:Lemma:UpperboundOfP0Ps-2-1}
	{\kappa _s}{\left\| {{s_k}} \right\|^2}\ge\left\| {g\left( {{x_{k + 1}}} \right)} \right\| ,
	\end{align}
	where 		
	\[\begin{array}{l}
	{\kappa _s} = \min \left\{ {\frac{{2{\epsilon_B} + \left( {{L_H} + {\sigma _k}} \right) + 2{\kappa _\theta }{\epsilon_g} + {\kappa _\theta }{L_{\nabla f}}}}{{\left( {1- 	{\theta _k}} \right)}},\frac{{{L_H} + {\sigma _k} + {\kappa _\theta }{L_{\nabla f}}}}{{1 - 	{\theta _k} - {\zeta _1} - {\zeta _2}}}} \right\},\\
	0<{\zeta _1},{\zeta _2} < 1,{\epsilon_B} \le {\zeta _1}\left( {{\epsilon_{\nabla f}} 	- {\epsilon_g}} \right),{\epsilon_g} \le {\zeta _2}\left( {{\epsilon_{\nabla f}} - {\epsilon_g}} \right).
	\end{array}\]
\end{lemma}


Different from the STR, we can also derive the relationship between $\|g(x_{k+1})\|$ and $\|s_k\|$. The core process of the proof is based on the cubic regularization of the Newton method \cite{nesterov2006cubic}. Such a relationship leads to the improved iteration complexity. 
Besides the lower bound of the numerator in (\ref{Newton:SARC:1-p}), we can also obtain the corresponding upper bound of the denominator, which is similar to Lemma 
\ref{Newton:STR:Lemma:Upperbound_fxs-msx}. 
Thus, we remove the proof.
\begin{lemma}\label{Newton:SARC:Lemma:Upperbound_p-h}
	Suppose Assumptions \ref{Newton:Assumption:function}, 
	\ref{Newton:Assumption:approximation}, and \ref{Newton:Assumption:Bound:Variance} hold, at 
	$k$-iteration, we have
	\begin{align*}
	&{p_k}\left( 0 \right) - {p_k}\left( {{s_k}} \right) - \left( {h\left( {{x_k}} \right) - 
		h\left( {{x_k} + {s_k}} \right)} \right)\\
	\le& 2( {\frac{\mathbb{I}{( {| {{{\cal S}_h}} | < n} )}}{{|{{{\cal S}_h}} |}}{H_1} + {\epsilon_g}} )\| {{s_k}} \| + \frac{3}{2}( {\frac{\mathbb{I}{( {| {{{\cal S}_h}} | < n} )}}{{| {{{\cal S}_h}}|}}{H_2} + {\epsilon_B}} ){\| {{s_k}} \|^2}
	+ ( {\frac{3}{2}{L_H} - \frac{1}{3}{\sigma _k}} ){\| {{s_k}} \|^3}.
	\end{align*}
	If $\mathcal{S}_h=\mathcal{S}_g$, we have $2( {\frac{\mathbb{I}{( {| {{{\cal S}_h}} | < n} )}}{{|{{{\cal S}_h}} |}}{H_1} + {\epsilon_g}} )\| {{s_k}} \| =0$.
\end{lemma}
Based on the above lemmas, we can derive the upper bound of the  adaptive parameter $\sigma$, which is used to analyze the iteration complexity. Furthermore, the parameters' settings, such as $\epsilon_g$, $\epsilon_B$,  and $\epsilon_h$ are similar 
to that of Lemma \ref{Newton:STR:Lemma:Bound:Radius}.

\begin{lemma}\label{Newton:SARC:Lemma:bound_sigmal}
In Algorithm \ref{Newton:SARC:Algorithm}, 
	suppose Assumptions  
	\ref{Newton:Assumption:function}-\ref{Newton:Assumption:Bound:Variance} hold, let $r_2>1$ and $| 	\mathcal{S}_h |$ =$ \text{min} \{ n,\text{max} \{ H_1/\epsilon_g,H_2/\epsilon_B 
	\} \}$, the parameter 	$\sigma$ is bounded by 
	\begin{align}
	\label{Newton:SARC:Lemma:bound_sigmal-1-main}
	    {\sigma _{\text{max}}} = \max  \left\{ {{\sigma _\text{max1}},{\sigma _\text{max2}}} 	\right\},	\end{align}	
	where $\sigma _\text{max1} = {\kappa _4}\frac{1}{{({\epsilon_{\nabla f}} - {\epsilon_g})}},{\sigma _{\text{max2}}} = \frac{9}{2}{r_2}{L_H},$ and 
	\begin{align}
	    {\kappa _4} =& {r_2}\max\left\{ {\kappa _H^2,\frac{{{{\left( {304\left( {3{\epsilon_B} +2{\epsilon_h}} \right)} \right)}^2}}}{{\left( {1 - \eta }\right)}},\frac{9}{2}({\epsilon_{\nabla f}}-{\epsilon_g}){L_H}}
	    \right\},\nonumber\\
	    {\epsilon_g} =& \frac{1}{{220}}\left( {1 - \eta } \right)({\epsilon_{\nabla f}} - 
	    {\epsilon_g}),
	    {\epsilon_B} = {\epsilon_h} =\frac{1}{{36}}\left( {1 - \eta } \right)\left( {{\epsilon_H} - {\epsilon_B}} \right).\nonumber
	\end{align}
	In particular, 	$\sigma_\text{min1}$ belongs to the case of $x\in 	\mathcal{S}_{\nabla f}$ and $\sigma_\text{min2}$ belongs to the case of $x\in \mathcal{S}_{H}$.	
\end{lemma}

\noindent\textbf{Proof of Theorem \ref{Newton:SARC:theorem:Iteration}}

\begin{proof}
	First of all, we consider the unsuccessful iteration. The unsuccessful iteration $T_{\text{unsuc}}$ is similar to that of STR. Based on Lemma \ref{Newton:SARC:Lemma:bound_sigmal}, that is	${\sigma _{\max }} = r_1^{{T_\text{suc}}}r_2^{{T_\text{unsuc}}}{\sigma _{\min }}.$	
	We can obtain 
	\begin{align} \label{Newton:SARC:theorem:Iteration_eq1}
	{T_{{\rm{unsuc}}}} 
	=& \frac{1}{{ - \log {r_1}}}\left( {\log \left( {\frac{{{\sigma _{{\rm{max}}}}}}{{{\sigma _{{\rm{min}}}}}}} \right) - {T_{{\rm{suc}}}}\log {r_2}} \right)	\nonumber\\
	\le& \frac{1}{{ - \log {r_1}}}\log \left( {\frac{{{\sigma _{{\rm{max}}}}}}{{{\sigma _{{\rm{min}}}}}}} \right) + {T_{{\rm{suc}}}}.
	\end{align}	
	
	Now, we consider two kinds of successful iteration complexity:
		\begin{itemize}
			\item 	For the case of $||\nabla f\left( {{x_k}} \right)|| \ge {\epsilon_{\nabla f}}$, based on Lemma 
			\ref{Newton:SARC:Lemma:UpperboundOfP0Ps}, if $k$-th iteration is successful, we obtain
			\begin{align*}
			&f\left( {{x_k}} \right) - f\left( {{x_{k + 1}}} \right) 
   \ge {\eta}\left( {p\left( 	0\right) - {p_k}\left( s_k \right)} \right)\\
			\ge& \frac{1}{10}{\eta}({\epsilon_{\nabla f}} - {\epsilon_g})\sqrt 	{\frac{{{\epsilon_{\nabla f}} - {\epsilon_g}}}{{{\sigma _{{\rm{max1}}}}}}}  
			= \frac{1}{10}{\eta}{({\epsilon_{\nabla f}} - {\epsilon_g})^2}\kappa _4^{ - 1/2},
			\end{align*}
			where inequality  follows from (\ref{Newton:SARC:Lemma:Upperbound_mx-msx:inequality1}) and the definition of  $\sigma_{\text{max1}}$ in 
			(\ref{Newton:SARC:Lemma:bound_sigmal-1-main}). Let $T_3$ be the number of  successful iterations, we obtain
			\begin{align*}
			f\left( {{x_0}} \right) - f_{\text{low}} 
			\ge \frac{1}{5}T_3{\eta}{({\epsilon_{\nabla f}} - {\epsilon	_g})^2}\kappa_4^{ - 1/2}.
			\end{align*}				
			
			\item For the case of ${\lambda _{\text{min} }}( \nabla^2f(x_k) ) 
			\le  - 	{\epsilon_H}$, based on Lemma 	\ref{Newton:SARC:Lemma:UpperboundOfP0Ps-2}, if 	$k$-th 	iteration is successful, we have
			\begin{align*}
			&f\left( {{x_k}} \right) - f\left( {{x_k} + {s_k}} \right) 
			\ge  \frac{{{\sigma _k}}}{6}\eta{\left\| {{s_k}} \right\|^3} = \frac{1}{{6\sigma _k^2}}\eta\sigma _k^3{\left\| {{s_k}} \right\|^3}\\
			& \ge - \frac{1}{{6{\sigma_\text{max2}}}}\eta{\left({\frac{{s_k^T{B_k}{s_k}}}{{{{\left\| {{s_k}} \right\|}^2}}}} \right)^3}
			 \ge  \frac{1}{{6{\sigma _\text{max2}}}}\eta{\left( {{\epsilon_H} - {\epsilon_B}} \right)^3},
			\end{align*}
			where  inequalities are based on  the condition of $s_k$ that 	satisfies 
			(\ref{Newton:SARC:Assumption_s1}),  that is 
			$
			{\sigma _k}\left\| {{s_k}} \right\| \ge  - \frac{{s_k^TB\left( {{x_k}} \right){s_k}}}{{{{\left\| {{s_k}} \right\|}^2}}}
			$, $\sigma_{\text{max2}}$ in (\ref{Newton:SARC:Lemma:bound_sigmal-1-main}), and equality (\ref{Newton:ARC:Lemma:Upperbound_fx-hsx_Lambda:inequality1}). Let $T_4$ 
			be the number of successful iterations, then we obtain	
			\[f( {{x_0}} ) - f_\text{low} \ge T_4\eta {\left( {{\epsilon_H} - {\epsilon_B}} \right)^3}/(6{\sigma _\text{max2}}).\]	
		\end{itemize}
		Thus, the total number of successful iterations is \[T_{\text{suc1}}={T_3} + {T_4} = {\kappa _5}\max 
		\{ 	{{{({\epsilon_{\nabla f}} - {\epsilon_g})}^{ - 2}},{{( {{\epsilon_H} - {\epsilon_B}} )}^{ - 3}}} \},\]
		where ${\kappa _5} = ( {f( {{x_0}} ) - {f_{{\rm{low}}}}} )\max \{ 	{5/( {\eta \kappa _4^{ - 1/2}}),6{\sigma _{{\rm{max2}}}}/\eta } \}$. 
		Combine with the successful iteration and unsuccessful iteration in (\ref{Newton:SARC:theorem:Iteration_eq1}), we can obtain the total iteration complexity,
		\begin{align*}
		{T_\text{suc1}} + {T_\text{unsuc1}} \le &  \frac{1}{{ - \log {r_1}}}\log \left( {\frac{\sigma _{{\max}}}{\sigma _{{\min}}}} \right) + 2{T_{{\rm{suc}}}}\\
		=& \mathcal{O}( {{\max}}\{ {{{( {{\epsilon _{\nabla f}} - {\epsilon _g}} )}^{ - 2}},{{( {{\epsilon _H} - {\epsilon _B}} )}^{ - 3}}} \} ).
		\end{align*}

		For the case that condition 
		(\ref{Newton:SARC:Assumption_s1})-(\ref{Newton:SARC:Assumption_s3}) and $
		g(x_k)>(\epsilon_{\nabla f}-\epsilon_g)$ hold, we have
		\begin{align*}
		f\left( {{x_k}} \right) - {f}\left( {{x_k} + {s_k}} \right) 
		\ge	\frac{{{\sigma _k}}}{6}{\left\| {{s_k}} \right\|^3}
		\ge &\frac{{{\sigma _{\text{min} }}}\kappa_s^{-\frac{3}{2}}}{6}{\left\| {g\left( {{x_k}} \right)} \right\|^{\frac{3}{2}}}\\
		\ge& \frac{{{\sigma_{\text{min}}}}\kappa_s^{-\frac{3}{2}}}{6}{\left({{\epsilon_{\nabla f}} - {\epsilon_g}} \right)^{\frac{3}{2}}},
		\end{align*}
		where  inequalities are based on Lemma \ref{Newton:SARC:Lemma:UpperboundOfP0Ps-2} and
		$\sigma_{{\text{min}}}$ in
		Algorithm \ref{Newton:SARC:Algorithm}.
		Let $T_5$ be the number of successful iterations, we obtain	
		\begin{align*}
		f({x_0}) - {f_\text{low}} \ge \frac{{\sigma _{\text{min} }}\kappa_s^{-3/2}}{6}T_5\eta {\left( {\epsilon_{\nabla 	f}} - {\epsilon_g} \right)^{3/2}}.
		\end{align*}
		The total number of successful iterations for such case is
		\[T_{\text{suc2}}={T_3} + {T_5} = {\kappa _6}\max \{ {{{({\epsilon_{\nabla f}} - {\epsilon_g})}^{ - 3/2}},{{( {{\epsilon_H} - {\epsilon_B}} )}^{ - 3}}} \},\]
		where ${\kappa _6} = ( {f( {{x_0}} ) - {f_{{\rm{low}}}}} )\max \{ {6\kappa _s^{3/2}/( {\eta {\sigma _{{\rm{min}}}}} ),6{\sigma _{{\rm{max2}}}}/\eta } \}$.
		Combine the successful iteration and unsuccessful iteration in (\ref{Newton:SARC:theorem:Iteration_eq1}), we can obtain the total iteration complexity,
		\begin{align*}
		{T_\text{suc1}} + {T_\text{unsuc1}} \le &  \frac{1}{{ - \log {r_1}}}\log \left( {\frac{{{\sigma _{{\rm{max}}}}}}{{{\sigma _{{\rm{min}}}}}}} \right) + 2{T_{{\rm{suc}}}}\\
		=& \mathcal{O}( {{\max}}\{ {{{( {{\epsilon _{\nabla f}} - {\epsilon _g}} )}^{ - 3/2}},{{( {{\epsilon _H} - {\epsilon _B}} )}^{ - 3}}} \} ).
		\end{align*}
	
\end{proof}	

\section{proof of Lemmas}
We present the detailed proof of the Lemmas used for the iteration complexities of STR and SARC.
\subsection{Lemmas for STR}
\noindent\textbf{Proof of Lemma \ref{Newton:STR:Lemma:Upperbound_mx-msx}}
\begin{proof}
	For the case $x_k\in \mathcal{S}_{\nabla f}$,
	through adding and subtracting the term $\nabla f( x_k ) $, we 
	have the lower bound of $\|g(x)\|$,
	\begin{align}
	\| {g( x_k )} \| &= \| {g( x_k ) -	 \nabla f( x_k ) + \nabla f( x_k )} 
	\|\nonumber\\
	&\ge \| {\nabla f( x_k )} \| -  \| {g( x_k	) - \nabla f( x_k )} \|
	\ge {\epsilon_{\nabla f}} - {\epsilon_g},\nonumber
	\end{align}
	where the last inequality is based on the approximation of $\nabla f(x_k)$ in Assumption 
	\ref{Newton:Assumption:approximation}. Following the lower bound 
	on 	the decrease of the proximal quadratic 
	function ${m_k}( {{s}} )$ from  
	(4.20) in \cite{nocedal2006numerical}, we have
	\begin{align*}
	m_k(0) - {m_k}(s_k)
	&
	\ge \frac{1}{2}\| g( x_k )
	\|\min 	\left\{ {\Delta _k},\frac{{\| g( x_k ) \|}}{\| B( x_k ) \|} \right\}
	\\&
	\ge \frac{\epsilon_{\nabla f} - {\epsilon_g}}{2}\min 
	\left\{ {\Delta _k},\frac{{\epsilon_{\nabla f}} - {\epsilon_g}}{\kappa _H} \right\},
	\end{align*}
	where the last inequality is from the above inequality and the bound 
	of 	${\left\| {B({x_k})} \right\|}$ in Assumption 
	\ref{Newton:Assumption:bound}.

	For the case $x_k\in \mathcal{S}_H$,	through adding and subtracting the term $\nabla ^2f(x_k ) $, we have
	\begin{align}
	\frac{{s_k^{T} B({x_k}){s_k}}}{{\left\| {{s_k}} \right\|^2}} =& 
	\frac{{s_k^T(B({x_k}) - \nabla^2f(x_k) + \nabla^2f(x_k)){s_k}}}{{\left\| {{s_k}} \right\|^2}}\nonumber\\
	=& \frac{{s_k^T(B({x_k}) - \nabla^2f(x_k)){s_k}}}{{\left\| {{s_k}} \right\|^2}} + 
	\frac{{s_k^T\nabla^2f(x_k){s_k}}}{{\left\| {{s_k}} \right\|^2}}\nonumber\\
	\le& \left\| { B({x_k}) - \nabla^2f(x_k)} \right\| + 
	\frac{{s_k^T\nabla^2f(x_k){s_k}}}{{\left\| {{s_k}} \right\|^2}}\nonumber\\
	\label{Newton:STR:Lemma:Upperbound_fx-hsx_Lambda:inequality1}
	\le& v_0{\epsilon_B} - {v_0}\epsilon_H
	=  - {v_0}( {{\epsilon_H} - {\epsilon_B}}),
	\end{align}
	where the second inequality is based on the Assumption
	\eqref{Newton:Assumption:approximation} and the condition of ${s_k^T\nabla^2f(x_k){s_k}}<v_{0}\lambda_\text{min}(\nabla^2f(x_k))
	\| s_k \|^2$. Using the definition of 
	$m_k(s)$ in (\ref{Newton:Definition_m}), we have
	\begin{align*}
	m_k( 0) - {m_k}( {{s_k}} ) 
	&=  - \left\langle {g({x_k}),{s_k}} \right\rangle  - 
	\frac{1}{2}s_k^TB\left( {{x_k}} \right){s_k}
	\\
	&\mathop  \ge \limits^{{\scriptsize \textcircled{\tiny{1}}}}  - 
	\frac{1}{2}s_k^TB\left( {{x_k}} \right){s_k}
	\mathop  \ge \limits^{{\scriptsize \textcircled{\tiny{2}}}} 
	\frac{1}{2}\left( {{\epsilon_H} - {\epsilon_B}} 
	\right)v_0{\left\| {{s_k}} \right\|^2}
	\\&
	\mathop =\limits^{{\scriptsize \textcircled{\tiny{3}}}} \frac{1}{2} 
	( {{\epsilon_H} - {\epsilon_B}} )v_0\Delta _k^2,
	\end{align*}
	where inequality $\scriptsize \textcircled{\tiny{1}}$ is based on  $\langle 	{g({x_k}),{s_k}} \rangle \le 0$, inequality $\scriptsize 
	\textcircled{\tiny{2}}$ follows from 
	(\ref{Newton:STR:Lemma:Upperbound_fx-hsx_Lambda:inequality1}), and 
	inequality $\scriptsize 
	\textcircled{\tiny{3}}$ is based on $\left\| {{s_k}} \right\| = {\Delta 
		_k}$.
\end{proof}

\noindent\textbf{Proof of Lemma \ref{Newton:STR:Lemma:Upperbound_fxs-msx}}

\begin{proof}
	Consider the Taylor expansion for $h(x_k+s_k)$ at $x_k$,
	\[h\left( {{x_k} + {s_k}} \right) = h\left( {{x_k}} \right) + \left\langle 
	{{s_k},\nabla h\left( {{x_k}} \right)} \right\rangle  + 
	\frac{1}{2}s_k^T\nabla^2h\left( {{\xi _k}} \right){s_k},\]
	where ${\xi  _k} \in \left[ {{x_k},{x_k} + {s_k}} \right]$.	
	Based on the definition of $m_k(s)$ in 
	(\ref{Newton:Definition_m}) and the Taylor expansion of the  
	function $ h(x_{k}+s_k)$ at $x_k$ above, we have 	
	\begin{align}
	&m_k(0) - m_k(s_k ) - (h(x_k) - h(x_k + s_k ) )
	\nonumber\\
	=	& \!-\! \langle g(x_k),s_k\rangle  \!-\! \frac{1}{2}s_k^TB(x_k)s_k \!+\! ( \langle \nabla h(x_k),s_k \rangle  \!+\! \frac{1}{2}s_k^T\nabla ^2h(\xi_k)s_k )
	\nonumber\\
	\mathop  = \limits^{{\scriptsize \textcircled{\tiny{1}}}}
	& \left\langle {\nabla h\left( {{x_k}} \right) - g({x_k}),{s_k}} \right\rangle  + 	\frac{1}{2}s_k^T\left( {{\nabla ^2}h\left( {{\xi _k}} \right) - B({x_k})} \right){s_k}\nonumber\\
	=& \left\langle {\nabla h\left( {{x_k}} \right) - \nabla f\left( {{x_k}} \right) + \nabla f\left( {{x_k}} \right) - g({x_k}),{s_k}} \right\rangle 
	\!+\! \frac{1}{2}s_k^T(\nabla ^2 h (\xi _k) 
	\!-\! \nabla ^2 f(\xi _k) 
	\!+\! \nabla ^2f(\xi_k)
	\!-\! {\nabla ^2}f(x_k) \!+\! {\nabla ^2}f(x_k) \!-\! B(x_k)){s_k}
	\nonumber\\
	\mathop  \le \limits^{{\scriptsize \textcircled{\tiny{2}}}}
	& 2\left( {\left\| {\nabla h\left( {{x_k}} \right) - \nabla f\left( {{x_k}} \right)} \right\| + \left\| {\nabla f\left( {{x_k}} \right) - g({x_k})} \right\|} \right)\left\| 	{{s_k}} \right\| \nonumber\\
	+& \frac{3}{2}(\| \nabla ^2h(\xi _k ) - \nabla ^2f( \xi _k ) \| + \| \nabla ^2f( \xi _k )-\nabla ^2f( x_k ) \| 
	+ \| \nabla ^2f( x_k ) - B(x_k) \| )\| s_k \|^2
	\nonumber\\
	\mathop  \le \limits^{{\scriptsize \textcircled{\tiny{3}}}}
	& 2\left( {\frac{{\mathbb{I}\left( {\left| {{\mathcal{S}_h}} \right| < n} \right)}}{{\left| {{\mathcal{S}_h}} \right|}}H_1 + {\epsilon_g}} \right)\left\| {{s_k}} \right\|
	 + \frac{3}{2}\left( {\frac{{\mathbb{I}\left( {\left| 	{{\mathcal{S}_h}} \right| < n} \right)}}{{\left| {{\mathcal{S}_h}} \right|}}{H_2} + L_H\left\| {{s_k}} \right\| + 	{\epsilon_B}} \right){\left\| {{s_k}} \right\|^2}\nonumber\\
	\mathop  \le \limits^{{\scriptsize \textcircled{\tiny{4}}}}
	& 2\left( {\frac{{\mathbb{I}\left( {\left| {{\mathcal{S}_h}} \right| < n} \right)}}{{\left| {{\mathcal{S}_h}} \right|}}H_1 + 	{\epsilon_g}} \right){\Delta _k} 
	+ \frac{3}{2}\left( {\frac{{\mathbb{I}\left( {\left| {{\mathcal{S}_h}} \right| < n} 	\right)}}{{\left| {{\mathcal{S}_h}} \right|}}{H_2} + L_H{\Delta _k} + 	{\epsilon_B}} \right)\Delta _k^2\nonumber,
	\end{align}
	where inequality $\scriptsize \textcircled{\tiny{2}}$ follows from the 	Holder's inequality, inequality $\scriptsize \textcircled{\tiny{3}}$  is based on the approximation of $\nabla f(x)$ 	and $\nabla^2f(x)$ in 
	Assumption \ref{Newton:Assumption:approximation},
	Assumption \ref{Newton:Assumption:Bound:Variance}, Lemma \ref{Newton:Appendix:Tool:RandomSubset}, and the Lipschitz 	continuous 	of Hessian matrix of $f(x)$ in Assumption  \ref{Newton:Assumption:function}, inequality $\scriptsize \textcircled{\tiny{4}}$ follows from the constraint condition as in the 	objective (\ref{Newton:STR:Objective}).
	Furthermore, for the case of $\mathcal{S}_h=\mathcal{S}_g$, the first term of $\scriptsize \textcircled{\tiny{1}}$ is equal to zero, then we have
	\begin{align*}
	&{m_k}\left( 0 \right) - {m_k}\left( {{s_k}} \right) - \left( {h\left( {{x_k}} \right) - h\left( {{x_k} + {s_k}} \right)} \right)
	\\
	\le&\frac{3}{2}\left( {\frac{{\mathbb{I}\left( {\left| {{\mathcal{S}_h}} \right| < n} \right)}}{{\left| {{\mathcal{S}_h}} \right|}}{H_2} + {\Delta _k} + {\epsilon	_B}} \right)\Delta 	_k^2.
	\end{align*}	
\end{proof}

\noindent\textbf{Proof of Lemma \ref{Newton:STR:Lemma:Bound:Radius}}


\begin{proof}	
	By setting $\frac{{\mathbb{I}\left( {\left| {{\mathcal{S}_h}} \right| < n} \right)}}{{\left| {{\mathcal{S}_h}} 	\right|}}{H_1} 
	\le {\epsilon_g}$, 
	and $\frac{{\mathbb{I}\left( {\left| {{\mathcal{S}_h}} \right| < n} \right)}}{{\left| {{\mathcal{S}_h}} \right|}}{H_2} 
	\le {\epsilon_B}$,  we consider two cases:
	
	For the case $||\nabla f(x)||>\epsilon_{\nabla f(x)}$, 	we assume that 
	${\Delta _k} \le \frac{{{\epsilon_{\nabla f}} - {\epsilon
				_g}}}{{{\kappa _H}}}$, which is used for Lemma 
	\ref{Newton:STR:Lemma:Upperbound_mx-msx}. Combine with (\ref{Newton:STR:1-p}) and
	the results in Lemma 
	\ref{Newton:STR:Lemma:Upperbound_mx-msx}, Lemma 
	\ref{Newton:STR:Lemma:Upperbound_fxs-msx}, and $\|s_k\|^2<\Delta_k^2$, 
	we have
	\begin{align}
	&1 - \rho 
	\nonumber\\
	\le	& \frac{{2( {\frac{{\mathbb{I}\left( {\left| {{S_h}} \right| < n} \right)}}{{\left| {{S_h}}\right|}}{H_1} \!+\! {\epsilon_g}} ){\Delta _k} \!+\! \frac{3}{2}( {\frac{{\mathbb{I}\left( 	{\left| {{S_h}} \right| < n} \right)}}{{\left| {{S_h}} \right|}}{H_2} \!+\! L_H{\Delta _k} \!+\! {\epsilon_B}} )\Delta _k^2 \!+\! 2{\epsilon_h}{\Delta^2 _k}}}{{\frac{1}{2}( {{\epsilon_{\nabla f}} - {\epsilon_g}} ){\Delta _k}}}\nonumber\\
	\label{Newton:STR:Lemma:Bound:Radius-3}
	=&\frac{{4{\epsilon_g} + \frac{3}{2}\left( {{L_H}{\Delta _k} + 2{\epsilon_B} + \frac{4}{3}{\epsilon_h}} \right){\Delta _k}}}{{\frac{1}{2}\left( {{\epsilon_{\nabla f}} - {\epsilon_g}} \right)}}.
	\end{align}
	In order to have the lower bound radius $\Delta_k$ such that $1 - \rho  \le1-\eta$, we 
	consider 
	the parameters' setting:
	\begin{itemize}
		\item For the first term, 
		we define
		${\epsilon_g} \le \frac{1}{{16}}\left( {1 - \eta } \right)\left( {{\epsilon_{\nabla f}} - {\epsilon_g}} \right)$.
		\item For the second term $\frac{3}{2}\left( {L_H{\Delta _k} + 2{\epsilon_B}+ 4{\epsilon_h}/3} \right){\Delta 	_k} \le \frac{1}{4}\left( {1 - \eta } \right)\left( {{\epsilon_{\nabla f}} - {\epsilon_g}} \right)$, as $\epsilon_B<1,\epsilon_h<1,\epsilon_{\nabla f}-\epsilon_g<1$,
		Thus, as long as 
		\begin{align}
		\label{Newton:STR:Lemma:Bound:Radius-1}
		{\Delta _k} \le \left( {{\epsilon_{\nabla f}} - {\epsilon_g}} \right)\min 
		\left\{ 
		{\frac{1}{{48}}\left( {1 - \eta } \right),\sqrt {\frac{1}{{12{L_H}}}\left( {1 - \eta } 	\right)} } \right\},
		\end{align}
		we can obtain that $1 - \rho  \le1-\eta$. 	
	\end{itemize}
	At the $k$-iteration, when the radius $\Delta_k$ satisfies the above condition, the update is a successful iteration, as in Algorithm \ref{Newton:STR:Algorithm}, the radius $\Delta_k$ will increase by 
	a factor $r_2$. 
	
	For the case ${\lambda _{\text{min} }}( {\nabla^2f(x_k)} ) \le  - 
	{\epsilon_H}$, based on the results in Lemma 
	\ref{Newton:STR:Lemma:Upperbound_mx-msx} and Lemma 
	\ref{Newton:STR:Lemma:Upperbound_fxs-msx}, we have
	\begin{align*}
	1 - \rho  \le \frac{{\frac{3}{2}\left( {{L_H}{\Delta _k} + 2{\epsilon_B} + 	\frac{4}{3}{\epsilon_h}} \right){\Delta _k}}}{{\frac{1}{2}\left( {{\epsilon_H} - {\epsilon_B}} \right){\Delta _k}}}.
	\end{align*}
	In order to have the lower bound radius $\Delta_k$ such that $1 - \rho  \le1-\eta$, we 
	consider 
	the parameters' setting:
	\begin{align}\label{Newton:STR:Lemma:Bound:Radius-2}
	{\epsilon_h}\le{\epsilon_B} \le   \frac{1}{{20}}( {1 - \eta } )( 
	{{\epsilon_H} - {\epsilon_B}}),{\Delta _k} \le \frac{1}{{6{L_H}}}( {1 - 
		\eta } )( {{\epsilon_H} - {\epsilon_B}} ).
	\end{align}
	Based on the above analysis and combine the assumption bound of $\Delta$ at the beginning, there exists a minimal radius ${\Delta _{{\rm{min}}}} = \min \{ {{\epsilon_{\nabla f}} - {\epsilon_g},{\epsilon_H} - {\epsilon_B}} \}{\kappa_1}$\footnote{Note that $\epsilon_{\nabla f}>\epsilon_g$ and $\epsilon_H > \epsilon_B$ by default} and 
	\[{\kappa _1} = {r_1}\min \{ {\frac{1}{{{\kappa 	_H}}},\frac{1}{{48}}( {1 - \eta } ),\sqrt 
		{\frac{1}{{12{L_H}}}( 	{1 - \eta } )} ,\frac{1}{{10{L_H}}}( {1 - \eta } )} \},\]
	where $0<r_1<1$. (multiply $r_1$ is due to the fact that 
	(\ref{Newton:STR:Lemma:Bound:Radius-1}) 
	and 
	(\ref{Newton:STR:Lemma:Bound:Radius-2})
	plus a small constant may lead to a successful iteration such that   $\Delta_k$ will be decreased by 
	a factor $r_1$.)					
\end{proof}

\subsection{Lemmas for SARC}

\noindent\textbf{Proof of Lemma \ref{Newton:SARC:Lemma:UpperboundOfP0Ps-2}}
\begin{proof}
	Based on the definition of $p_k(s)$ in (\ref{Newton:SARC:definition_P}), we have
	\begin{align*}
	&p_k( 0) - {p_k}( s_k )
	\\
	=&  - \left\langle 	{g\left( {{x_k}} \right),{s_k}} \right\rangle  - \frac{1}{2}s_k^TB\left( {{x_k}} \right){s_k} - \frac{{{\sigma _k}}}{3}{\left\| {{s_k}} \right\|^3}
	\\
	=& \underbrace { - \left\langle {g\left( {{x_k}} \right),{s_k}} \right\rangle  - s_k^TB\left( {{x_k}} \right){s_k} - {{\left\| {{s_k}} \right\|}^3}}_{ = 0} 
	+ \frac{1}{2}s_k^TB\left( {{x_k}} 	\right){s_k} 
	+ 	\frac{{2{\sigma _k}}}{3}{\left\| {{s_k}} \right\|^3}\\
	\mathop  = \limits^{{\scriptsize 	\textcircled{\tiny{1}}}}& 
	\frac{1}{2}s_k^TB\left( {{x_k}} \right){s_k} + 	\frac{{2{\sigma _k}}}{3}{\left\| {{s_k}} \right\|^3}
	\mathop  \ge \limits^{{\scriptsize \textcircled{\tiny{2}}}} 
	\frac{{{\sigma 	_k}}}{6}{\left\| {{s_k}} \right\|^3},
	\end{align*}
	where ${\scriptsize \textcircled{\tiny{1}}}$ and ${\scriptsize 	\textcircled{\tiny{2}}}$ follow from 
	(\ref{Newton:SARC:Assumption_s1}).
	
	Consider the lower bound of $\|s_k\|$: Firstly, based on the definition of $m_k(s)$, for simplicity, we use ${g_k} = 
	g\left( {{x_k}} \right)$ and ${B_k} = B\left( {{x_k}} \right)$ instead, we 	have
	\begin{align*}
	&\left\| {g\left( {{x_{k + 1}}} \right)} \right\|
	\\ =& \left\| {g\left( {{x_{k + 1}}} \right) - \nabla {p_k}\left( s \right) + \nabla {p_k}\left( s \right)} 	\right\|
	\\
	\mathop\le\limits^{{\scriptsize \textcircled{\tiny{1}}}}& \| g( {{x_k}} ) \!+\! \int_0^1 {B( {{x_k} \!+\! \tau {s_k}} 	){s_k}d\tau }  \!-\! ( g( {{x_k}} ) \!+\! B( {{x_k}} ){s_k} 
	 \!+\! \sigma _k\| {{s_k}} \|{s_k}) \| 
	+ \| {\nabla {p_k}( {{s_k}} )} \|
	\\
	\mathop\le\limits^{{\scriptsize \textcircled{\tiny{2}}}}& \left\| {\int_0^1 {\left( {B\left( {{x_k} + \tau {s_k}} \right) - H\left( {{x_k} + \tau {s_k}} \right)} \right){s_k}d\tau } } \right\| 
	+ \left\| 
	{H\left( {{x_k} + \tau {s_k}} \right)s - H\left( {{x_k}} \right){s_k}} \right\|+ \left\| {H\left( {{x_k}} \right){s_k} - {B_k}{s_k}} \right\| + {\sigma _k}{\left\| 
		{{s_k}} \right\|^2} + \left\| {\nabla {p_k}\left( {{s_k}} \right)} 	\right\|\\
	\mathop\le\limits^{{\scriptsize \textcircled{\tiny{3}}}}& 
	{\epsilon_B}\left\| {{s_k}} \right\| + {L_H}{\left\| {{s_k}} \right\|^2} + 	{\epsilon_B}\left\| {{s_k}} \right\| + {\sigma _k}{\left\| {{s_k}} \right\|^2} + 	{\theta _k}\left\| {g\left( {{x_k}} \right)} \right\|\\
	=& 2{\epsilon_B}\left\| {{s_k}} \right\| + \left( {{L_H} + 	{\sigma _k}} \right){\left\| {{s_k}} \right\|^2} + {\theta _k}\left\| {g\left( {{x_k}} \right)} \right\|,
	\end{align*}	
	where inequality ${\scriptsize 
		\textcircled{\tiny{1}}}$ is based on the triangle inequality and  the Taylor expansion 
	of 	$g(x)$,	equality 	${\scriptsize 	\textcircled{\tiny{2}}}$ is obtained by adding and 
	subtracting the term of 	${H\left( {{x_k}} \right){s_k}}$ and ${H\left( {{x_k+\tau s_k}} 
		\right){s_k}}$, and triangle inequality; equality 
	${\scriptsize 	\textcircled{\tiny{3}}}$ follows from Assumption 
	\ref{Newton:Assumption:function}, \ref{Newton:Assumption:approximation} and the condition 
	in 	(\ref{Newton:SARC:Assumption_s3}). Secondly, consider $g(x_k)$, we have
	\begin{align*}
	\left\| {g\left( {{x_k}} \right)} \right\|
	=& \| g\left( {{x_k}} \right) - \nabla f\left( {{x_k}} \right) + 	\nabla f\left( {{x_k}} 	\right) - \nabla f\left( {{x_k} + {s_k}} \right) + \nabla f\left( {{x_k} + {s_k}} \right) 
	- g\left( {{x_{k + 1}}} \right) + g\left( {{x_{k + 1}}} \right) \|\\
	\le& \left\| {g\left( {{x_k}} \right) - \nabla f\left( {{x_k}} \right)} \right\| + \left\| 	{\nabla f\left( {{x_k}} \right) - \nabla f\left( {{x_k} + {s_k}} \right)} \right\| + \left\| 	{\nabla f\left( {{x_k} + {s_k}} \right) - g\left( {{x_{k + 1}}} \right)} 	\right\| + \left\| 	{g\left( {{x_{k + 1}}} \right)} \right\|\\
	\le& 2{\epsilon_g} + {L_{\nabla f}}\left\| {{s_k}} \right\| + \left\| 	{g\left( {{x_{k + 	1}}} \right)} \right\|,
	\end{align*}
	where the first and second inequalities are based on the triangle inequality, Lipschitz continuity of gradient ${\nabla f\left( {{x_k}} \right)}$ in Assumptions
	\ref{Newton:Assumption:function} and 
	\ref{Newton:Assumption:approximation}.
	
	Finally, replace the term $\left\| {\nabla g\left( {{x_k}} \right)} \right\|$, 
	we have
	\begin{align*}
	&\left( {1 - {\theta _k}} \right)\left\| {g\left( {{x_{k + 1}}} \right)} \right\| 
	\le  
	2{\epsilon_B}\left\| {{s_k}} \right\| + \left( {{L_H} + {\sigma _k}} \right){\left\| {{s_k}} \right\|^2} + 2{\theta _k}{\epsilon_g} + {\theta _k}{L_{\nabla f}}\left\| {{s_k}} \right\|.
	\end{align*}
	Considering the definition of ${\theta _k} 
	\le {\zeta _\theta }\min \left\{ {1,\left\| {{s_k}} \right\|} 
	\right\},{\zeta _\theta } < 1$, we analyze the bound from different rang of
	$\|s_k\|$:
	\begin{itemize}
		\item For the case of $\left\| {{s_k}} \right\| \ge 1$, we have
		\begin{align*}
		&\left( {1 - {\theta _k}} \right)\left\| {g\left( {{x_{k + 1}}} \right)} \right\|
			 \le 2{\epsilon_B}{\left\| {{s_k}} \right\|^2} + \left( {{L_H} + {\sigma _k}} \right){\left\| {{s_k}} \right\|^2} + 2{\kappa _\theta }{\left\| {{s_k}} 	\right\|^2}{\epsilon_g} + {\kappa _\theta }{L_{\nabla f}}{\left\| 
			{{s_k}} \right\|^2}.
		\end{align*}
		\item For the case of $\left\| {{s_k}} \right\| \le 1$, based 
		on the 	assumption on 
		${\epsilon_g}$ and ${\epsilon_g}$, that is 
		\begin{align*}
		{\epsilon_B} \le {\zeta _1}\left( {{\epsilon_{\nabla f}} - {\epsilon_g}} \right) \le {\zeta _1}\left\| {g\left( {{x_{k + 1}}} \right)} \right\|,
		\\	{\epsilon_g} \le {\zeta _2}\left( {{\epsilon_{\nabla f}} - {\epsilon_g}} \right) \le {\zeta _2}\left\| {g\left( {{x_{k + 1}}} \right)} \right\|,
		\end{align*}
		where ${\zeta _1},{\zeta _2} < 1$. We have
		\begin{align*}
		&\left( {1 - {\theta _k}} \right)\left\| {g\left( {{x_{k + 1}}} 	\right)} \right\| 
		\\
		\le &	2{\epsilon_B} + \left( {{L_H} + {\sigma _k}} \right){\left\| {{s_k}} \right\|^2} + 
		2{\kappa _\theta }{\epsilon_g} + {\kappa _\theta }{L_{\nabla f}}{\left\| {{s_k}} \right\|^2}
		\\
		\le& \left( {2{\zeta _1} + 2{\kappa _\theta }{\zeta _2}} \right)\left\| {g\left( {{x_{k + 1}}} \right)} \right\| + \left( {{L_H} + {\sigma _k} + {\kappa _\theta }{L_{\nabla f}}} \right){\left\| {{s_k}} \right\|^2}.
		\end{align*}
	\end{itemize}	
	Thus, in all, we can obtain $\left\| {g\left( {{x_{k + 1}}} \right)} \right\| \le {\kappa_s}{\left\| {{s_k}} \right\|^2}$ and 
	\[{\kappa _s} 
	\!=\! \min \left\{ {\frac{{2{\epsilon_B} \!+\! \left( {{L_H} \!+\! {\sigma _k}} \right) \!+\! 2{\kappa _\theta }{\epsilon_g} \!+\! {\kappa _\theta 	}{L_{\nabla f}}}}{{\left( {1 \!-\! 	{\theta _k}} \right)}},\frac{{{L_H} \!+\! {\sigma _k} \!+\! {\kappa _\theta }{L_{\nabla f}}}}{{1 \!-\! {\theta _k} 	\!-\! {\zeta _1} \!-\! {\zeta _2}}}} \right\}.\]
\end{proof}

\noindent\textbf{Proof of Lemma {\ref{Newton:SARC:Lemma:bound_sigmal}}}
\begin{proof}	
	We assume that ${\sigma _k} \ge \frac{9}{2}{L_H}$ and ${\sigma _k} \ge \frac{{\kappa _H^2}}{{({\epsilon_{\nabla f}} - {\epsilon_g})}}$, which are used for Lemma \ref{Newton:SARC:Lemma:UpperboundOfP0Ps} and Lemma \ref{Newton:SARC:Lemma:Upperbound_p-h}. 	
	By setting $\frac{{\mathbb{I}\left( {\left| {{\mathcal{S}_h}} \right| < n} 	\right)}}{{\left| 	{{\mathcal{S}_h}} 	\right|}}{H_1} \le {\epsilon_g}$, and $\frac{{\mathbb{I}\left( {\left| 	{{\mathcal{S}_h}} 	\right| < n} \right)}}{{\left| 	{{\mathcal{S}_h}} \right|}}{H_2} \le {\epsilon_B}$,  we consider two cases:
	
	For the case $\nabla f\left( {{x_k}} \right) \ge {\epsilon_{\nabla f}}$: Firstly, we 
	consider $g(x_k)$. Through adding and subtracting the term $\nabla f( x_k ) $, we 
	have the lower bound of $\|g(x)\|$,
	\begin{align}
	\| {g( x_k )} \| &= \| {g( x_k ) -	 \nabla f( x_k ) + \nabla f( x_k )} 
	\|\nonumber\\\label{Newton:SARC:Lemma:Upperbound_mx-msx:inequality1}
	&\ge \| {\nabla f( x_k )} \| -  \| {g( x_k ) - \nabla f( x_k )} \|
	\nonumber\\
	&\ge {\epsilon_{\nabla f}} - {\epsilon_g},
	\end{align}
	where the last inequality is based on the approximation of $\nabla f(x_k)$ in Assumption 
	\ref{Newton:Assumption:approximation}. Secondly, because of 
	\[{\sigma _k} \ge \frac{{\kappa _H^2}}{{({\epsilon_{\nabla f}} - {\epsilon_g})}} 
	\ge 
	\frac{{{{\left\| {B({x_k})} \right\|}^2}}}{{\left\| {g({x_k})} \right\|}} \Rightarrow 
	\frac{{\left\| {B({x_k})} \right\|}}{{{\sigma _k}}} \le \sqrt {\frac{{\left\| {g({x_k})} 
				\right\|}}{{{\sigma _k}}}}, \]
	combing with the upper bound of $s_k$ in Lemma \ref{Newton:SARC:Lemma:UpperboundOfP0Ps}, we 
	have 
	$\left\| 	{{s_k}} \right\| \le \frac{{11}}{4}\sqrt {\left\| {g({x_k})} \right\|/{\sigma _k}}$.	
	Finally, based on equality (\ref{Newton:SARC:1-p}), Lemma 
	\ref{Newton:SARC:Lemma:UpperboundOfP0Ps}, and Lemma \ref{Newton:SARC:Lemma:Upperbound_p-h},
	we 	have
	\begin{align*}
	1 - \rho  
	\le& 10\frac{{4{\epsilon_g}\left\| {{s_k}} \right\| +3{\epsilon_B}{{\left\| 	{{s_k}} \right\|}^2}+2{\epsilon_h}\|s_k\|^2}}{{\left\| {{g_k}} \right\|\text{min}\left\{ {{\left\| {{g_k}} \right\|}/\left\| {{B_k}} \right\|,\sqrt {\left\| {{g_k}} \right\|/{\sigma _k}} } \right\}}}.
	\end{align*}
	In order to ensure that there exists a lower bound of $\sigma$ such that satisfying $1-\rho\le 	1-\eta$, we consider the setting of the $\sigma$. Combing with the upper bound of $s_k$ in Lemma \ref{Newton:SARC:Lemma:UpperboundOfP0Ps}, if ${\epsilon_g} = 
	\frac{1}{{220}}\left( {1 - \eta }\right)({\epsilon_{\nabla f}} - {\epsilon_g}),{\sigma _k}\ge\frac{1}{{{\epsilon_{\nabla f}} - {\epsilon	_g}}}\frac{{{{\left( {304\left( {3{\epsilon_B} + 2{\epsilon_h}} \right)} \right)}^2}}}{{\left( {1 - \eta } \right)}}$, we have
	\begin{align*}
	&10\frac{{4{\epsilon_g}\left\| {{s_k}} \right\| + \left( {3{\epsilon_B} + 	2{\epsilon_h}} \right){{\left\| {{s_k}} \right\|}^2}}}{{\left\| {{g_k}} \right\|{\rm{min}}\left\{ {\left\| {{g_k}} \right\|/\left\| {{B_k}} \right\|,\sqrt 
	{\left\| {{g_k}} \right\|/{\sigma _k}} } \right\}}}\\ 
	\le& \frac{{110{\epsilon_g}}}{{\left\| {{g_k}} \right\|}} + \frac{{76\left( {3{\epsilon_B} + 2{\epsilon_h}} \right)\sqrt {\left\| {g({x_k})} \right\|/{\sigma _k}} }}{{\left\| {{g_k}} \right\|}} \\
	=& \frac{{110{\epsilon_g}}}{{\left\| 	{{g_k}} \right\|}} + \frac{{76\left( {3{\epsilon_B} + 2{\epsilon_h}} \right)}}{{\sqrt {\left\| {{g_k}} \right\|{\sigma _k}} }}\\
	\le& \frac{{110{\epsilon_g}}}{{({\epsilon_{\nabla f}} - {\epsilon_g})}} + \frac{{76\left( {3{\epsilon_B} + 2{\epsilon_h}} \right)}}{{\sqrt {({\epsilon_{\nabla f}} - {\epsilon_g}){\sigma _k}} }} \le \frac{1}{2}\left( {1 - \eta } \right).
	\end{align*}
	Thus, we can see that if the parameter
	$\sigma _{\rm{max1}} ={r_2} \frac{1}{({\epsilon_{\nabla f}} -\epsilon_g)}\text{max}\{ {\kappa_H^2,\frac{{{{\left( {304\left( {3{\epsilon_B} + 2{\epsilon_h}} \right)}\right)}^2}}}{{\left( {1 - \eta } \right)}},\frac{9}{2}({\epsilon_{\nabla f}} - {\epsilon_g}){L_H}} \}$, $r_2>1$, 
	we can obtain that 	$1 - \rho  \le1-\eta$. 
	
	For the case $ \lambda_\text{min} ({\nabla^2f(x )} ) \le - {\epsilon_H }$: 
	
	Firstly, based on the  Rayleigh quotient 
	\cite{conn2000trust} that if ${H\left( x \right)}$ is symmetric and the vector $s \ne 0 $, 
	then, we have
	\begin{align}
	\frac{{s_k^TB\left( {{x_k}} \right){s_k}}}{{{{\left\| {{s_k}} \right\|}^2}}} &= \frac{{s_k^T\left( {B\left( {{x_k}} 
	\right) - {\nabla ^2}f\left( {{x_k}} \right) + {\nabla ^2}f\left( {{x_k}} \right)} 	\right){s_k}}}{{{{\left\| {{s_k}} \right\|}^2}}}\nonumber\\
	&\mathop  \le \limits^{\scriptsize \textcircled{\tiny{1}}} 
	\frac{{s_k^T{\nabla ^2}f\left( {{x_k}} \right){s_k}}}{{{{\left\| {{s_k}} \right\|}^2}}} + 
	\left\| {{\nabla ^2}f\left( {{x_k}} \right) - B\left( {{x_k}} \right)} \right\|\nonumber\\
	\label{Newton:ARC:Lemma:Upperbound_fx-hsx_Lambda:inequality1}
	&\mathop  \le \limits^{\scriptsize \textcircled{\tiny{2}}} {\lambda _{\text{min}  }}\left( {{\nabla ^2}f\left( {{x_k}} \right)} \right) + {\epsilon_B}	\le  - 	{\epsilon_H } + {\epsilon_B},
	\end{align}
	where ${\scriptsize \textcircled{\tiny{1}}}$ is based on the triangle 	inequality, 
	${\scriptsize \textcircled{\tiny{2}}}$ follows from the  Rayleigh quotient 
	\cite{conn2000trust}  and approximation Assumption \ref{Newton:Assumption:approximation}.
	
	Secondly, based on 
	$s_k^T{B_k}{s_k} + {\sigma _k}{\left\| {{s_k}} \right\|^3} \ge 0$, we have 
	\begin{align*}
	{\sigma 
		_k}\left\| 	{{s_k}} \right\| \ge  - \frac{{s_k^T{B_k}{s_k}}}{{{{\left\| {{s_k}} \right\|}^2}}} \ge 
	\left( 	{{\epsilon_H} - {\epsilon_B}} \right).
	\end{align*}
	
	Thirdly, based on equality 	(\ref{Newton:SARC:1-p}),   Lemma 
	\ref{Newton:SARC:Lemma:UpperboundOfP0Ps}, and Lemma \ref{Newton:SARC:Lemma:Upperbound_p-h}, we have
	\begin{align*}
	1 - \rho  
	\le& \frac{{2{\epsilon_h}\|s_k\|^2}}{{\frac{{{\sigma _k}}}{6}{{\left\| {{s_k}} 	\right\|}^3}}} + \frac{{3{\epsilon_B}{{\left\| {{s_k}} \right\|}^2}}}{{\frac{{{\sigma _k}}}{6}{{\left\| {{s_k}} \right\|}^3}}} 
	=\frac{{12{\epsilon	_h}}}{{\sigma _k{{\left\| {{s_k}} \right\|}}}} + 18{\epsilon_B}\frac{1}{{{\sigma _k}\left\| {{s_k}} \right\|}}\\
	\le& \frac{{12{\epsilon_h}}}{{{{\left( {{\epsilon_H} - {\epsilon_B}} \right)}}}} +18{\epsilon_B}\frac{1}{{{\epsilon_H} - {\epsilon_B}}}.
	\end{align*}
	In order to have the lower bound radius $\Delta_k$ such that $1 - \rho  \le1-\eta$, we 	consider the parameters' setting:
	\begin{itemize}
		\item For the first term, then we define, \\
		${\epsilon_B} = \frac{1}{{36}}\left( {1 - \eta } \right)\left( {{\epsilon_H} - 	{\epsilon_B}} \right)$.
		\item For the second term, we define\\ ${\epsilon_h} \le \frac{1}{{24}}{\left( {{\epsilon_H} - {\epsilon_B}} \right)}\left( {1 - \eta } \right)$,	
	\end{itemize}
	Thus, we can see that if ${\sigma _{\text{max2}}} = \frac{9}{2}{r_2}{L_H}$, $r_2>1$, we can obtain that $1 - \rho  \le1-\eta$. 
	
	All in all, there is a large $\sigma$ such that leads to the successful iteration, that is 
	\[{\sigma _{\text{max}}} = \max \left\{ {{\sigma _\text{max1}},{\sigma _\text{max2}}} 
	\right\}.\]
\end{proof}

\subsection{ Lemma of Sub-Samplings}
\noindent\textbf{Proof of Lemma \ref{Newton:Subsample:Lemma} }
\begin{proof}
	Let us define $
	{X_i} = {f_i}( x ) - f( x ),i \in {\mathcal{S}_h}.$
	Based on Assumption 
	\ref{Newton:Assumption:bound}, we have
	\begin{align*}
	| {{X_i}} | = &| {{f_i}( x ) - f( x 
		)} | \le | {{f_i}( x )} | + 
	| {f( x )} | \le 2{\kappa _f}\\
	\Rightarrow &
	{| {{X_i}} |^2} \le 4\kappa _f^2,
	\end{align*}
	which satisfying the conditions 
	$	\mathbb{E}[ { {{X_i}} } ] = 0, \mathbb{E}[ {X_i^2} 
	] \le 4\kappa _f^2.
	$
	Define the new variable
	\begin{align*}
	{Z_1} =& {f_s}(x) - f\left( x \right) 
	= \frac{1}{{\left| {{\mathcal{S}_h}} 
			\right|}}\sum\nolimits_{j \in {{\mathcal{S}_h}}} {\left( 
		{{f_i}\left( x \right) - f\left( x \right)} \right)}, \\
	{Z_2} =& \sum\nolimits_{j \in  {{\mathcal{S}_h}}} {\left( 
		{{f_i}\left( x \right) - f\left( x \right)} \right)}. 
	\end{align*}	
	Based on  Operator-Bernstein inequality \cite{gross2010note}, We 
	give  probability about the condition in Assumption	
	\ref{Newton:Assumption:approximation}, we have $
	|Z_2| = |\mathcal{S}_h ||Z_1| \le | {{\mathcal{S}_h}} |{\epsilon_h}{\|s_k\|^2}\le |\mathcal{S}_h |{\epsilon_h}\max\{{(\Delta_\text{max})^2},(1/\sigma_\text{min})^2\}$, where $\Delta^2_c\mathop  = \limits^{{\rm{def}}}\max\{{(\Delta_\text{max})^2},(1/\sigma_\text{min})^2\}$ 
	then 
	\begin{align*}
	\Pr \left[ {\left| {{Z_2}} \right| > \left| {{\mathcal{S}_h}} 
		\right|{\epsilon_h}{\Delta^2 _{{c}}}} \right] \le 2d\exp \left( { - \frac{{{{( {| {{\mathcal{S}_h}} |{\epsilon_h}{\Delta^2_c}} )}^2}}}{{4\left| {{\mathcal{S}_h}} \right|4\kappa _f^2}}} \right) 
		\le \delta_0.
	\end{align*} 
	Thus, the cardinality of $\mathcal{S}_h$ should satisfy $\left| {{\mathcal{S}_h}} \right| 
	\ge 
	\frac{{16\kappa _f^2}}{{\epsilon
			_h^2\Delta_c^4}}\log \left( {\frac{2d}{\delta_0 }} 
	\right).$
	Furthermore,  based on Lemma \ref{Newton:Appendix:Tool:RandomSubset} and 
	Assumption 
	\ref{Newton:Assumption:Bound:Variance},  we can also have the upper 
	bounds with respect to the gradient 
	and the Hessian of $h(x)$,
	\begin{align*}
	    {\left\| {\nabla h\left( x \right) - \nabla f\left( x \right)} 
		\right\|^2} \le &\frac{{\mathbb{I}\left( {\left| {{\mathcal{S}_h}} \right| < n} \right)}}{{\left| 	{{\mathcal{S}_h}} \right|}}{H_1},\\
	{\left\| {{\nabla ^2}h\left( x \right) - {\nabla ^2}f\left( x \right)}
		\right\|^2} \le&\frac{{\mathbb{I}\left( {\left| {{\mathcal{S}_h}} \right| < n} 
			\right)}}{{\left| {{\mathcal{S}_h}} \right|}}{H_2}.
	\end{align*}
\end{proof}

\end{document}